\documentclass[10pt,twocolumn,letterpaper]{article}

\usepackage{cvpr}  

\usepackage[dvipsnames]{xcolor}

\usepackage{nicefrac,multirow}
\definecolor{BrickRed}{rgb}{0.6,0,0}
\definecolor{RoyalBlue}{rgb}{0,0,0.8}
\definecolor{TdGreen}{rgb}{0,0.5,0.2}

\definecolor{cvprblue}{rgb}{0.21,0.49,0.74}
\definecolor{jred}{rgb}{0.8,0,0}

\newcommand{\rpp}{$\mathtt{RPP}$}

\usepackage{tcolorbox}
\usepackage[pagebackref,breaklinks,colorlinks,citecolor=cvprblue]{hyperref}
\usepackage{float}
\usepackage{afterpage}
\usepackage[accsupp]{axessibility} %

\def\paperID{17663}
\def\confName{CVPR}
\def\confYear{2024}

\title{In Search of a Data Transformation That Accelerates Neural Field Training}

\author{Junwon Seo\thanks{equal contribution} \quad Sangyoon Lee$^*$ \quad Kwang In Kim \quad Jaeho Lee \\
Pohang University of Science and Technology (POSTECH)\\
{\tt\small \{junwon.seo, sangyoon.lee, kimkin, jaeho.lee\}@postech.ac.kr}
}

\begin{document}
\maketitle
\begin{abstract}
Neural field is an emerging paradigm in data representation that trains a neural network to approximate the given signal. A key obstacle that prevents its widespread adoption is the encoding speed---generating neural fields requires an overfitting of a neural network, which can take a significant number of SGD steps to reach the desired fidelity level. In this paper, we delve into the impacts of data transformations on the speed of neural field training, specifically focusing on how permuting pixel locations affect the convergence speed of SGD. Counterintuitively, we find that randomly permuting the pixel locations can considerably accelerate the training. To explain this phenomenon, we examine the neural field training through the lens of PSNR curves, loss landscapes, and error patterns. Our analyses suggest that the random pixel permutations remove the easy-to-fit patterns, which facilitate easy optimization in the early stage but hinder capturing fine details of the signal.\footnote{code: {\scriptsize \url{https://github.com/effl-lab/DT4Neural-Field}}}

\end{abstract}
\section{Introduction}
\label{sec:intro}

Neural field is a form of data representation that parameterizes each target signal as a neural network that maps spatiotemporal coordinates to the signal values \citep{xie22}.
For example, a colored image can be represented by a model that maps ($\mathtt{X}$,$\mathtt{Y}$) pixel coordinates to the corresponding ($\mathtt{R}$,$\mathtt{G}$,$\mathtt{B}$) values.
This parameterization enjoys many advantages in faithfully and efficiently representing high-dimensional signals with fine detail, and thus is being widely used for modeling signals of various modalities, such as image \citep{liif}, video \citep{nvp}, 3D scene \citep{nerf}, or spherical data \citep{generalized_inr}.


A key obstacle that prevents the widespread adoption of neural fields is their \textit{training cost}. To represent each datum as a neural field, one must train a neural network using many SGD iterations. For instance, NeRF requires at least 12 hours of training time on GPUs to represent a single 3D scene \citep{nerf}. Representing a set of data thus requires a considerable amount of computation and time, making it very difficult to develop practical applications of neural fields.

Many prior works view the ``optimization bias'' as a major cause behind the long training time of neural fields \citep{ffn}. In particular, the spectral bias of the SGD-based training is known to bias the neural network to prioritize fitting the low-frequency components of the target signal and leave high-frequency components for the late stage of training \citep{rahaman}. It has been observed that such a tendency greatly hinders the neural fields from expressing natural data (\eg, images) with fine, high-frequency details \citep{ffn,dictionary}.

The prevailing strategy to mitigate such bias is to introduce a useful \textit{prior} (or inductive bias) that can help neutralize the negative effects of the bias. One popular approach is to develop new network components that bring a favorable \textit{architectural prior}, such as the Fourier features \citep{nerf}, sinusoidal activation \citep{siren}, or spatial encoding \citep{acorn,instantngp}. Other works also attempt to introduce the prior in the form of \textit{initial parameters} that have been meta-learned from a large set of signals \citep{metasdf,learnit}. Such meta-learned initializations tend to have rich high-frequency spectra, which can help fitting natural signals within a small number of SGD steps \citep{learnit}.

\begin{figure*}[t]
\centering
\includegraphics[width=0.98\linewidth]{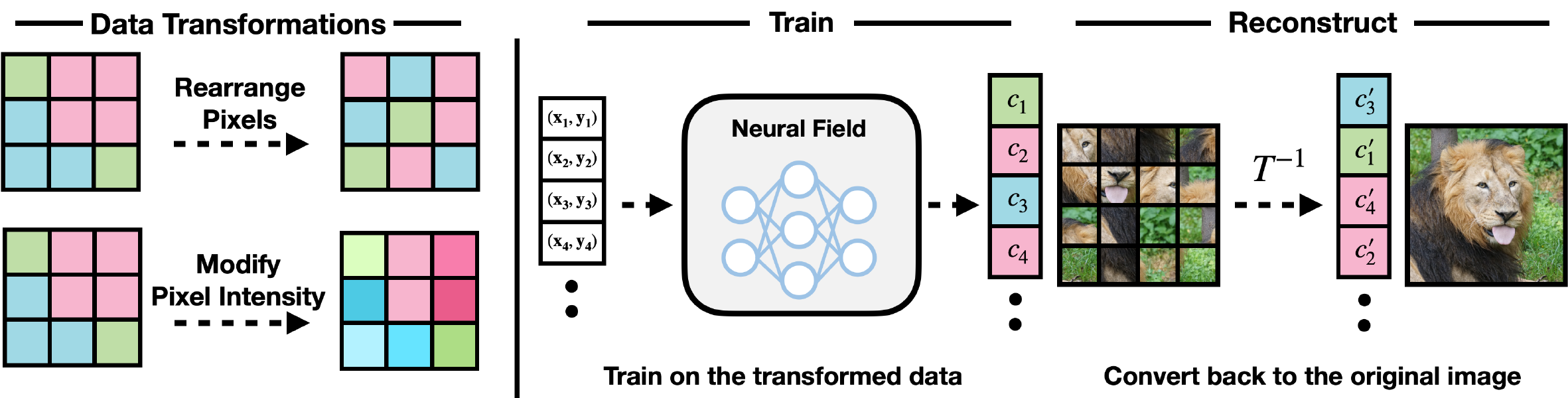}
\vspace{-0.5em}
\caption{\textbf{Overall pipeline.} We consider a three-step procedure to train neural fields. (1) Apply a data transformation to the target datum. (2) Train a neural field to fit the transformed data. (3) Reconstruct the original data by generating the transformed datum from the neural field, and then applying the inverse of the data transformation. By selecting the data transformation carefully, we can significantly reduce the computational cost required to train a neural field that achieves the desired quality of approximation.}
\label{fig:pipeline}
\vspace{-1em}
\end{figure*}

In this paper, we approach the problem from a different angle.
In particular, we ask
the following question:
\begin{center}
\vspace{-0.5em}
\textit{``Can we \textbf{exploit} the optimization bias of SGD,\\ instead of fighting against it?''}
\vspace{-0.5em}
\end{center}
Precisely, we ask whether we can \textit{transform} the datum in a way that the optimization bias acts favorably in fitting the transformed data with a neural field. If there exists such a transformation, and if it admits an efficiently computable inverse, we may be able to use the following strategy to reduce the neural field training cost: we train a neural field that expresses the transformed signal, and the original signal can be recovered from the model by applying an inverse transformation to the signal generated by the model (\cref{fig:pipeline}).

\noindent\textbf{Contribution.} As a first step to answer this question, we conduct an extensive empirical study on how applying simple data transformations affect the computational cost of training the corresponding neural field. In particular, we compare the number of SGD steps needed to fit the transformed data to a certain fidelity level (\eg, PSNR 50 for images), with the steps to fit the original data. We try total seven different data transformations, from the one that permutes the location of the pixels to the one that scales the intensity values of each pixel.

From the experiment, we make a surprising observation: we find that the \textit{random pixel permutation} (abbr. {\rpp}) provides a consistent acceleration on a range of datasets and various neural field architectures with sophisticated encoding schemes. The {\rpp} is in fact the only data transformation that provided acceleration on every experimental setups that we considered. We find that the original data requires, on average, $\sim$30\% greater number of SGD steps to achieve the similar fidelity level than the {\rpp} data. As {\rpp} tends to bias the data toward high-frequency, it is quite counter-intuitive that they can be fit faster than the original signals.

\textit{Why} does the random pixel permutation accelerate neural field training? To explain this phenomenon, we articulate the following ``blessings of no pattern'' hypothesis:
\begin{tcolorbox}[boxsep=0pt,colback=black!5]
Original data often have smooth, representative patterns that facilitate easy optimization, particularly in the early phase of learning. However, this smoothness quickly transforms into an obstacle when aiming for a sufficiently high level of fidelity. Random pixel permutation removes these easy-to-fit patterns, which accelerates optimization in the long run. ($\bigstar$)
\end{tcolorbox}

To corroborate our hypothesis $\bigstar$, we take an in-depth look into the training dynamics of the original and the {\rpp} data. We find that, indeed the training speed on the {\rpp} data is slower than on the original data during the early training phase (\cref{ssec:lossdynamics}). However, after training for a sufficient number of epochs, the neural field trained on {\rpp} data finds a ``linear loss highway'' on which the optimization is very easy; the optimization on original data fails to find one (\cref{ssec:losslandscape}). Furthermore, we find that the errors in the {\rpp} images are more evenly distributed over pixels and lack visually distinguishable structures; errors in original images tend to have periodic or axis-aligned patterns, which might be the artifacts of the encoding scheme (\cref{ssec:lossvariance}).

To sum up, our contribution can be summarized as:
\begin{enumerate}[leftmargin=*,parsep=0pt,topsep=3pt]
\item Through systematic study, we find that simple data transformations can dramatically change the training speed of neural field ($\times0.1$--$\times20$), even on state-of-the-art neural fields architectures with sophisticated encodings \citep{instantngp}.
\item We discover that random pixel permutations ({\rpp}) provide consistent acceleration over fitting the original image, providing $\times1.08$--$\times1.50$ speedups.
\item We conduct an in-depth analysis which sheds light on how {\rpp} speeds up the training by removing the easy-to-fit patterns that slow down the training eventually.
\end{enumerate}

Despite the limitation of {\rpp} that it may be difficult to be applied for applications where generalization is critical, our study has at least two potential impact areas. First, the {\rpp} itself may be useful in the applications of neural field where a strong training fidelity is the core performance criterion, such as data compression \citep{modality}. In fact, one of the core challenges in neural-field-based data compression is the slow encoding speed (see, \eg, COOL-CHIC \citep{coolchic}). Second, our study provides a concrete starting point in developing a new data transformation that strikes the balance between the training speed and the generalizability.

\section{General framework}
\label{sec:framework}

\begin{figure*}[!t]
\centering
    \begin{subfigure}[t]{0.246\linewidth}
        \centering
        \includegraphics[width=\linewidth]{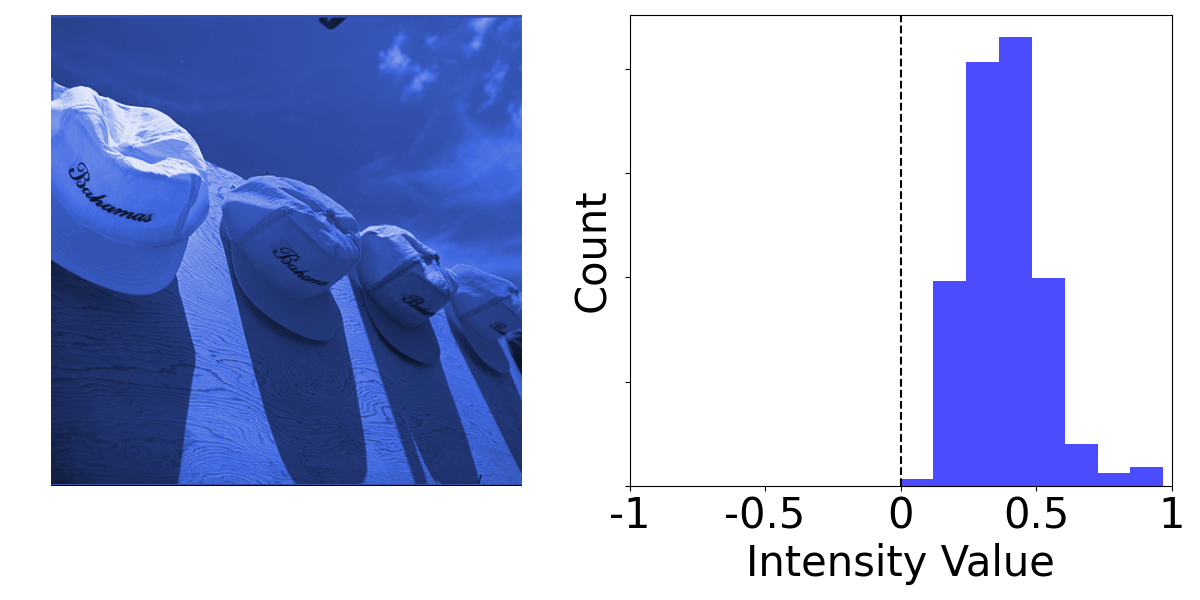}
        \captionsetup{skip=0pt}
        \caption{Original}
        \label{fig:visualize_intensity_original}
    \end{subfigure}
    \hfill
    \begin{subfigure}[t]{0.246\linewidth}
        \centering
        \includegraphics[width=\linewidth]{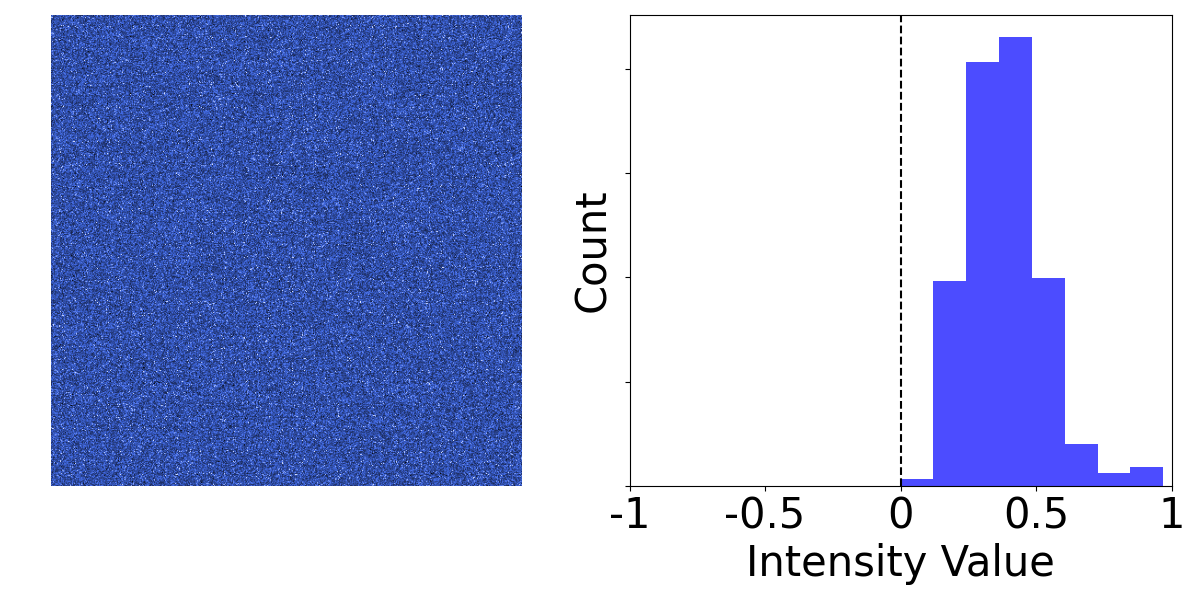}
        \captionsetup{skip=0pt}
        \caption{Random pixel permutation}
        \label{fig:visualize_intensity_rpp}
    \end{subfigure}
    \hfill
    \begin{subfigure}[t]{0.246\linewidth}
        \centering
        \includegraphics[width=\linewidth]{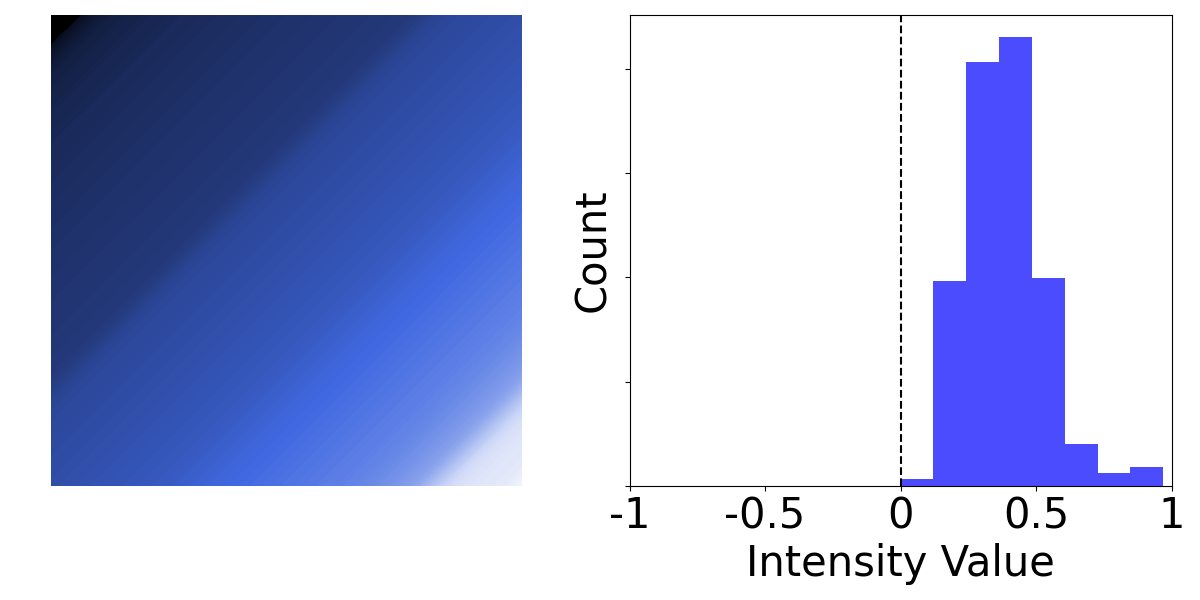}
        \captionsetup{skip=0pt}
        \caption{Zigzag permutation}
        \label{fig:visualize_intensity_zigzag}
    \end{subfigure}
    \hfill
    \begin{subfigure}[t]{0.246\linewidth}
        \centering
        \includegraphics[width=\linewidth]{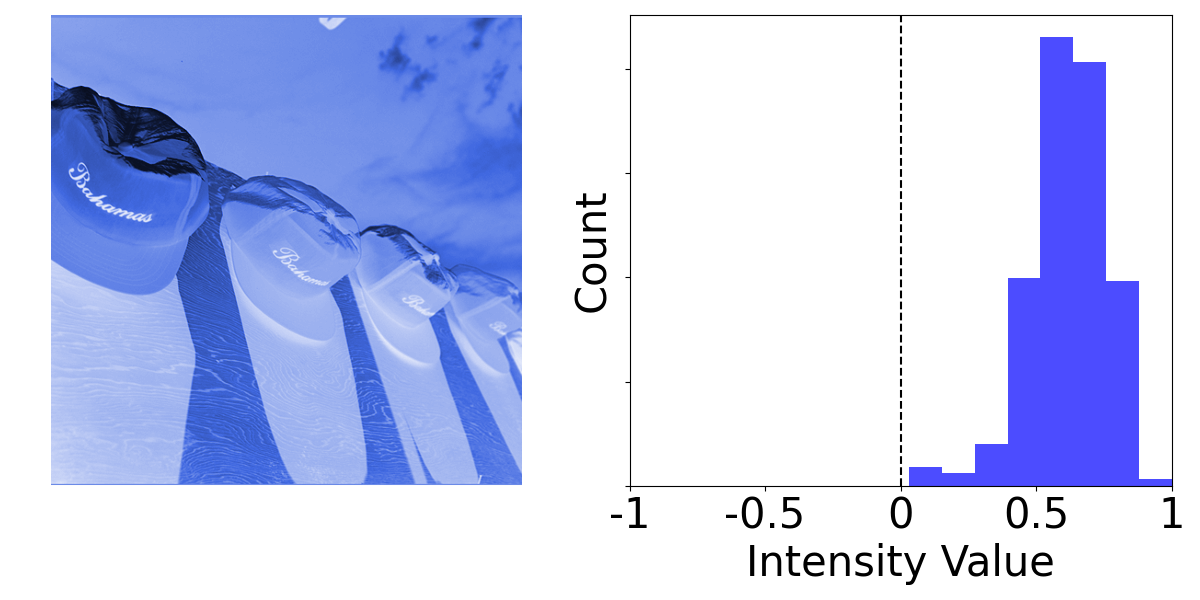}
        \captionsetup{skip=0pt}
        \caption{Inversion}
        \label{fig:visualize_intensity_inversion}
    \end{subfigure}
    \vspace{0.1cm}
    
    \begin{subfigure}[t]{0.246\linewidth}
        \centering
        \includegraphics[width=\linewidth]{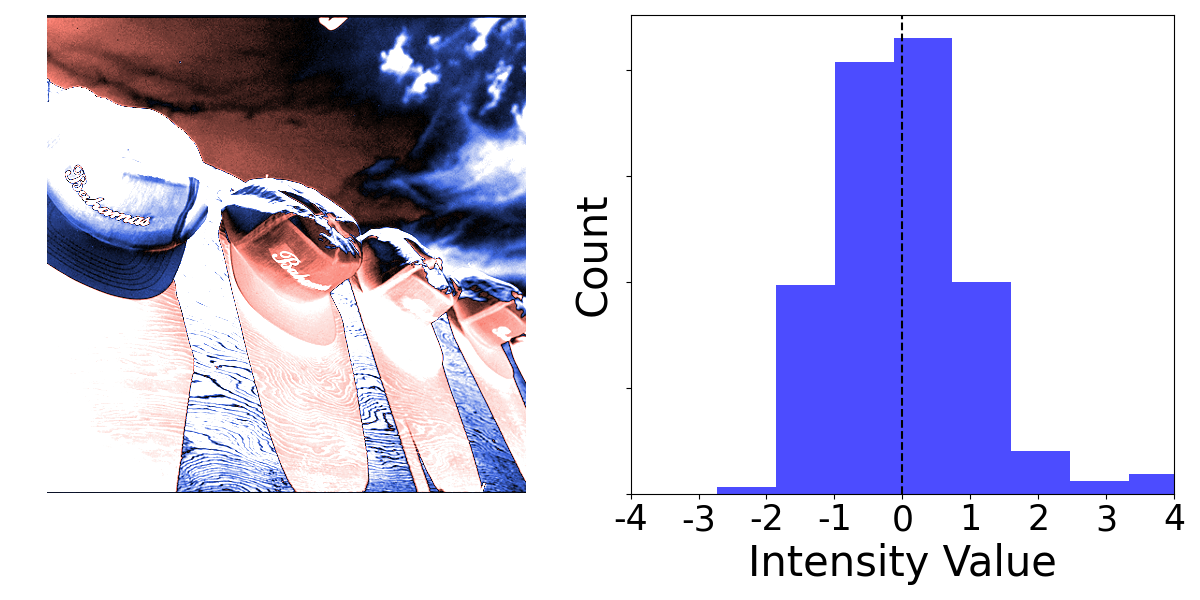}
        \captionsetup{skip=0pt}
        \caption{Standardization}
        \label{fig:visualize_intensity_std}
    \end{subfigure}
    \hfill
    \begin{subfigure}[t]{0.246\linewidth}
        \centering
        \includegraphics[width=\linewidth]{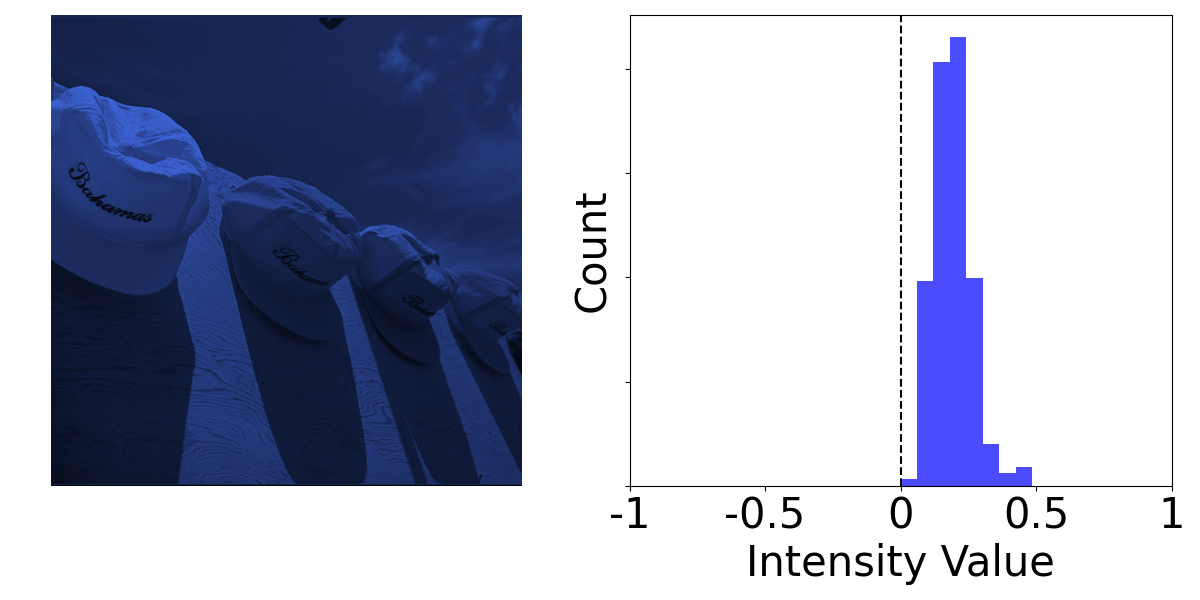}
        \captionsetup{skip=0pt}
        \caption{Linear Scaling $(t=0.5)$}
        \label{fig:visualize_intensity_linear}
    \end{subfigure}
    \hfill
    \begin{subfigure}[t]{0.246\linewidth}
        \centering
        \includegraphics[width=\linewidth]{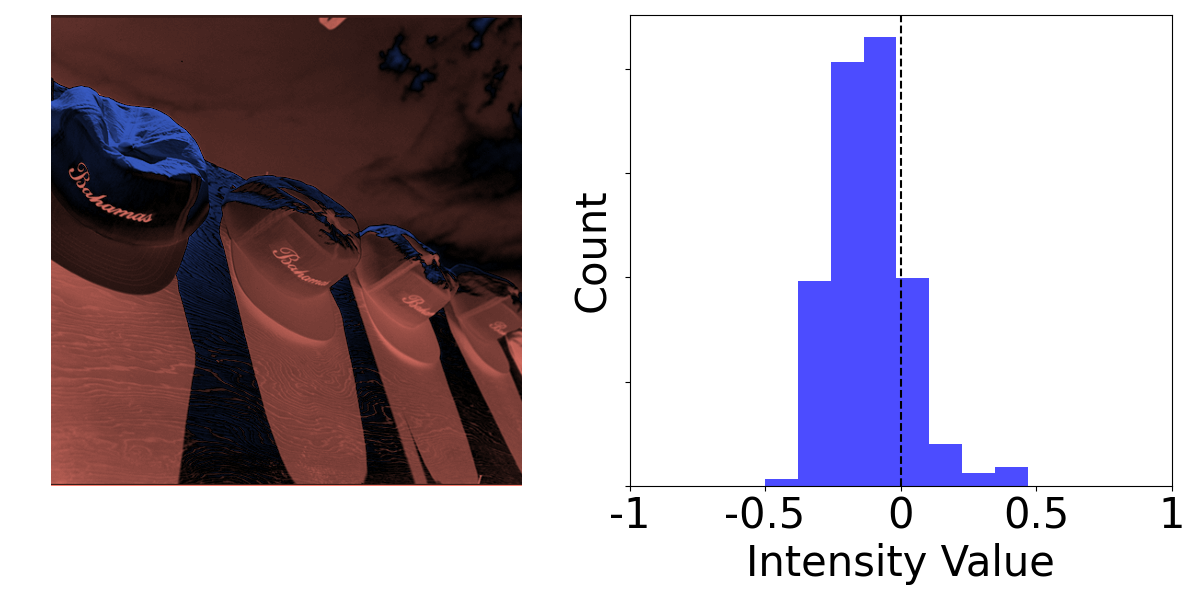}
        \captionsetup{skip=0pt}
        \caption{Centering $(t=1)$}
        \label{fig:visualize_intensity_centering}
    \end{subfigure}
    \hfill
    \begin{subfigure}[t]{0.246\linewidth}
        \centering
        \includegraphics[width=\linewidth]{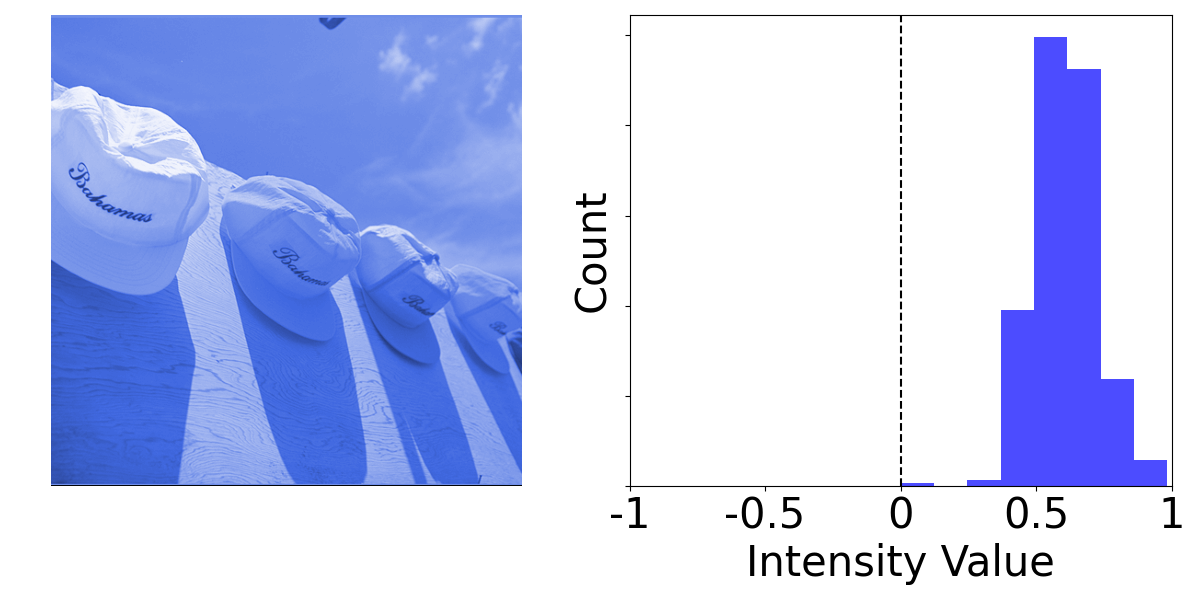}
        \captionsetup{skip=0pt}
        \caption{Gamma correction $(\gamma=2.0)$}
        \label{fig:visualize_intensity_gamma}
    \end{subfigure}
    \vspace{0.1cm}

    \begin{subfigure}[t]{0.5\linewidth}
        \centering
        \includegraphics[width=\linewidth]{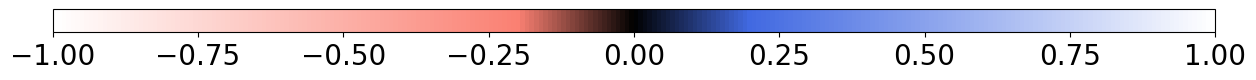}
        \label{fig:colorbar}
    \end{subfigure}
    \vspace{-0.6cm}

    \caption{\textbf{Data transformations considered.} In each subfigure, we visualize the data transformations by illustrating how the transformed image (left) and the intensity histogram (right) looks like on a Kodak image.
    }
    \label{fig:dt}
    \vspace{-1em}
\end{figure*}

The general framework of finding data transformations that accelerate neural field training can be formalized as an optimization problem, as we describe in this section.

\subsection{Formalisms}

Suppose that we want to use the neural field parameterization to approximate some \textit{signal} $\mathbf{x} \in \mathcal{X}$. Here, the \textit{signal space} $\mathcal{X}$ is the space of all signals that have the same data type. For example, we can let $\mathcal{X}$ be the set of all $256 \times 256$ RGB images; in such case, we may have $\mathcal{X} \subseteq \mathbb{R}^{256\times256\times3}$.

The \textit{data transformation} $T:\mathcal{X} \to \mathcal{X}$ maps an element in the signal space to another element. We define the \textit{transformation space} $\mathcal{T}$ as a set of data transformations on $\mathcal{X}$ that has an inverse.

Our goal is to find a data transformation $T$ that minimizes the training cost of fitting the transformed signal $T(\mathbf{x})$. To formalize this, we define the \textit{training cost}
\begin{align}
\mathsf{cost}: \mathcal{X} \times \mathcal{T} \to \mathbb{R}_+.
\end{align}
Here, the function $\mathsf{cost}(\mathbf{x},T)$ measures the computational burden required to train a neural field for $T(\mathbf{x})$, whenever its outcome is inverted back via $T^{-1}$, approximates the original signal $\mathbf{x}$ with the desired level of precision. The burden may be measured in various ways, \eg, the number of SGD steps, FLOPs, or wall-clock time. We note that the training cost of $\mathbf{x}$ using $T(\cdot)$ need not be identical to the training cost of fitting $T(\mathbf{x})$ with the same precision, \ie, 
\begin{align}
\mathsf{cost}(\mathbf{x},T) \ne \mathsf{cost}(T(\mathbf{x}),\mathrm{Id}),
\end{align}
especially when the transformation involves scaling of the signal. Also, the training cost may heavily depend on the choice of neural field architectures, hardware type, and the batch size---using a larger batch size tends to reduce the number of steps until convergence \citep{shallue}, but it also requires training with a larger memory.

Given these tools, we work to consider the optimization
\vspace{-1.5em}
\begin{align}
\mathrm{minimize}\quad \mathsf{cost}(\mathbf{x},T),\qquad \mathrm{subject\:to}\quad T \in \mathcal{T}^*, \label{eq:goal}
\end{align}
where $\mathcal{T}^* \subseteq \mathcal{T}$ is a subset of the transformation space that satisfies some desired properties. The set $\mathcal{T}^*$ can be configured in many different ways, to account for the expected usages of the trained neural fields. This problem can also be extended to a version where we have a probability  distribution of the signals $\mathbf{x}$, and find a single transformation $T$ that minimizes the expected cost.

\noindent\textbf{Scope of this paper.} While the most general framework is to solve the optimization (\ref{eq:goal}) directly, the problem is intractible unless we have an expressive, well-paramaterized transformation space $\mathcal{T}^*$ that admits an effective optimization method. In this paper, we focus on the \textit{proof-of-concept} that there exists some choice of $T$ such that
\begin{align}
\mathsf{cost}(\mathbf{x},T) < \mathsf{cost}(\mathbf{x},\mathrm{Id}), \label{eq:whatwedo}
\end{align}
(\cref{sec:observation}), and deepening our understanding on when and why such $T(\cdot)$ can accelerate the training (\cref{sec:in_depth}).


\subsection{Example cases}\label{ssec:examplecases}

Here are some examples desired properties of data transformation, and how they relate to practical applications. We provide a more in-depth discussions in the \cref{sec:comprehensive_review}.

\noindent\textbf{Efficient invertibility.} The inverse $T^{-1}$ should be able to be computed efficiently, \eg, by performing a linear operation. Ideally, one may wish to be able to combine the inverse transformation into the neural field by modifying the parameters of the trained neural field directly. For instance, if $T(\mathbf{x}) = -\mathbf{x}$, we can incorporate the inverse in the neural field by negating the final layer weights of the neural field. This property is useful whenever we expect many repeated inferences of the neural field, \eg, for real-time interaction.

\noindent\textbf{Retains interpolatability.} We want our transform $T$ to admit a principled way to sample the value of interpolated coordinates from the neural field that approximates the transformed signal. In some neural field applications, \eg, super-resolution \citep{liif}, this interpolating capability plays an essential role. On the other hand, some applications like data compression do not involve any interpolation \citep{modality}.


\section{Data transformations vs. training speed}\label{sec:observation}

\renewcommand{\arraystretch}{1.2}
\begin{table*}[t!]

\centering
\resizebox{\textwidth}{!}{%

\begin{tabular}{@{}llcccccccccc@{}}
\hline
                             \multirow{2}{*}{Architecture}& \multirow{2}{*}{Dataset}        & \multirow{2}{*}{\begin{tabular}[c]{@{}c@{}}Random\\ pixel perm.\end{tabular}} & \multirow{2}{*}{\begin{tabular}[c]{@{}c@{}}Zigzag\\ perm.\end{tabular}} & \multirow{2}{*}{Inversion} & \multirow{2}{*}{Standardization} & \multicolumn{2}{c}{Linear Scaling} & \multicolumn{2}{c}{Centering} & \multicolumn{2}{c}{Gamma Correction} \\ \cline{7-12} 
                 & & & & & & $t=0.5$ & $t=2.0$ & $t=1.0$ & $t=2.0$ &  $\gamma=0.5$ & $\gamma=2.0$\\ \hline
\multirow{3}{*}{SIREN}       & Kodak   & \colorbox{blue!10}{1.26$\times$}                                                                        & \colorbox{blue!10}{17.90$\times$}                  & 0.72$\times$                      & \colorbox{blue!10}{2.20$\times$}              & 0.79$\times$            & \colorbox{blue!10}{1.39$\times$}                 &     0.76$\times$          &    0.97$\times$           & 0.36$\times$             & 0.80$\times$                    \\
                             & DIV2K   & \colorbox{blue!10}{1.30$\times$}                                                                         & \colorbox{blue!10}{21.74$\times$}                  & 0.70$\times$                          & \colorbox{blue!10}{2.36$\times$}                      & 0.82$\times$            & \colorbox{blue!10}{1.45$\times$}        & 0.97$\times$             & \colorbox{blue!10}{1.22$\times$}             & 0.30$\times$                 & 0.90$\times$                         \\
                             & CLIC    & \colorbox{blue!10}{1.08$\times$}                                                                             & \colorbox{blue!10}{17.07$\times$}                       & 0.68$\times$                          & \colorbox{blue!10}{2.18$\times$}                  & 0.76$\times$                & \colorbox{blue!10}{1.36$\times$}                  & 0.87$\times$             & \colorbox{blue!10}{1.14$\times$}             & 0.24$\times$                 & 0.82$\times$                           \\ \hline
\multirow{3}{*}{Instant-NGP} & Kodak   & \colorbox{blue!10}{1.50$\times$}                                                                         & 0.86$\times$                   & 0.21$\times$                      & 0.45$\times$             & \colorbox{blue!10}{1.15$\times$}            & 0.75$\times$                     & \colorbox{blue!10}{1.24$\times$}         & 0.80$\times$         & 0.29$\times$             & 0.38$\times$                 \\
                             & DIV2K   & \colorbox{blue!10}{1.32$\times$}                                                                         & 0.80$\times$                   & 0.20$\times$                      & 0.61$\times$               & \colorbox{blue!10}{1.12$\times$}            & 0.64$\times$                & \colorbox{blue!10}{1.39$\times$}         & 0.97$\times$         & 0.14$\times$             & 0.33$\times$                    \\
                             & CLIC    & \colorbox{blue!10}{1.35$\times$}                                                                         & 0.78$\times$                   & 0.07$\times$                       & 0.53$\times$            & \colorbox{blue!10}{1.03$\times$}             & 0.68$\times$                  & 0.99$\times$          & 0.66$\times$         & 0.12$\times$              & 0.32$\times$                      \\ \hline
\end{tabular}
}
\vspace{-0.5em}

\caption{\textbf{Acceleration factors of data transformations.} We compare the average acceleration factor (eq.~\ref{eq:accfac}) of seven data transformations under six different combinations of datasets and models. We take average of cost ratios, instead of ratios of averages, to avoid the statistic being driven by a small number of outliers that require very long training time. The \colorbox{blue!10}{shaded} figures denote the speedup.}
\label{table:big_table}
\vspace{-1em}
\end{table*}

In this section, we compare the training cost of the neural field training on various data transformation schemes. As it turns out, there indeed exists a nice data transformation that satisfies ineq.~\ref{eq:whatwedo}: the \textit{random pixel permutation} ({\rpp}).

\subsection{Experimental setup}

We focus on the task of 2D image regression. This choice reduces the computational burden of fitting each datum, so that we can make experimental validations on multiple datasets and models with extensive hyperparameter tuning.

For the training cost, we use \textit{the number of SGD steps} until we reach the training PSNR over 50dB (a near-perfect reconstruction for 8-bit images). The number of SGD steps is an informative indicator of the total FLOPs and runtime for any fixed choice of hardware, model architecture, and data. Importantly, we tune the learning rate for each (transformed) image, making it close to the scenario where all settings have been optimized to minimize the runtime.

For each choice of an image and the data transformation, we measure the \textit{acceleration factor}, \ie,
\begin{align}
\mathsf{acc}(\mathbf{x},T) \triangleq \frac{\mathsf{cost}(\mathbf{x},\mathrm{Id})}{\mathsf{cost}(\mathbf{x},T)}. \label{eq:accfac}
\end{align}
Other configurations are as follows.

\noindent\textbf{Datasets.} We use three different image datasets:
\begin{itemize}[leftmargin=*,parsep=0pt,topsep=0pt]
\item \textit{Kodak.} Consists of 24 different images from the Kodak lossless true color image suite \citep{kodak}.
\item \textit{DIV2K.} Consists of 100 images in the validation split of the DIV2K image super-resolution challenge (HR) \citep{div2k}.
\item \textit{CLIC.} Consists of 100 images from the validation split of the dataset for the Challenge on Learned Image Compression 2020 (CLIC) \citep{clic}. There are total 102 images in the validation split, but we removed two images that have a sidelength shorter than 512 pixels.
\end{itemize}
We pre-process each image in the datasets as follows. Following BACON \citep{bacon}, we resize each image into 512$\times$512 pixels by first center-cropping the image to a square with sidelengths equal to the shorter sidelength of the original image, and then resizing with the Lanczos algorithm. Next, we convert the image to grayscale. Finally, we apply sRGB to Linear RGB operation, as in Instant-NGP \citep{instantngp}.

The resulting images have 512$\times$512 pixels (thus total $2^{18}$ pixels) with intensities lying in the interval $[0,1]$.

\noindent\textbf{Models.} We use two different neural field architectures.
\begin{itemize}[leftmargin=*,parsep=0pt,topsep=0pt]
\item \textit{SIREN.} A classical architecture that uses multi-layered perceptrons with sinusoidal activation functions \citep{siren}. We configure the model to have three hidden layers with 512 neurons in each layer and output dimension 1. We use the default frequency scaling hyperparameter $\omega_0 = 30$.
\item \textit{Instant-NGP.} A more recently proposed neural field with multi-resolution hash encodings \citep{instantngp}. We use the default setup for the image regression, with the maximum hash table size and 16-bit parameters.
\end{itemize}

\noindent\textbf{Training.} We use full-batch gradient descent, \ie, the batch size $2^{18}$. This batch size, in general, minimizes the number of SGD steps required to fit the target image. We note that we sample directly from the coordinate grid, instead of sampling from interpolated coordinates (as in, \eg, \citep{instantngp}).

\noindent\textbf{Hyperparameters.} We tune the learning rate using the grid search. For SIREN, we tune the learning rate in the range $\{2^{-8},\ldots,2^{-16}\}$. For Instant-NGP, we tune the learning rate in the range $\{2^{-4},\ldots,2^{-15}\}$.

\noindent\textbf{Data transformations.} We consider total seven elementary data transformations (\cref{fig:dt}): Two transformations that only change the location of the pixels (random pixel permutation, zigzag permutation), and five transformations that change the intensity values of each pixels (inversion, standardization, linear scaling, centering, and gamma correction).
\begin{itemize}[leftmargin=*,topsep=0pt,parsep=0pt]
\item \textit{Random pixel permutation} ({\rpp}). We randomly permute the location of each pixels. This transformation generally increases the frequency spectrum of the image.
\item \textit{Zigzag permutation.} We sort all the pixels in the ascending order of their intensities. Then, we place the sorted pixels in the zigzag order, starting from the upper left corner. This produces a low-frequency image.
\item \textit{Inversion.} We invert the intensity of each pixel, \ie, perform $z \mapsto 1-z$ on each intensity values.
\item \textit{Standardization.} We measure the mean $\mu$ and standard deviation $\sigma$ for the intensities of all pixels in the image. Then we standardize the intensities via $z \mapsto \frac{z-\mu}{\sigma}$.
\item \textit{Linear scaling.} We scale the intensities of each pixel by $t$ without any centering, \ie $z \mapsto tz$.
\item \textit{Centering.} We center the interval $[0,1]$ at zero and scale the intensities by $t$, \ie, $z \mapsto t(z-\nicefrac{1}{2})$.
\item \textit{Gamma correction.} We nonlinearly scale the intensity of each pixels as $z \mapsto z^{1/\gamma}$. Choosing $\gamma > 1$ makes images brighter, and $\gamma < 1$ makes darker.

\end{itemize}




\subsection{Results}
We report the average acceleration factors in \cref{table:big_table}. From the table, we make several observations:
\begin{itemize}[leftmargin=*,topsep=0pt,parsep=0pt]
\item {\rpp} provides a consistent acceleration over the original image; see \cref{sec:in_depth} for an in-depth analysis.
\item The zigzag permutation works tremendously well on SIREN---speeding up training by over $\times17$---but slows down training Instant-NGP. We hypothesize that this is due to the axis-aligned inductive bias imposed by the spatial grid encoding; see \cref{ssec:lossvariance} for more discussion.
\item Inversion consistently slows down the training. This fact, ironically, implies that inversion might have been an effective accelerator if all ``natural images'' looked like the color-inverted images.
\item Scaling up the intensities tends to have opposite effects on different architectures. Scaling up speeds up training in SIRENs, but slows down in Instant-NGPs.
\item Gamma correction slows down the training, regardless of taking the power of $2$ or $\nicefrac{1}{2}$. We suspect that this is due to the precision errors from taking powers. In other words, excessive scaling operations may not be helpful in accelerating the training.
\end{itemize}


\section{A closer look at the random permutation}
\label{sec:in_depth}
\begin{figure}[!t]
\centering
    \begin{subfigure}[t]{0.49\linewidth}
        \centering
        \includegraphics[width=\linewidth]{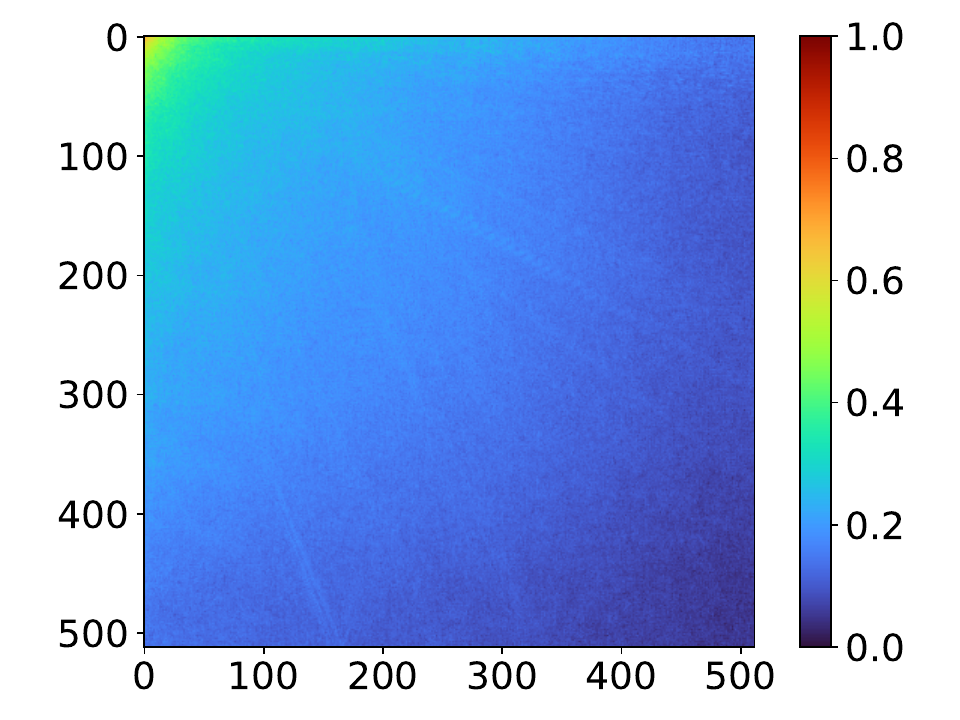}
        \caption{Original}
        \label{fig:dct_orig_kodak}
    \end{subfigure}
    \hfill
    \begin{subfigure}[t]{0.49\linewidth}
        \centering
        \includegraphics[width=\linewidth]{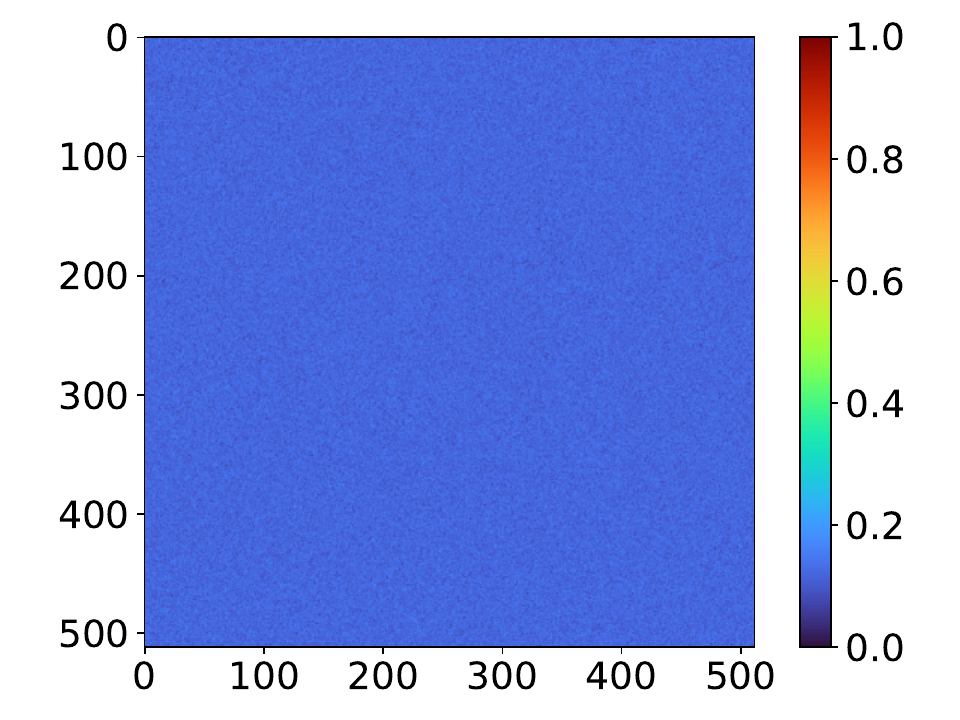}
        \caption{{\rpp}}
        \label{fig:dct_per_kodak}
    \end{subfigure}
    \vspace{-1em}
    \caption{\textbf{Frequency spectra of original vs. {\rpp}.} We compare the average DCT coefficients of the original and {\rpp} Kodak images. Upper left region denotes the low-frequency, and the lower right region denotes the high-frequency.}
    \label{fig:vis_dct_kodak}
    \vspace{-1em}
\end{figure}
We now focus on a specific type of data transformation: the random pixel permutation (\rpp). The {\rpp} relocates each pixel to a new random location, \ie,
\begin{align}
(\mathtt{coords}_i,\mathtt{values}_i) \mapsto (\mathtt{coords}_{\pi(i)},\mathtt{values}_i)
\end{align}
for some random permutation $\pi(\cdot)$. The marginal distributions of the input coordinates or the output values remain the same. As we have seen in \cref{table:big_table}, the {\rpp} transformation accelerates the fitting of the corresponding neural field, which is very unexpected and difficult to explain.

\noindent\textbf{Why is this strange?} From the perspective of the \textit{spectral bias}, the {\rpp} operation should have slowed down the training speed, instead of accelerating it. As demonstrated by \citet{rahaman}, typical neural networks first rapidly fit the low-frequency components of the target signal, and fit the high-frequency components much later. The {\rpp} transformation, in this sense, should have been detrimental to the training. In fact, {\rpp} images tend to have more higher-frequency components than the original images. In \cref{fig:vis_dct_kodak}, we provide the average discrete cosine transform (DCT) coefficients of the original and {\rpp} images; for visualization, we raise the coefficients to the power of $0.03$, similar to \citep{dctvis}. We observe that the original image is more biased toward low-frequency than {\rpp}. However, we also observe that {\rpp} images fit faster than the original images.

\begin{figure}[t]
\centering
\includegraphics[width=0.95\linewidth]{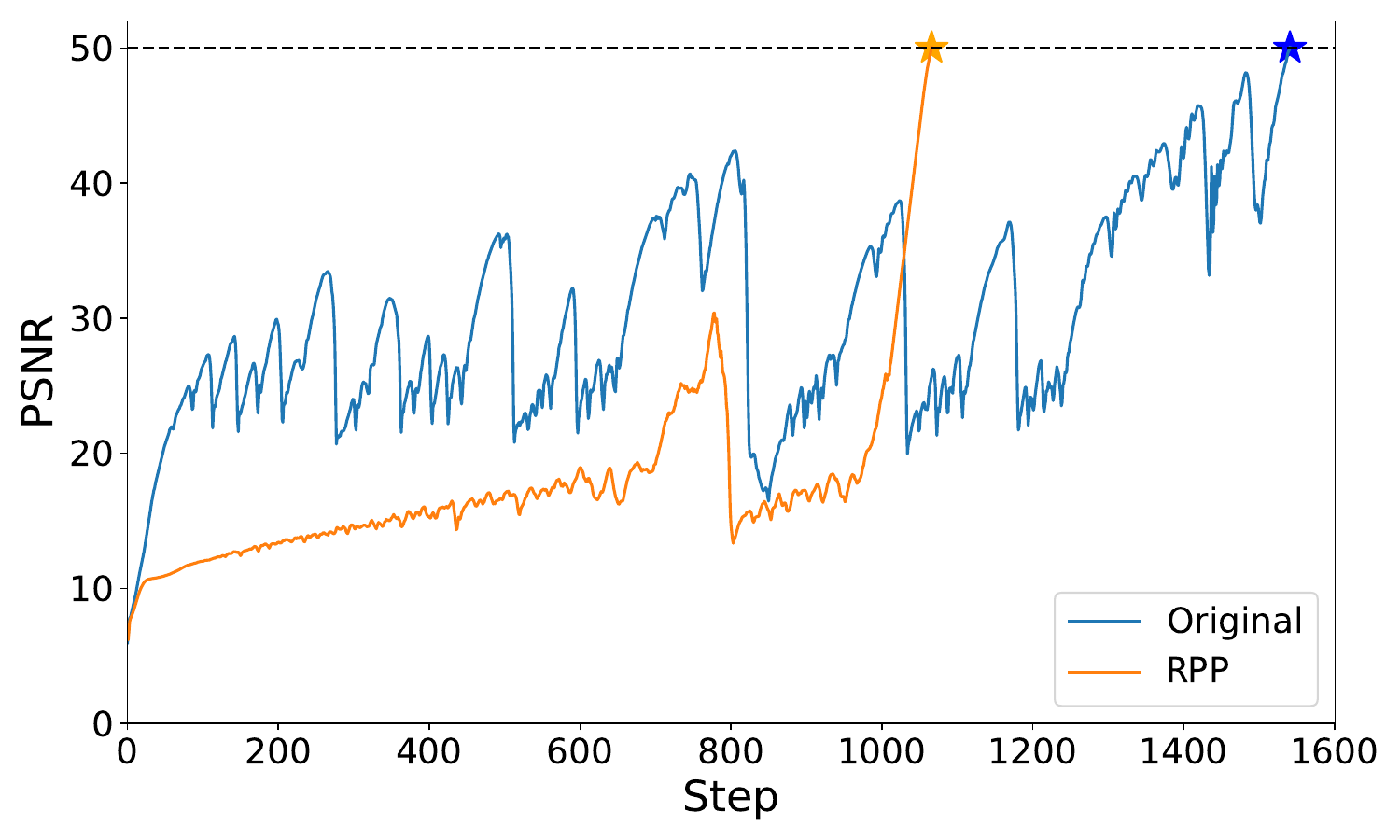}
\vspace{-1em}
\caption{\textbf{PSNR curves for a single Kodak image.} The original image (orange) excels during the early stage of training but the {\rpp} image quickly reaches PSNR 50 in the middle of the training.}
\label{fig:learning_curve}
\vspace{-1em}
\end{figure}
\renewcommand{\arraystretch}{1.1}
\begin{table}[t]
\centering
\resizebox{0.35\textwidth}{!}{%
\begin{tabular}{@{}lccl@{}}
\hline
                   & \multicolumn{2}{c}{Target PSNR} & \multirow{2}{*}{Ratio} \\ \cline{2-3}
                   & 30dB             & 50dB             &                        \\ \hline
Original           & 105.8          & 1371.4         & 0.08                   \\
Random pixel perm. & 851.8          & 1100.5         & 0.79                   \\ \hline
\end{tabular}
}
\vspace{-0.5em}
\caption{\textbf{The number of steps to reach PSNR 30dB and 50dB.} We measure the average steps to reach moderate and high PSNR levels over Kodak images, and report the average ratio.}
\label{table:learning_curve_ratio}
\vspace{-1em}
\end{table}

In the remainder of this section, we identify three different aspects that the training dynamics of {\rpp} images are critically different from that of original images. These aspects, when put together, support our ``blessings of no pattern'' hypothesis ($\bigstar$) that {\rpp} accelerates training by removing easy-to-fit patterns that can be harmful to reaching a high level of fidelity. In particular, we show that
\begin{itemize}[leftmargin=*,topsep=0pt,parsep=0pt]
\item {\rpp} is slow to reach moderate PSNR levels, but is fast to reach high PSNR (\cref{ssec:lossdynamics}).
\item there is a linear path that connects the moderate-to-high PSNR points in the {\rpp} loss landscape (\cref{ssec:losslandscape}).
\item model trained on {\rpp} images tends to have less structured error, while vanilla training leads to error patterns that reflect the underlying encoding schemes (\cref{ssec:lossvariance}).
\end{itemize}

\subsection{The PSNR curve: Slower to approximate well, but faster to approximate ``very well''}\label{ssec:lossdynamics}

First, we observe that {\rpp} images tend to achieve moderate PSNRs (\eg, 30dB) much later than the original images, but they quickly reach high PSNRs (\eg, 50dB) afterwards.

In \cref{fig:learning_curve}, we provide an example plot of fitting a SIREN on a natural image from the Kodak dataset. From the figure, we can immediately make the following observations:
\begin{itemize}[leftmargin=*,topsep=0pt,parsep=0pt]
\item \textit{RPPs are slow starters.} The training PSNR of the original image quickly arrives at $\sim$ 30dB the early stage of training, even before taking 300 SGD iterations. On the other hand, the PSNR of the {\rpp} image remains very small, under 15 dB at the similar number of steps.
\item \textit{Explosive surge in later stage.} After staying at a low PSNR level (under 20 dB) for a while, the PSNR of {\rpp} images bursts explosively. The PSNR jumps from $\sim$ 15dB to 50dB in less than 100 SGD steps. On the other hand, the PSNR curve of the original image oscillates wildly at 30--35dB level for a long time, and gradually reaches the 50dB at around 1600 steps.
\end{itemize}

These observations are consistent over the choice of the target image; we provide a comprehensive collection of figures in the supplementary materials.

For a more quantitative comparison, we compare the average number of SGD steps to arrive the PSNR of 30dB and 50dB for the original and {\rpp} images (\cref{table:learning_curve_ratio}). Again, we use the images from the Kodak dataset and fit with SIRENs.

We observe that, for original images, the number of steps to reach 30dB is around 100 steps, while the number of SGD steps to reach the 50dB is over 1300 steps. That is, it takes almost $13\times$ more steps
to reach the high PSNR than for moderate PSNRs. On the other hand, for {\rpp} images, it takes over 800 steps to reach the moderate PSNR level, while it takes only 300 more steps to reach the high PSNR.

\subsection{Linear paths in the loss landscape: RPP images find a ``linear expressway''}\label{ssec:losslandscape}

\begin{figure}[t]
\centering
        \begin{subfigure}[t]{0.49\linewidth}
            \centering
            \includegraphics[width=\linewidth]{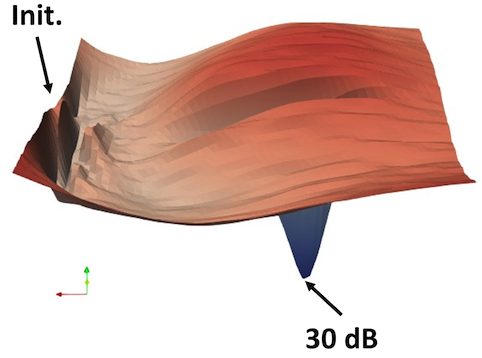}
            \caption{Elevated view; original}
            \label{fig:original_loss_landscape_0_30_hole}
        \end{subfigure}
        \hfill
        \begin{subfigure}[t]{0.49\linewidth}
            \centering
            \includegraphics[width=\linewidth]{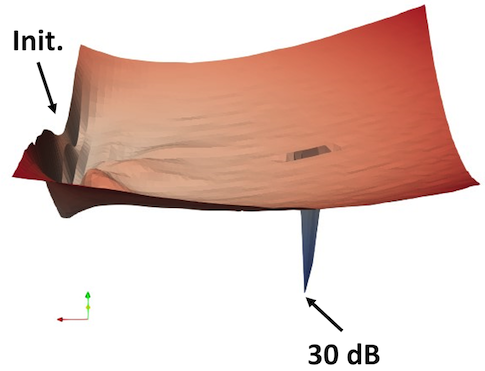}
            \caption{Elevated view; {\rpp}}
            \label{fig:permute_loss_landscape_0_30_hole}
        \end{subfigure}
        \begin{subfigure}[t]{0.49\linewidth}
            \centering
            \includegraphics[width=\linewidth]{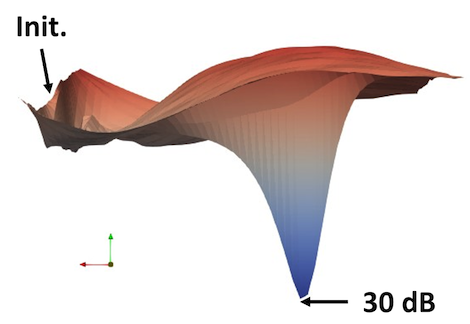}
            \caption{Side view; original}
            \label{fig:original_loss_landscape_0_30_valley}
        \end{subfigure}
        \hfill
        \begin{subfigure}[t]{0.49\linewidth}
            \centering
            \includegraphics[width=\linewidth]{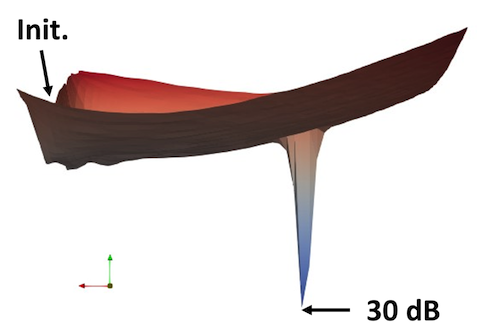}
            \caption{Side view; {\rpp}}
            \label{fig:permute_loss_landscape_0_30_valley}
        \end{subfigure}

        \caption{\textbf{SIREN loss landscape: from initial point to 30dB point.} For original and {\rpp} versions of a Kodak image, we plot the loss landscape between the initial point and the parameter that achieves PSNR 30. For {\rpp}, the minima is much narrower and lacks a clear pathway toward it, unlike in the original image.}
\label{fig:loss_land_scape_0_30}
\vspace{-1em}
\end{figure}

        

\begin{figure}[t]
\centering
        \begin{subfigure}[t]{0.49\linewidth}
            \centering
            \includegraphics[width=\linewidth]{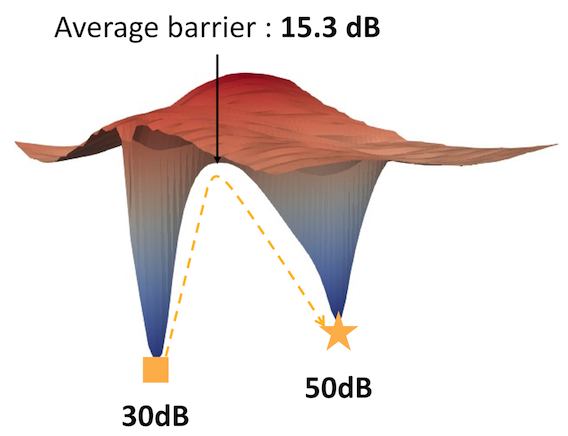}
            \caption{Original}
            \label{fig:original_loss_land_scape_30_50}
        \end{subfigure}
        \hfill
        \begin{subfigure}[t]{0.49\linewidth}
            \centering
            \includegraphics[width=\linewidth]{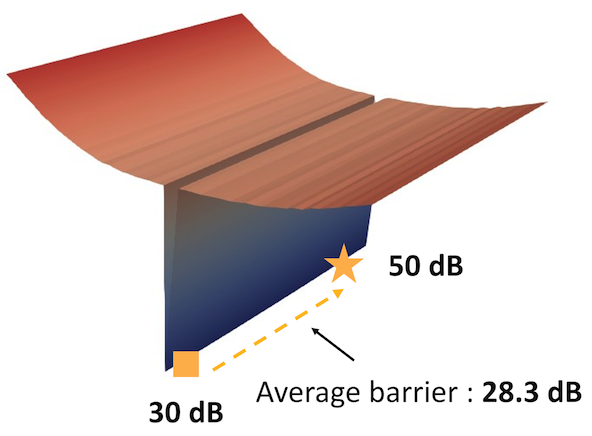}
            \caption{\rpp}
            \label{fig:permute_loss_land_scape_30_50}
        \end{subfigure}
        \caption{\textbf{SIREN loss landscape: from 30dB point to 50dB point.} For original and {\rpp} versions of a Kodak image, we plot the loss landscape between the parameters that first achieve PSNR 30 ({\color{Dandelion}$\blacksquare$}) and PSNR 50 ({\color{Dandelion}$\bigstar$}), respectively. The loss barriers between two minima are dramatically different; 15.3dB for the original image, and 28.3dB for the {\rpp} (averaged over all Kodak images).
        }
\label{fig:loss_land_scape_30_50}
\vspace{-1em}
\end{figure}

\begin{figure*}[!t]
\centering
    \begin{subfigure}[t]{0.32\linewidth}
        \centering
        \includegraphics[width=\linewidth]{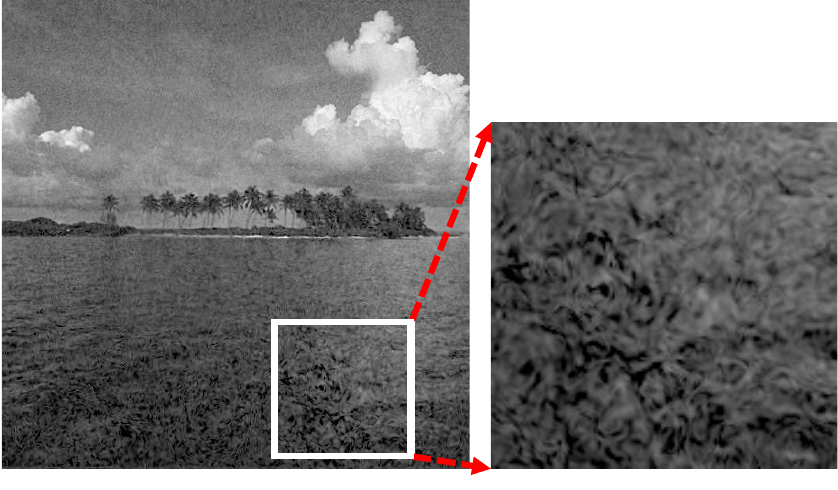}
        \captionsetup{skip=0pt}
    \end{subfigure}
    \hfill
    \begin{subfigure}[t]{0.32\linewidth}
        \centering
        \includegraphics[width=\linewidth]{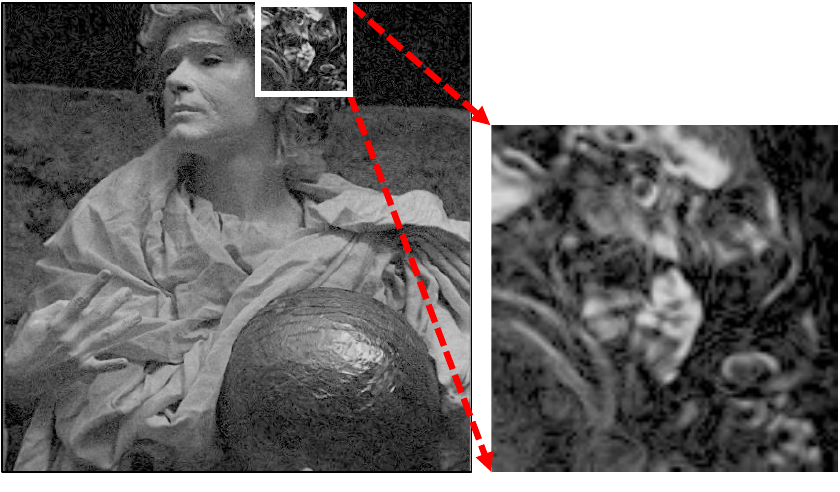}
        \captionsetup{skip=0pt}
        \caption{Original}
        \label{fig:recon_siren_statue_orig}
    \end{subfigure}
    \hfill
    \begin{subfigure}[t]{0.32\linewidth}
        \centering
        \includegraphics[width=\linewidth]{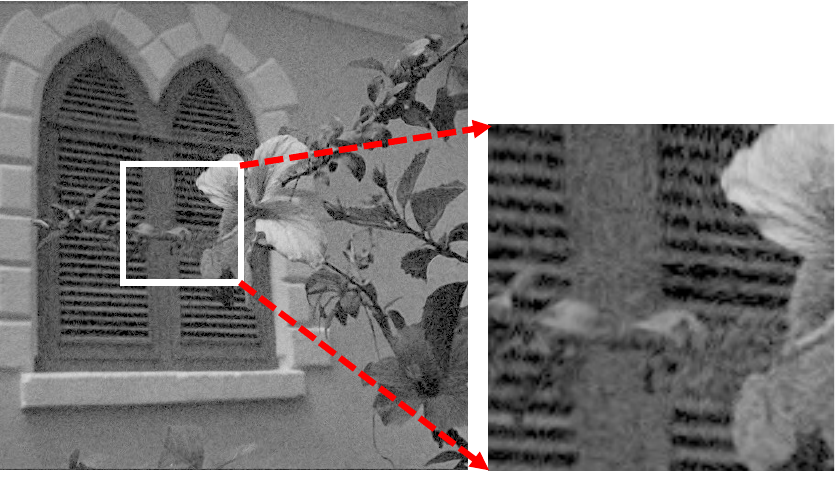}
        \label{fig:recon_siren_flower_orig}
    \end{subfigure}
    
    \vspace{0.1cm}
    \begin{subfigure}[t]{0.32\linewidth}
        \centering
        \includegraphics[width=\linewidth]{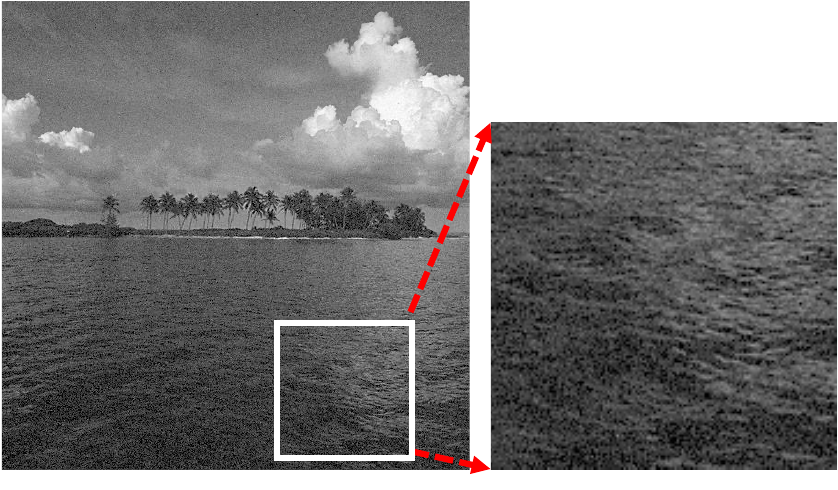}
        \captionsetup{skip=0pt}
    \end{subfigure}
    \hfill
    \begin{subfigure}[t]{0.32\linewidth}
        \centering
        \includegraphics[width=\linewidth]{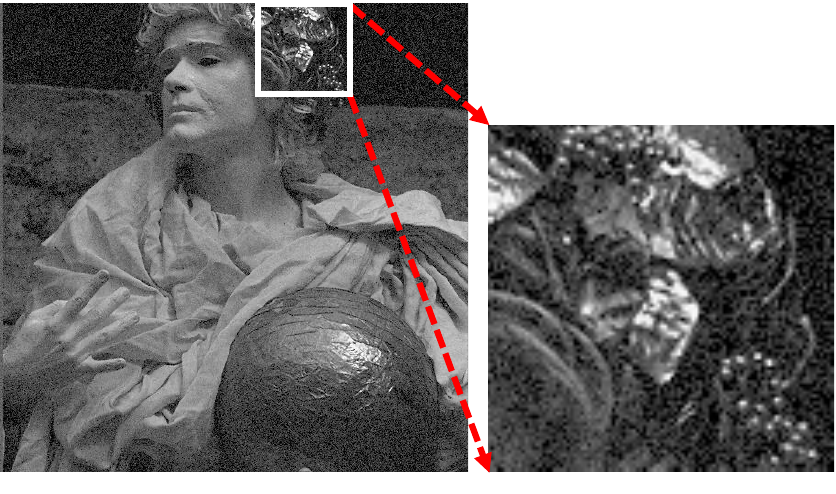}
        \captionsetup{skip=0pt}
        \caption{{\rpp}}
        \label{fig:recon_siren_statue_permute}
    \end{subfigure}
    \hfill
    \begin{subfigure}[t]{0.32\linewidth}
        \centering
        \includegraphics[width=\linewidth]{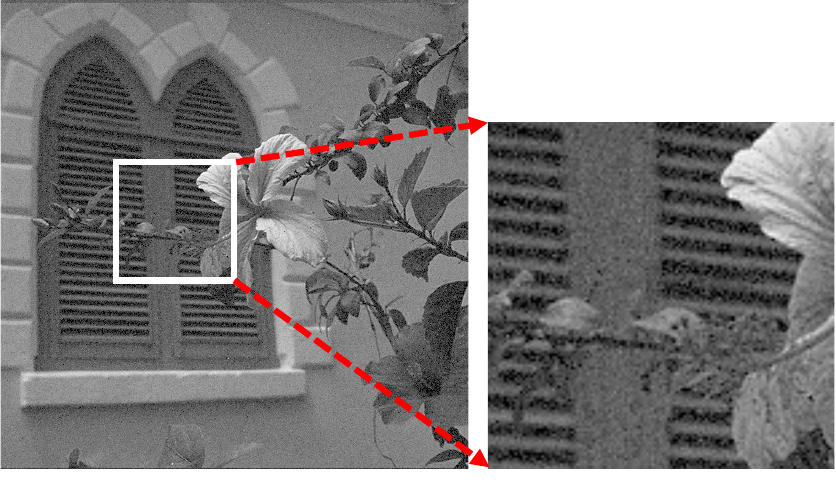}
        \captionsetup{skip=0pt}
    \end{subfigure}
    \vspace{-1em}
    \caption{\textbf{Example reconstructions at a target PSNR value of 30 (SIREN).} Random pixel permutation enables the network to prioritize the restoration of high-frequency, fine details, whereas the original representation tends to steer the network toward restoring low-frequency components initially. While this behavior may pose challenges during the early stages of training, it ultimately facilitates achieving visually more plausible reconstructions much earlier than the original representation.}
    \label{fig:recon_siren}
    \vspace{-1em}
\end{figure*}

From the previous observation, it seems likely that, for {\rpp} images, there may exist an \textit{expressway} in the neural field parameter space that connects the moderate-PSNR parameter with the high-PSNR parameter; on this expressway, the SGD may not encounter too many hills to circumvent, and may mostly follow a smooth linear path.

To validate this intuition, we visualize the loss landscape of the SIREN trained on Kodak images. In particular, we capture two different phases of training.
\begin{enumerate}[leftmargin=*,topsep=0pt,parsep=0pt]
\item \textit{Early phase.} We plot the linear path from the initialization to 30dB point, \ie, the parameter that first achieves PSNR 30dB during the training (\cref{fig:loss_land_scape_0_30})
\item \textit{Late phase.} We plot the linear path from the 30dB point to the PSNR 50dB point (\cref{fig:loss_land_scape_30_50}).
\end{enumerate}
To generate such visualization, we project the parameter space into a two-dimensional space, as in \citep{losslandscape}. Unlike \citep{losslandscape}, we fix one axis to a directional vector between two parameter points that we want to capture (\eg, 30dB point and 50dB point); the other axis has been decided randomly.\footnote{A similar method has been considered by \citet{goodlandscape}.}

During the early phase (\cref{fig:loss_land_scape_0_30}), we observe that the loss landscape of {\rpp} is quite hostile; unlike in original images, the minima that 30dB point belongs to is quite narrow and there is no clear pathway toward the 30dB point.

During the late phase (\cref{fig:loss_land_scape_30_50}), for the {\rpp} image, once the parameter arrives at the minima, there exists a linear path that connects the 30dB point to the 50dB point. If we measure the loss barrier, \ie the maximum loss on the linear path between 30dB and 50dB points, it is as low as 28.3dB for the {\rpp} versions of Kodak images. For original images, the loss barrier is quite high (15.3dB).





\subsection{Patterns in the error: The errors are less structured in RPP images}\label{ssec:lossvariance}

\begin{figure}[!t]
\centering
    \begin{subfigure}[t]{0.98\linewidth}
        \centering
        \includegraphics[width=\linewidth]{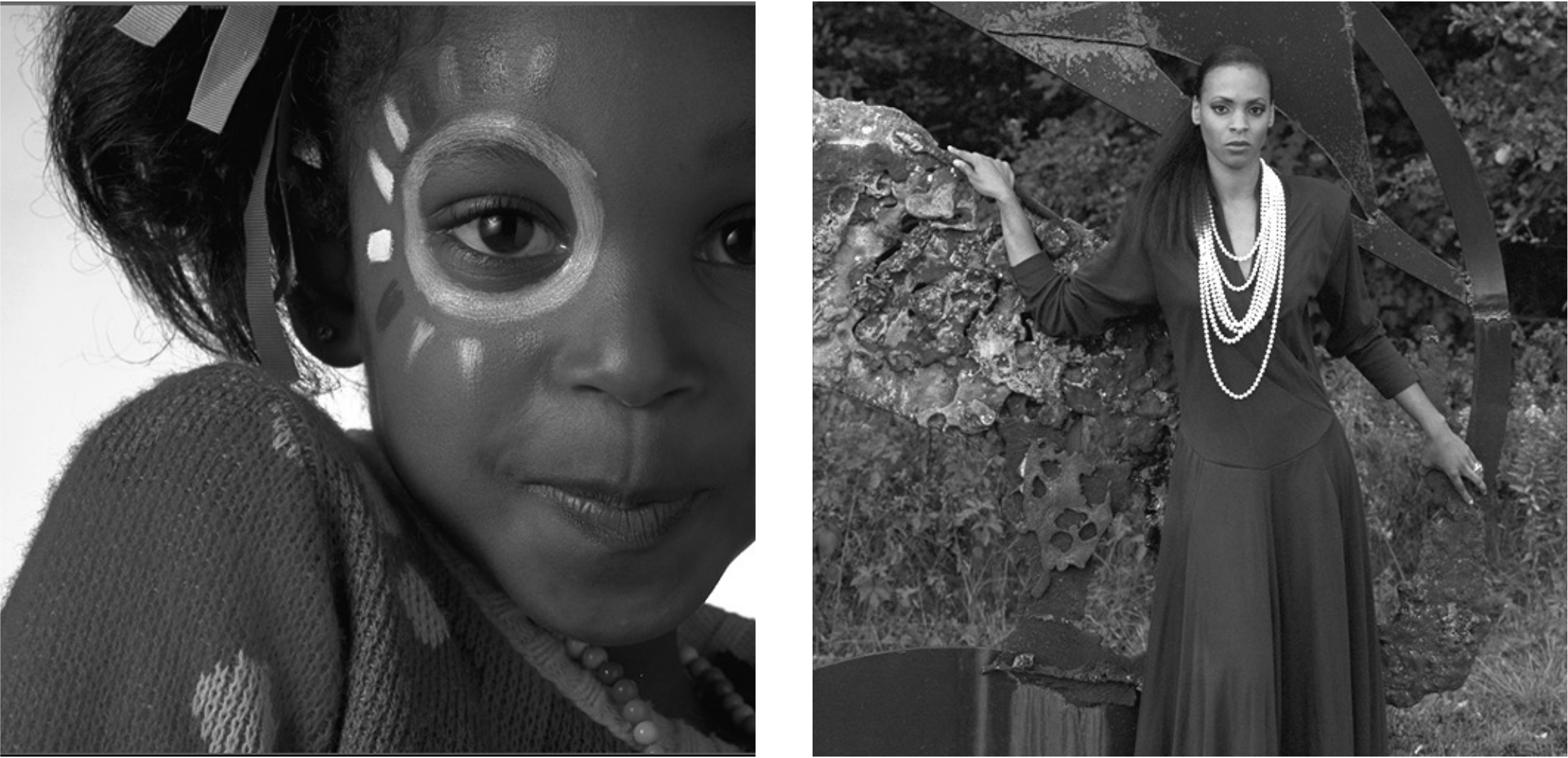}
        \caption{Ground Truth}
        \label{fig:recon_ngp_15_target}
    \end{subfigure}

    \begin{subfigure}[t]{0.98\linewidth}
        \centering
        \includegraphics[width=\linewidth]{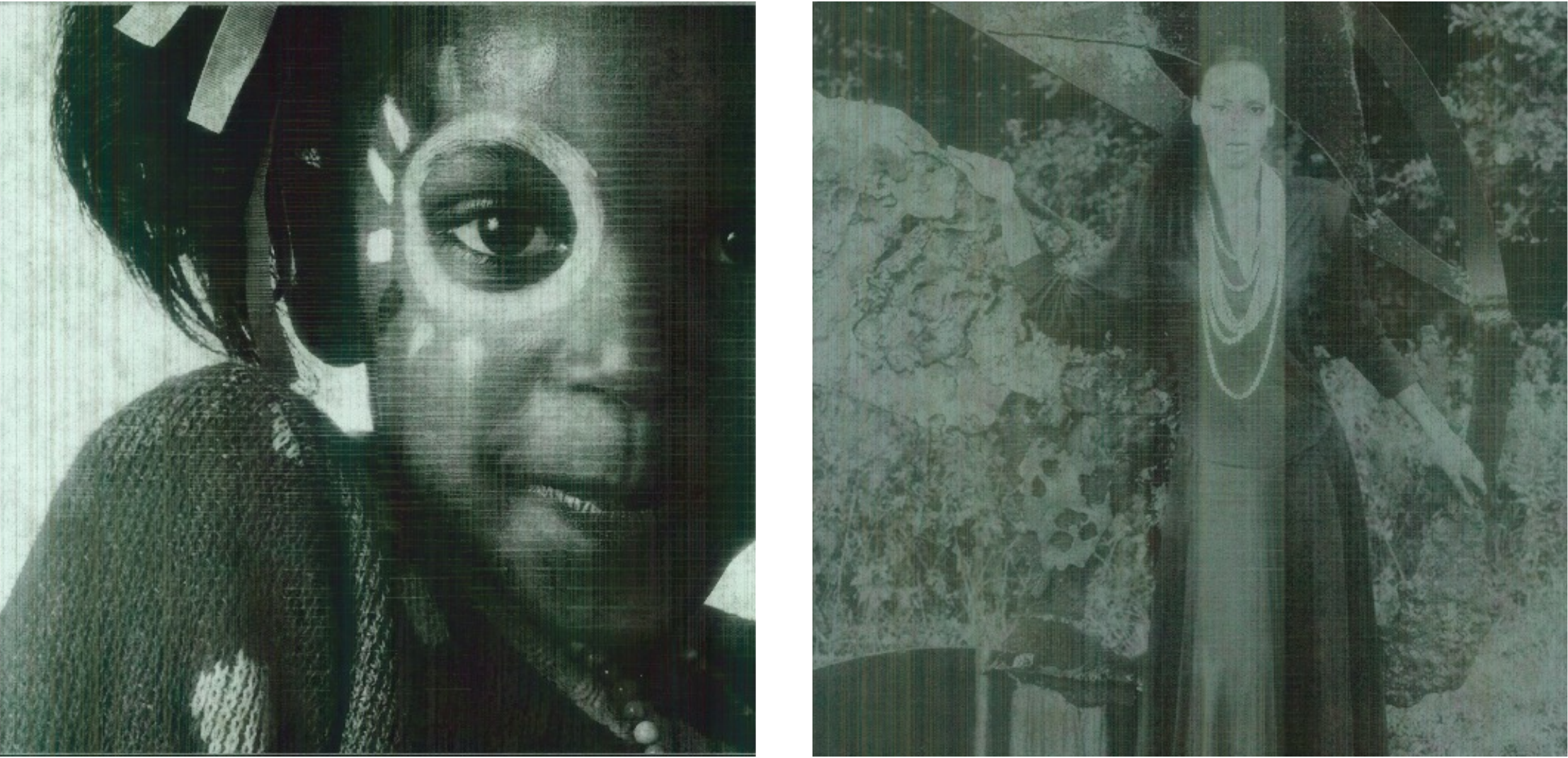}
        \caption{Original}
        \label{fig:recon_ngp_15_psnr20_shuffled}
    \end{subfigure}

    \begin{subfigure}[t]{0.98\linewidth}
        \centering
        \includegraphics[width=\linewidth]{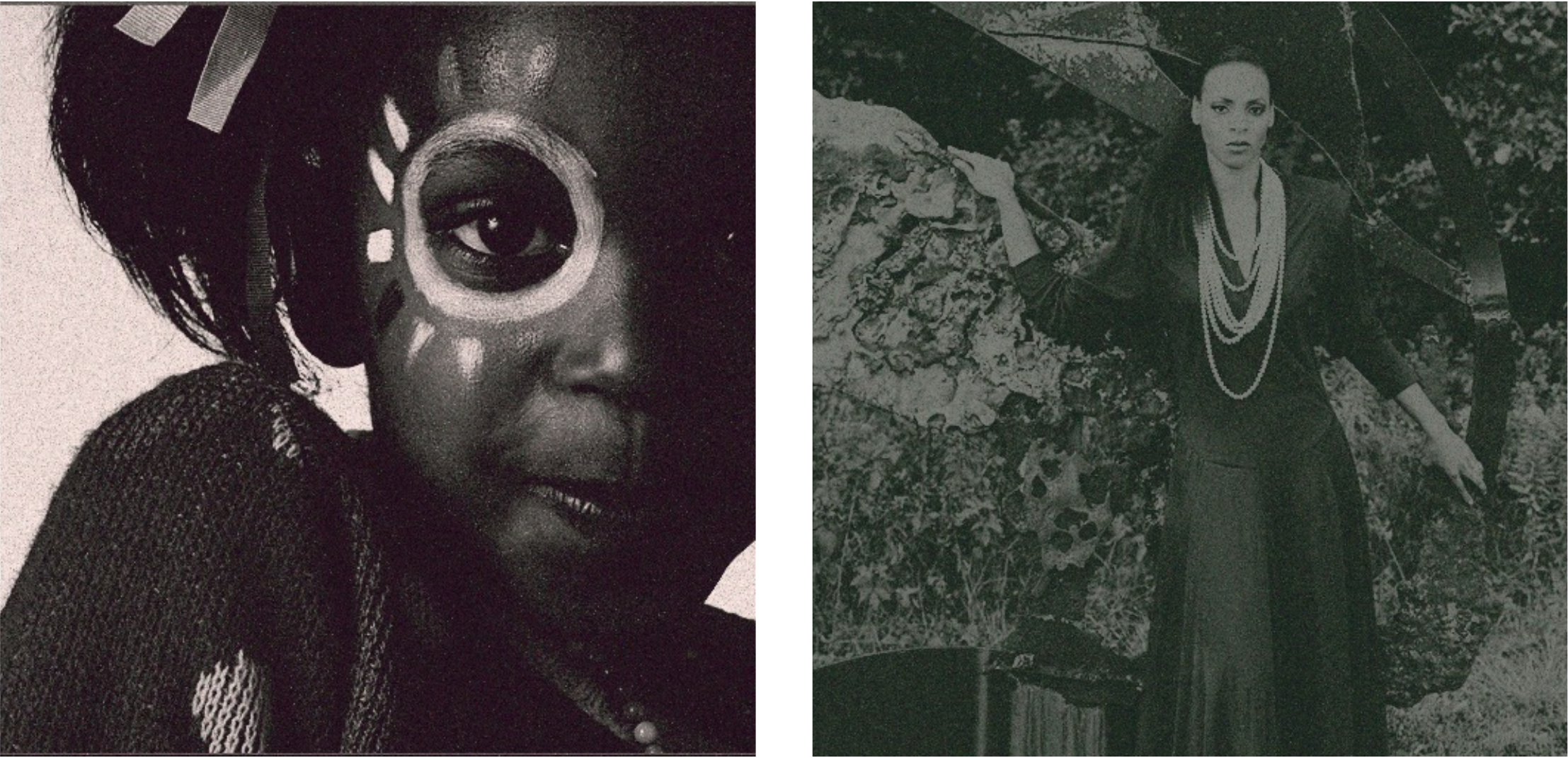}
        \caption{{\rpp}}
        \label{fig:recon_ngp_18_psnr20_orig}
    \end{subfigure}
    \vspace{-1em}
    \caption{\textbf{Example reconstructions at a target PSNR value of 20 (Instant-NGP).} Similarly to the SIREN case, we observe that Instant-NGPs trained on the {\rpp} images have much less visually distinguishable artifacts. The models trained on original images tend to have axis-aligned blocks of error, which is likely to have originated from the spatial grid encoding of the model.}
    \label{fig:recon_ngp}
    \vspace{-1em}
\end{figure}

\renewcommand{\arraystretch}{1.2}
\begin{table}[t]
\centering
\resizebox{0.47\textwidth}{!}{%
\begin{tabular}{@{}lrrrr@{}}
\hline
                   & \multicolumn{4}{c}{Target PSNR}           \\ \cline{2-5} 
                   & \multicolumn{1}{c}{20}       & \multicolumn{1}{c}{30}       & \multicolumn{1}{c}{40}       & \multicolumn{1}{c}{50}       \\ \hline
Original             & 10.51e-04 & 23.68e-06 & 19.46e-08 & 5.90e-10 \\
Random pixel perm. & 5.68e-04 & 5.89e-06 & 7.59e-08 & 5.81e-10  \\ \hline
Average ratio      & \multicolumn{1}{c}{2.03$\times$}     & \multicolumn{1}{c}{5.20$\times$}    & \multicolumn{1}{c}{5.80$\times$}    & \multicolumn{1}{c}{1.55$\times$}    \\ \hline
\end{tabular}
}
\vspace{-0.5em}
\caption{\textbf{Loss variance over pixels.}
We measure the average pixel loss variance in the SIREN trained on Kodak dataset, measured at various levels of PSNR. The ``average ratio'' denotes the mean of $\nicefrac{\mathrm{Var}(\mathrm{original})}{\mathrm{Var}(\mathtt{RPP})}$. {\rpp} has a lower loss variance at all PSNRs.}
\label{table:Loss_variance_ratio}
\vspace{-1em}
\end{table}


Another interesting property of {\rpp} is that the errors that the neural field makes are more evenly distributed among the pixels, lacking a clearly distinguishable pattern. It turns out that the \textit{whiteness} of the {\rpp} error can help generating more visually sharp and satisfactory images than neural fields trained on original images.

In \cref{table:Loss_variance_ratio}, we measure the pixel loss variance in both original and {\rpp} images trained to various target PSNR levels. From the table, we observe that the loss variance is up to $5.80\times$ larger in the neural fields trained on the original images than on {\rpp} images. This implies that, for original images, some pixels in the original image are fit much faster than other pixels. Putting that differently, the neural field trained on the unpermuted images tends to prioritize learning a specific type of patterns first, while the neural field trained on {\rpp} images does less so.

Exactly what pattern is {\rpp} avoiding to fit? In \cref{fig:recon_siren,fig:recon_ngp}, we compare the images generated by the neural fields trained on original and {\rpp} images on SIREN and Instant-NGP, respectively. In \cref{fig:recon_siren}, we observe that SIRENs trained on original images have thicker wavy patterns, which is likely to be a consequence of the sinusoidal encoding structure of SIREN. The models trained on {\rpp} images, on the other hand, have much sharper texture with fine-grained noise. A similar phenomenon is observed in the case of Instant-NGP (\cref{fig:recon_ngp}). Here, we observe that the models trained on original images attain axis-aligned artifacts; black (or white) horizontal (or vertical) blocks appear in the rendered images. On the other hand, the error is much less structured in models trained on {\rpp} images.



\section{Related work}
\label{sec:related}

\noindent\textbf{Optimization bias of neural networks.} It has been well known, both theoretically and empirically, that SGD-based optimization algorithms are biased toward learning ``simpler solutions,'' for various notions of simplicity. A prominent example is the \textit{spectral bias}, a tendency to prioritize learning lower-frequency components of the target function \citep{frequency_bias,rahaman}. Other types of biases have also been studied in the literature: SGD tends to prefer learning smaller-norm solutions \citep{implicitregularization,gunasekar} and low-rank solutions \citep{simpler,yangsalman}. These works mostly focus on explaining how such simplicity biases are advantageous in learning a solution that can generalize well (see, \eg, \citep{cohenrazin}). Our work, in contrast, describes how such bias can be \textit{disadvantageous} in terms of the training speed, instead of the generalization performance.

\noindent\textbf{Random labels and memorization.} The question of how neural nets behave when trained on random labels has been actively studied since its connection to the generalization capability of deep learning has been established \citep{rethinking}. \citet{memorization} finds that neural networks tend to learn simple patterns first and then \textit{memorize} the unexplainable samples---such as random label and data---in the later phase. A more recent study argues that training on randomly labeled data can actually be advantageous \citep{whatdo}; pre-training on random data helps fitting the training data, when eventually fine-tuning on the dataset with correct labels. Different from these works, our work discovers how random label can be beneficial in the neural field context, without any consideration on further fine-tuning.

\noindent\textbf{Faster training of neural fields.} Other than the works described in the introduction, there have been many other attempts to accelerate the neural field training by designing better encoding schemes. Positional encodings \citep{nerf,ffn,siren,mfn} and spatial grids \citep{nsvf,plenoxel,instantngp,acorn} are the most popular options, and some works also use tree-like structures \citep{implicit3D,plenoctree}. Our work is complementary to this line of works, and argues that transforming the data in the original data space can provide auxiliary training speed boosts. We also note that a concurrent work also aims to boost the training speed by modifying the given data \citep{partition}. In particular, \citep{partition} finds that separating the visual signals into many sub-segments can facilitate the convergence of SIRENs and proposes a meta-learning algorithm to speed up training. Our work considers a more general class of data transformations, and focus on understanding how transformations accelerate training.


\section{Conclusion}
\label{sec:conc}

Learning neural fields poses a unique challenge, as the primary objective is to \emph{overfit} to the given data---an aim contrary to the typical goals in other problem domains where learning attempts to achieve \emph{regularization} to prevent overfitting. In our study of this distinctive problem, we discovered that certain data transformations can expedite the network's acquisition of fine details more rapidly than when learning from the original images. Specifically, we demonstrated that random pixel permutations, which explicitly make learning low-frequency details more challenging, guide the network away from fixating on low-frequency components. This resulted in significantly enhanced PSNR within a given computational budget and, furthermore, yielded visually more plausible reconstructions at the same PSNR level. Our insights are substantiated by thorough empirical evaluations.

\noindent\textbf{Limitation \& future direction.} A notable limitation of our approach is that its direct applicability may be constrained when learning neural networks for other problem domains such as classification and regression, where regularization is crucial. As a future work, we aim to conduct more comprehensive studies to understand how data transformations affect both trainability and generalization of neural fields (and other neural networks). Such exploration may offer new insights on how to modify the given workload to jointly optimize the training cost and the model performance.

Another, more narrowly scoped future direction will be to gain a more concrete, theoretical understanding on the effect of random pixel permutations on the training dynamics of neural fields. Although our empirical analyses provide some insights into how {\rpp} can help accelerate the training, we are yet to fully understand why fitting the easy patterns slows down the later phase of training. Deeper understandings on this matter may lead to impactful lessons applicable to any deep learning domain that involves long training.

\noindent\textbf{Acknowledgments.}
This work was partly supported by the National Research Foundation of Korea (NRF) grant (RS-2023-00213710, Neural Network Optimization with Minimal Optimization Costs), and partly by the Institute of Information \& communications Technology Planning \& Evaluation (IITP) grant (No.2022-
0-00713, Meta-learning applicable to real-world problems, No.2019-0-01906, Artificial Intelligence Graduate School Program (POSTECH)), all funded by the Korea government (MSIT).

{
    \small
    \bibliographystyle{ieeenat_fullname}
    \bibliography{main}
}

\clearpage
\newpage
\setcounter{page}{1}
\maketitlesupplementary

\appendix
\renewcommand{\thesection}{S\arabic{section}}

\renewcommand{\thefigure}{S\arabic{figure}}
\renewcommand{\thetable}{S\arabic{table}}



\hypersetup{%
  linkcolor  = black
}
\makeatletter
\newcommand \Dotfill {\leavevmode \cleaders \hb@xt@ .99em{\hss .\hss }\hfill \kern \z@}
\makeatother

\renewcommand{\labelenumii}{\theenumii}
\renewcommand{\theenumii}{\theenumi.\arabic{enumii}.}

\section*{Contents}
\begin{enumerate}[label=S\arabic*]
    \item \hyperref[sec:wall_clock]{\textbf{The actual speedup on Kodak dataset.}} \hfill
    \textbf{\pageref{sec:wall_clock}} \\
    \item \hyperref[sec:sup_average]{\textbf{The average number of steps for achieving target\\ PSNR}} \hfill \textbf{\pageref{sec:sup_average}}\\
    \item \hyperref[sec:single_rpp]{\textbf{Utilizing a single permutation matrix for RPP}} \hfill
    \textbf{\pageref{sec:single_rpp}} \\
    \item \hyperref[sec:sup_dct]{\textbf{DCT coefficient of other datasets}} \hfill \textbf{\pageref{sec:sup_dct}}\\
    \item \hyperref[sec:sup_lc_curve]{\textbf{Full PSNR curves on the Kodak dataset}} \hfill \textbf{\pageref{sec:sup_lc_curve}}\\
    \item \hyperref[sec:sup_reconstruct_ngp]{\textbf{Reconstructed images in Instant-NGP}} \hfill \textbf{\pageref{sec:sup_reconstruct_ngp}}\\
    \item \hyperref[sec:sup_loss_landscape]{\textbf{Additional loss landscapes}} \hfill \textbf{\pageref{sec:sup_loss_landscape}}
    \begin{enumerate}
        \item \hyperref[subsec:sup_loss_landscape_other_images]{Landscapes on other images} \Dotfill \quad\pageref{subsec:sup_loss_landscape_other_images}
        \item \hyperref[subsec:sup_loss_landscape_different_vector]{Projections on different direction vectors} \Dotfill \quad \pageref{subsec:sup_loss_landscape_different_vector}
    \end{enumerate}
\end{enumerate}

\hypersetup{%
  linkcolor  = red
}




\renewcommand{\arraystretch}{1.2}
\begin{table*}[h]

\centering
\resizebox{\textwidth}{!}{%

\begin{tabular}{@{}llccccccccccc@{}}
\hline
                             \multirow{2}{*}{Architecture}& \multirow{2}{*}{Dataset}   & \multirow{2}{*}{Original}     & \multirow{2}{*}{\begin{tabular}[c]{@{}c@{}}Random\\ pixel perm.\end{tabular}} & \multirow{2}{*}{\begin{tabular}[c]{@{}c@{}}Zigzag\\ perm.\end{tabular}} & \multirow{2}{*}{Inversion} & \multirow{2}{*}{Standardization} & \multicolumn{2}{c}{Linear Scaling} & \multicolumn{2}{c}{Centering} & \multicolumn{2}{c}{Gamma Correction} \\ \cline{8-13} 
                 &  &      &                                                                         &                         &                            &                                  & $t=0.5$             & $t=2.0$             & $t=1.0$               & $t=2.0$              & $\gamma=0.5$              & $\gamma=2.0$             \\ \hline
\multirow{3}{*}{SIREN}       & Kodak & 1371.4 & 1100.5                                                                         & 79.8                  & 2271.4                      & 706.7            & 1875.2           & 1051.0                    &     2071.7         &    1591.7          & 4297.1           & 1835.4                \\
                             & DIV2K & 1568.6 & 1366.3                                                                        & 73.3                 & 2512.1                       & 778.8     & 2033.6         & 1153.6                         & 1892.2            & 1532.3           & 6630.2                & 2102.6                     \\
                             & CLIC &  1257.1 & 1233.7                                                                             & 73.5                      & 2089.9                         & 683.6  & 1832.4              & 1014.3                              & 1663.3            & 1334.1            & 6359.3                 & 1806.8                            \\ \hline
\multirow{3}{*}{Instant-NGP} & Kodak   & 207.3  & 217.1 & 	240.6 & 	3121.9 & 	761.7 & 	168.3 & 	304.0 & 	167.1 & 	258.6 & 	1861.2 & 	677.3  \\
                             & DIV2K   & 194.0 & 146.7 & 	247.1 & 	4240.3 & 	439.1 & 	164.6 & 	505.6 & 	151.4 & 	245.2 & 	4207.9 & 	1369.1 \\
                             & CLIC    & 199.0 & 151.4 & 	238.5 & 	3881.8 & 	490.4 & 	168.2 & 	502.3 & 	152.0 & 	257.4 & 	4773.7 & 	1418.4  \\ \hline
\end{tabular}
}
\caption{\textbf{Average steps for achieving PSNR 50dB.} We report the average steps of 8 data transformations (including original) under six different combinations of datasets and models. The instances that do not achieve the target PSNR with any of the learning rates are excluded from the average calculation. Gamma correction with $\gamma$ = 2.0 fails to fit (1, 2, 1) images on Kodak, DIV2K, and CLIC datasets, respectively. Similarly, $\gamma$ = 0.5 and Linear Scaling with t = 2.0 do not fit (0, 6, 2) and (1, 2, 1) images in these datasets, respectively.}
\label{table:big_table_sup}
\end{table*}


\newpage

\quad
\newpage

\section{Wall-clock latency comparison}
\label{sec:wall_clock}

We primarily focus on evaluating the speedup through our proposed \textit{acceleration factor}, which is derived from the number of SGD steps. In \cref{table:required_time}, we show whether this factor indeed leads to an acceleration of neural field training. To demonstrate this point empirically, we conduct a comparison between the average wall-clock runtimes of two key stages: data pre-processing and loading (referred to as “Load”) and the actual SGD iterations (referred to as “Train”) on original and {\rpp} Kodak images. We find that the extra computation spent during the “Load” is much smaller in scale than the speedup from “Train”, resulting in a net speedup. This corresponds to a time savings of 54543 ms, a whopping 25\% time reduction, in the SIREN architecture, where SGD iterations make up the majority of the training.

\renewcommand{\arraystretch}{1.2}
\begin{table}[h]
\centering
\resizebox{0.48\textwidth}{!}{%
\begin{tabular}{@{}lrrrr}
\hline
         & \multicolumn{2}{c}{SIREN} & \multicolumn{2}{c}{Instant-NGP} \\ \cline{2-5} 
         & Load (ms) & Train (ms) & Load (ms) & Train (ms) \\ \hline
Original & 2,913        & 212,170      & 1,133           & 8,484           \\
\rpp      & 2,933        & 157,607      & 1,324           & 7,813           \\
 & {\color{red}\small(+20)} & {\color{blue}\small(-54,563)} & {\color{red}\small(+191)} & {\color{blue}\small(-671)}\\ \hline
Speedup & \multicolumn{2}{r}{54,543} & \multicolumn{2}{r}{480}         \\ \hline
\end{tabular}
}

\caption{\textbf{Wall-clock speedup (ms) on Kodak dataset.} \\
\textit{\underline{Note.}} The net speedup may depend on the choice of hardware; we measure the wall-clock speed on GPU server with a NVIDIA GeForce RTX 3090, and AMD EPYC 7313 16-Core CPU. The speedup can be even larger in CPU-only setups where training is much slower (e.g., mobile devices).}
\label{table:required_time}
\end{table}
\newpage

\section{Average number of steps for achieving target PSNR}
\label{sec:sup_average}

In \cref{table:big_table_sup}, we present the average number of steps required to achieve a 50dB Peak Signal-to-Noise Ratio (PSNR). We observe that this table does not exactly reflect the trends observed in \cref{table:big_table} of the main paper. This discrepancy is due to the presence of \textit{outliers}; some images that require much bigger number of steps than other images can dominate the overall statistic. For example, for Instant-NGP, the original Kodak\#20 image requires 1279 number of steps until convergence, which is 7.96 times greater than other images on average. In such case, the average depends heavily on this sample; without Kodak\#20, the average for original is 160.7 steps and for {\rpp} is 111.0 steps. The acceleration factor, which we used for the main table, is relatively robust against this issue.

\noindent\textbf{Note.} For some data augmentations, several augmented images could not have been fit with Instant-NGP to PSNR 50dB, with any of the learning rates that we tried. In particular, Gamma correction with $\gamma = 2.0$ cannot fit $(1,2,1)$ images on Kodak, DIV2K, and CLIC datasets, respectively. Likewise, Gamma correction with $\gamma = 0.5$ and Linear Scaling with $t= 2.0$ cannot fit $(0,6,2)$ and $(1,2,1)$ images in each datasets, respectively. For these cases, we report the average number of steps \textit{except} on unconverged cases. Nevertheless, the average number of steps for PSNR 50dB on other data points for these augmentations are typically very large; thus it is very unlikely that these modified averages will lead to a misleading conclusion that such augmentations are beneficial for the training speed.

\newpage
\section{Extended discussions on the potentials and limitations of data transformations}
\label{sec:comprehensive_review}

While {\rpp} provides consistent speedups for fitting the given datum, it has limited applicability to scenarios that require the trained model to have certain characteristics, such as interpolatability (as briefly discussed \cref{ssec:examplecases}). This section provides a more comprehensive discussion on this matter, to elucidate both the potential benefits and limitations.


\noindent\textbf{Image fitting} is to represent the target 2D image with a corresponding neural field \citep{siren, acorn}. Here, the main challenge is to \textit{overfit} to a singular data instance with high fidelity, and our framework is well-suited for this application. By applying the same permutation matrix across all images, {\rpp} can accelerate training without additional memory overhead (\cref{sec:single_rpp}).


\noindent\textbf{Image superresolution} \citep{liif, jiif, super_hsi} and \textbf{image inpainting} \citep{inpainting_inr, wire} utilizes neural field trained on a set of seen coordinates to predict the signal value on unseen coordinates. These applications strongly rely on the ability of neural field to capture the spatial patterns of the target signal from the seen coordinates and interpolate them.
However, {\rpp} introduces a significant challenge in learning the meaningful implicit information in terms of spatial relationships. {\rpp} can disrupt the neural field’s understanding of locality and continuity, which is vital for both superresolution and inpainting. Other transformations that adjust only pixel intensities (\eg, standardization), retain the local structure of data, and can thus be used.

\noindent\textbf{Data compression} \citep{coin, coin++,coolchic} reduces the required number of bits to store the data by representing the datum by a neural field with a small number of parameters. This idea, however, entails substantial computational resources for each datum during the encoding process, where we need to train a neural field. {\rpp} is potentially very useful for reducing the encoding time, by accelerating the neural field training.

\noindent\textbf{3D scene reconstruction and novel view synthesis} \citep{nerf,instantngp} utilizes a neural field trained on a small number of 2D images of a same 3D scene to generate a novel view of the scene. For this, we heavily rely on the capability of the neural field to infer implicit relationships from all input data. Data transformations involving pixel relocations present a significant hurdle in this context, complicating or outright precluding the networks to predict the adequate relationship.
Several concurrent works \citep{partition, divide&conquer} have explored a training paradigm which trains a single input with segmenting based on the inherent characteristics such as frequencies. Our hypothesis, “blessing of no pattern”, can provide a novel insight within this discourse. We demonstrate that there exist representative patterns in data, and they work completely differently depending on the learning phase. From this point of view, the representative patterns can serve as a new criterion for partitioning an instance.

\noindent\textbf{Inference with neural field weights} \citep{functa, inr2vec} regards the set of weight parameters derived from neural fields as a conventional input feature, based on which we perform inference. Data transformations can help constructing a large-scale dataset of neural field weights, which can be used to train a model that predicts based on the weight vectors.

\section{Utilizing a single permutation matrix for RPP}
\label{sec:single_rpp}

In the main paper, we used an independently drawn permutation matrix for each {\rpp} image. Despite the interesting results, this hinders applying {\rpp} to further applications due to its complete randomness and memory footprint of the matrix.

\renewcommand{\arraystretch}{1.2}
\begin{table}[h]
\centering
\resizebox{0.48\textwidth}{!}{%
\begin{tabular}{@{}lrr}
\hline
            & Independently drawn & Same permutation \\ \cline{1-3} 
SIREN       & 1.26$\times$  & 1.27$\times$          \\
Instant-NGP & 1.50$\times$  & 1.51$\times$      \\ \hline
\end{tabular}
}

\caption{Using independently drawn vs. same permutation.}
\label{table:srpp}
\vspace{-1em}
\end{table}

\noindent We find that {\rpp} can accelerate the neural field training even with \textit{a single unified permutation matrix}. In \cref{table:srpp}, we compare the acceleration factors of {\rpp} using the independently drawn random permutation (per image) and using the same permutation, on the Kodak dataset. We observe that the quantities are roughly identical to each other both in two architectures. 

\newpage
\section{DCT coefficient of other datasets}
\label{sec:sup_dct}

\begin{figure}[t]
\centering
    \begin{subfigure}[]{0.49\linewidth}
        \centering
        \includegraphics[width=\linewidth]{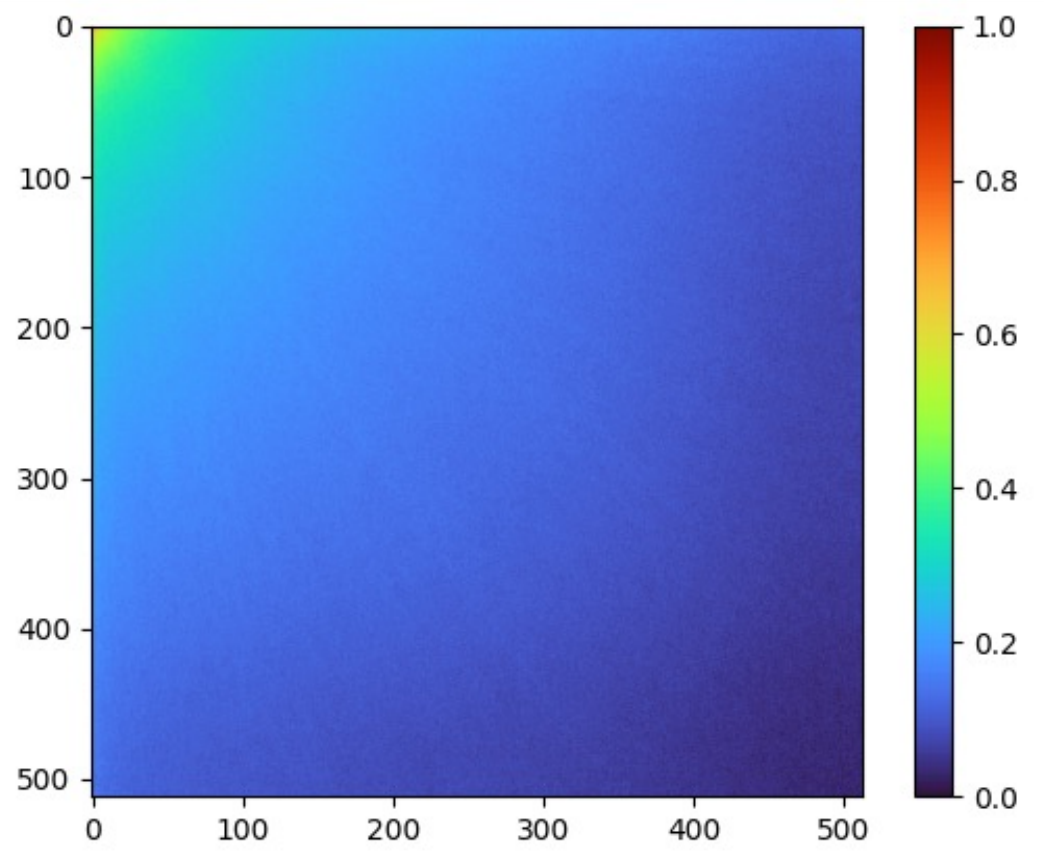}
        \caption{Original}
        \label{fig:dct_orig_div2k}
    \end{subfigure}
    \hfill
    \begin{subfigure}[]{0.49\linewidth}
        \centering
        \includegraphics[width=\linewidth]{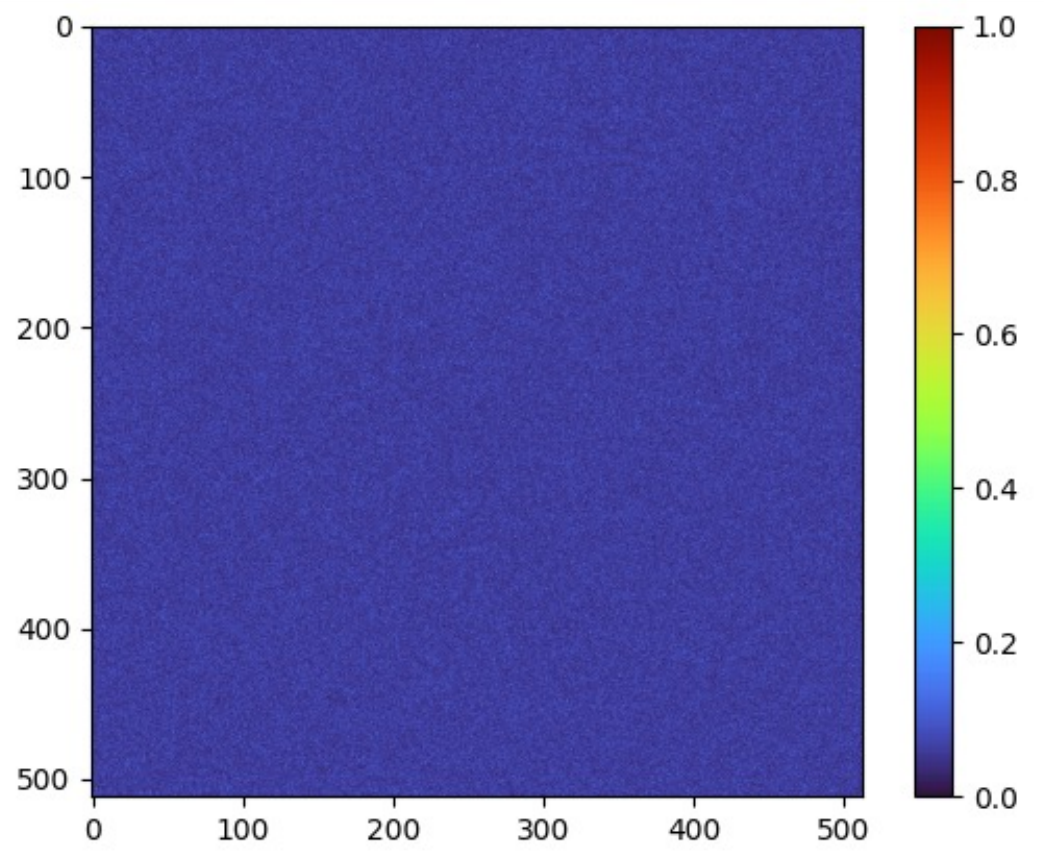}
        \caption{Permutation}
        \label{fig:dct_per_div2k}
    \end{subfigure}

    \caption{\textbf{Frequency spectra of original vs. {\rpp} (DIV2K).} We compare the average DCT coefficients of the original and {\rpp} DIV2K images. Upper left region denotes the low-frequency, and the lower right region denotes the high-frequency.}
    \label{fig:vis_dct_div2k}
\end{figure}

\begin{figure}[t]
\centering
    \begin{subfigure}[]{0.49\linewidth}
        \centering
        \includegraphics[width=\linewidth]{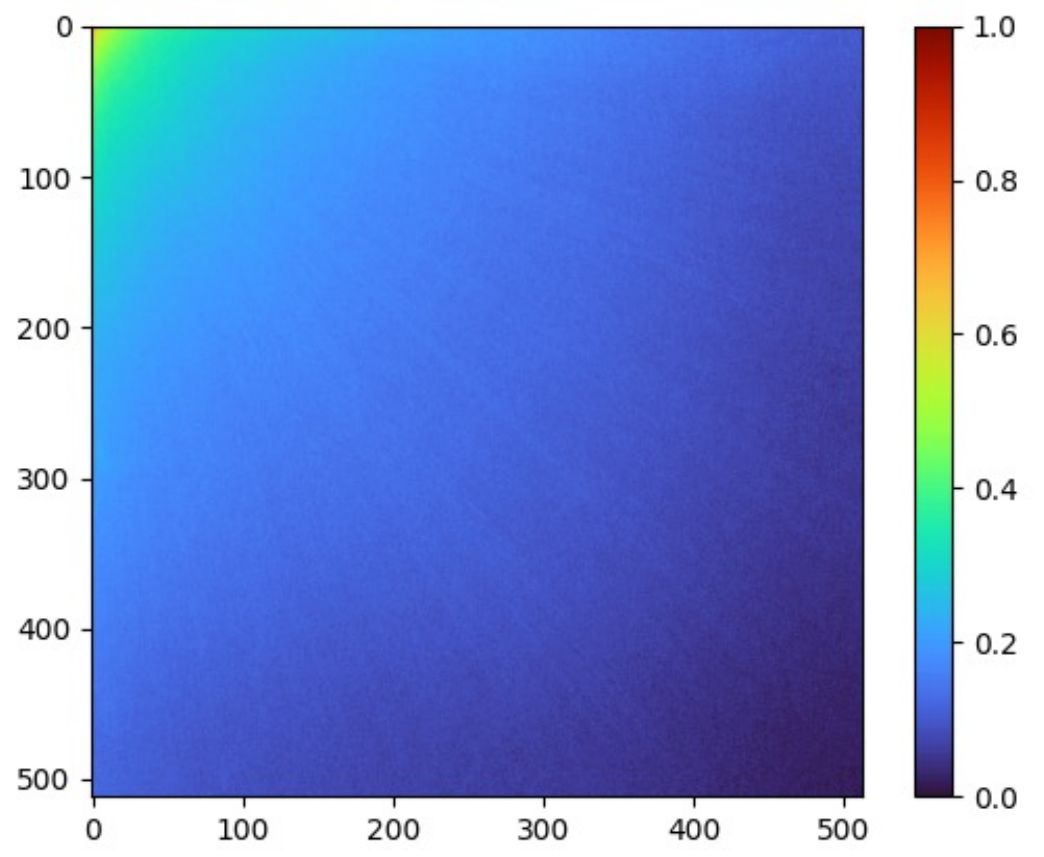}
        \caption{Original}
        \label{fig:dct_orig_clic}
    \end{subfigure}
    \hfill
    \begin{subfigure}[]{0.49\linewidth}
        \centering
        \includegraphics[width=\linewidth]{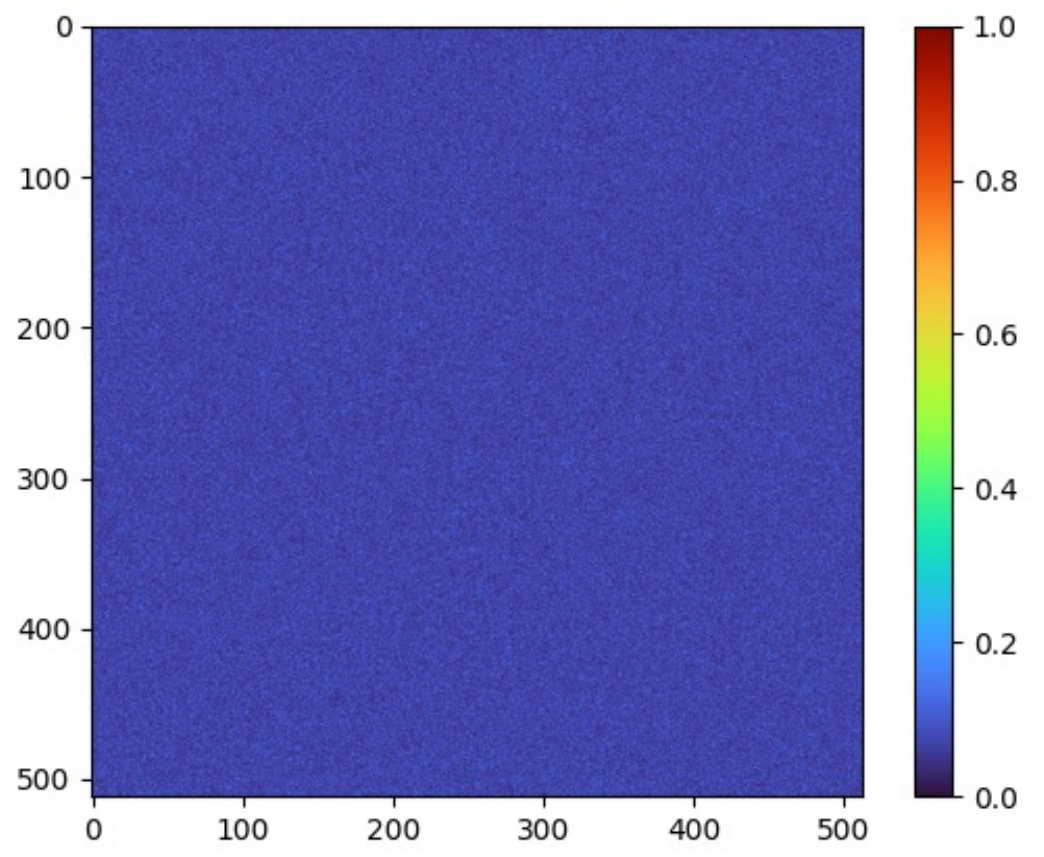}
        \caption{Permutation}
        \label{fig:dct_per_clic}
    \end{subfigure}

    \caption{\textbf{Frequency spectra of original vs. {\rpp} (CLIC).} We compare the average DCT coefficients of the original and {\rpp} CLIC images.}
    \label{fig:vis_dct_clic}
\end{figure}

In \cref{fig:vis_dct_kodak} of the main text, we have compared the average Discrete Cosine Transform (DCT) coefficients of original images and {\rpp} images on the Kodak dataset. \Cref{fig:vis_dct_kodak} demonstrates a discernible pattern: in the original images, we observe that the upper left region (low-frequency components) has much higher coefficients than other regions, whereas in {\rpp} images the scale of the coefficients are relatively uniform. We extend our analysis to other datasets, DIV2K and CLIC, to validate the consistency of our initial findings. The results from these datasets (\cref{fig:vis_dct_div2k,fig:vis_dct_clic}, respectively) mirror the observations on the Kodak images. In the original images from DIV2K and CLIC, we once again observed a prevalence of low-frequency components, particularly in the upper left region of the frequency spectra. The distribution is uniform on {\rpp} images.

\newpage

\section{Full PSNR curves on the Kodak dataset}
\label{sec:sup_lc_curve}

\begin{figure*}[h]
\centering
    \begin{subfigure}[t]{0.32\linewidth}
        \centering
        \fcolorbox{red}{white}{\includegraphics[width=\linewidth]{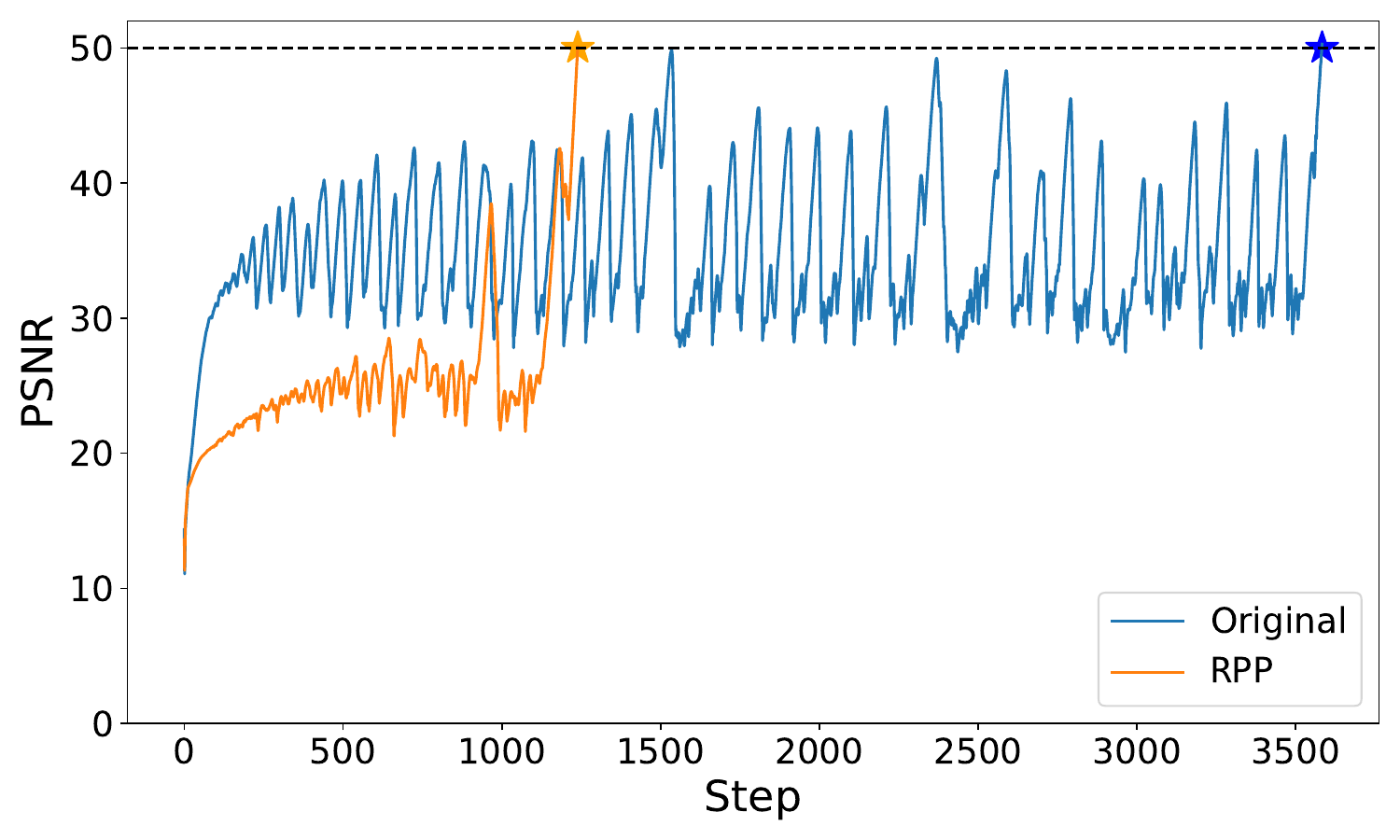}}
        \captionsetup{skip=0pt}
        \caption{Kodak\#01}
        \label{fig:lc_kodak1}
    \end{subfigure}
    \hfill
    \begin{subfigure}[t]{0.32\linewidth}
        \centering
        \includegraphics[width=\linewidth]{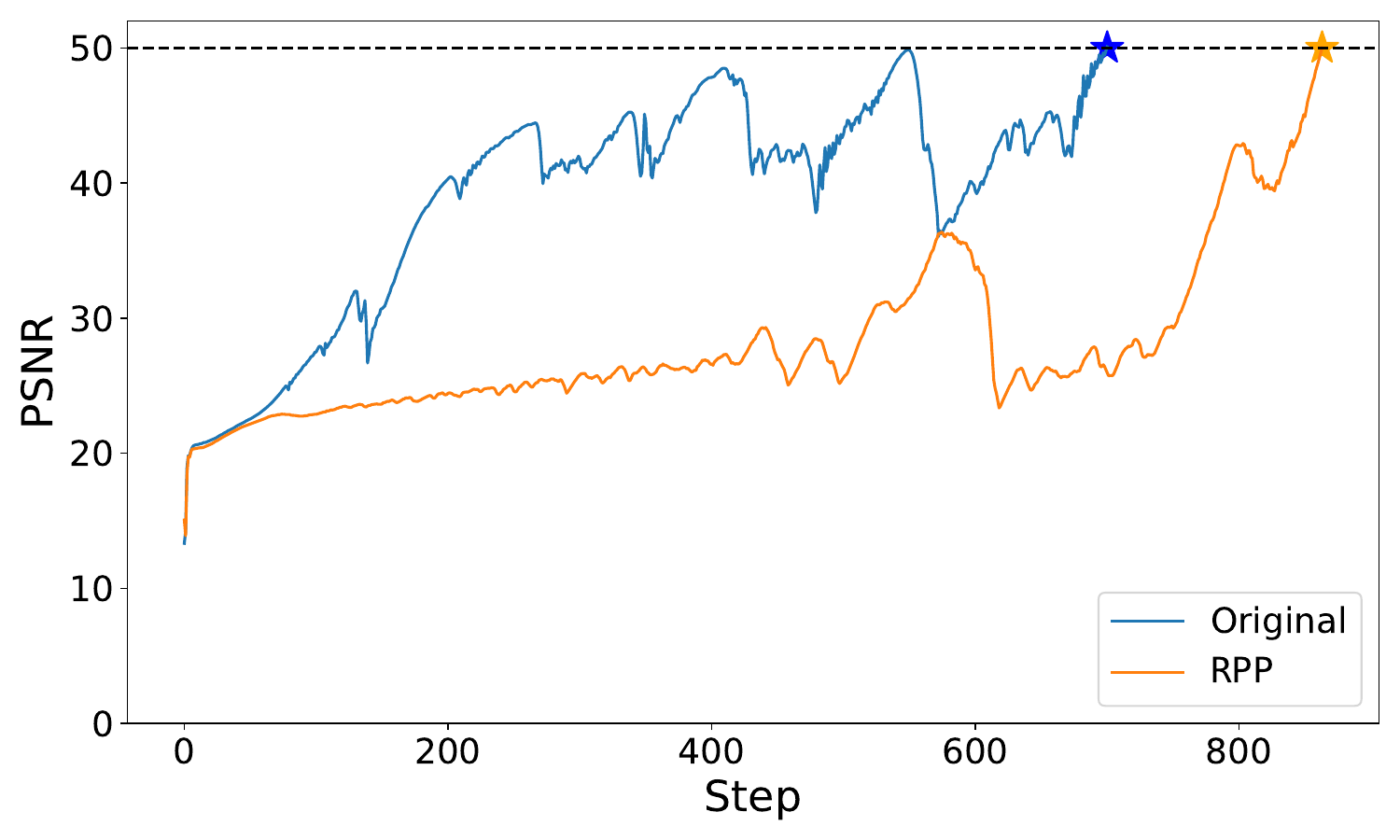}
        \captionsetup{skip=0pt}
        \caption{Kodak\#02}
        \label{fig:lc_kodak2}
    \end{subfigure}
    \hfill
    \begin{subfigure}[t]{0.32\linewidth}
        \centering
        \includegraphics[width=\linewidth]{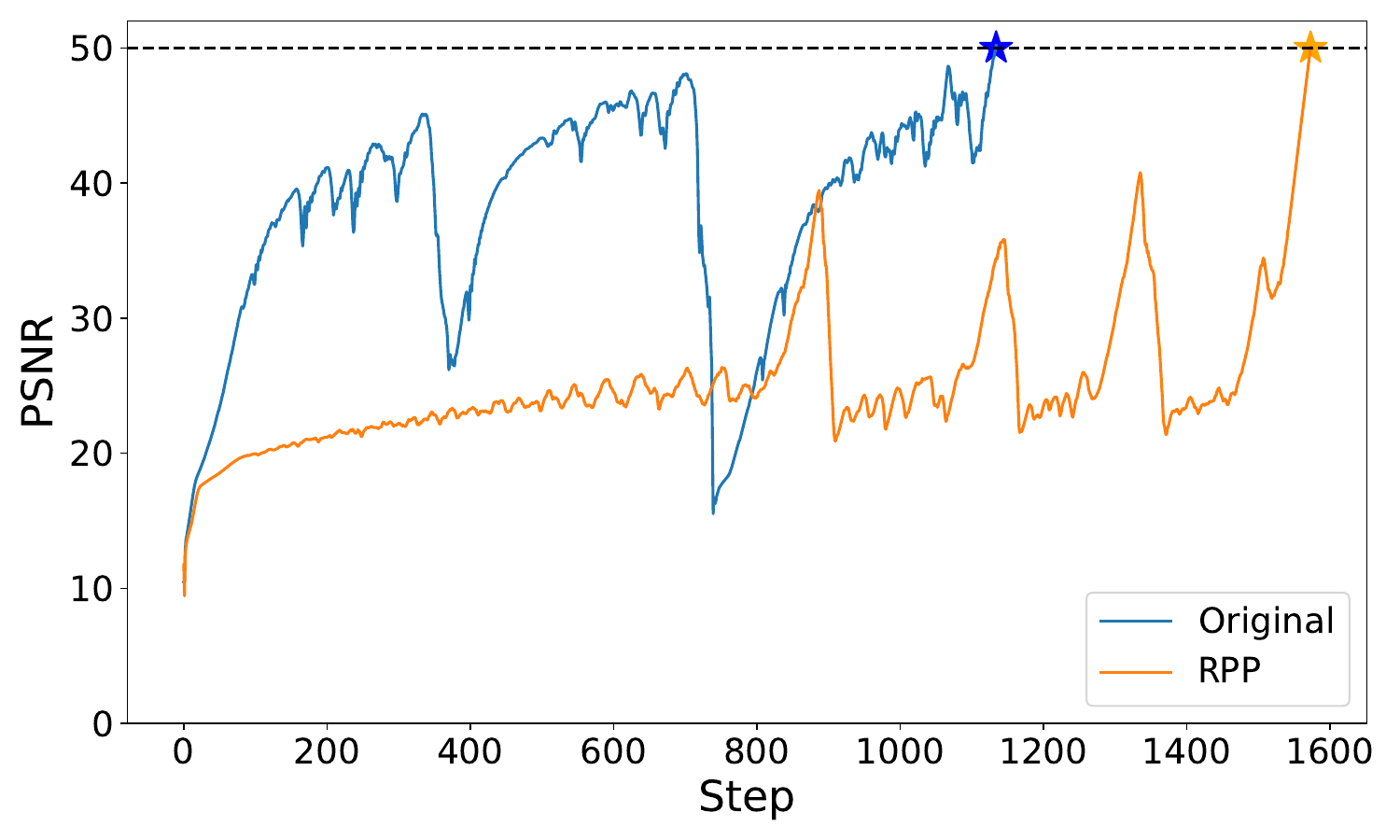}
        \captionsetup{skip=0pt}
        \caption{Kodak\#03}
        \label{fig:lc_kodak3}
    \end{subfigure}
    
    \vspace{0.1cm}
    \begin{subfigure}[t]{0.32\linewidth}
        \centering
        \includegraphics[width=\linewidth]{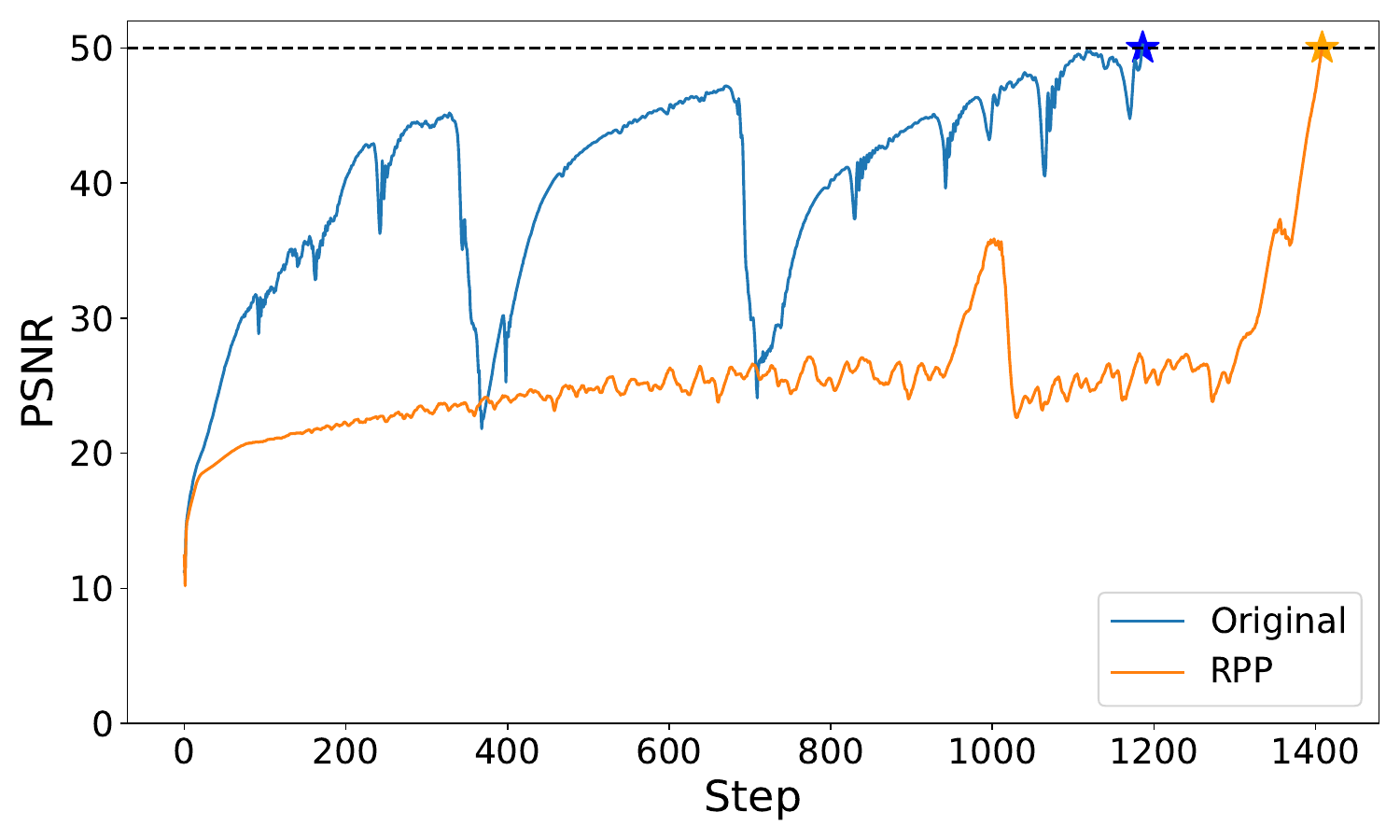}
        \captionsetup{skip=0pt}
        \caption{Kodak\#04}
        \label{fig:lc_kodak4}
    \end{subfigure}
    \hfill
    \begin{subfigure}[t]{0.32\linewidth}
        \centering
        \fcolorbox{red}{white}{\includegraphics[width=\linewidth]{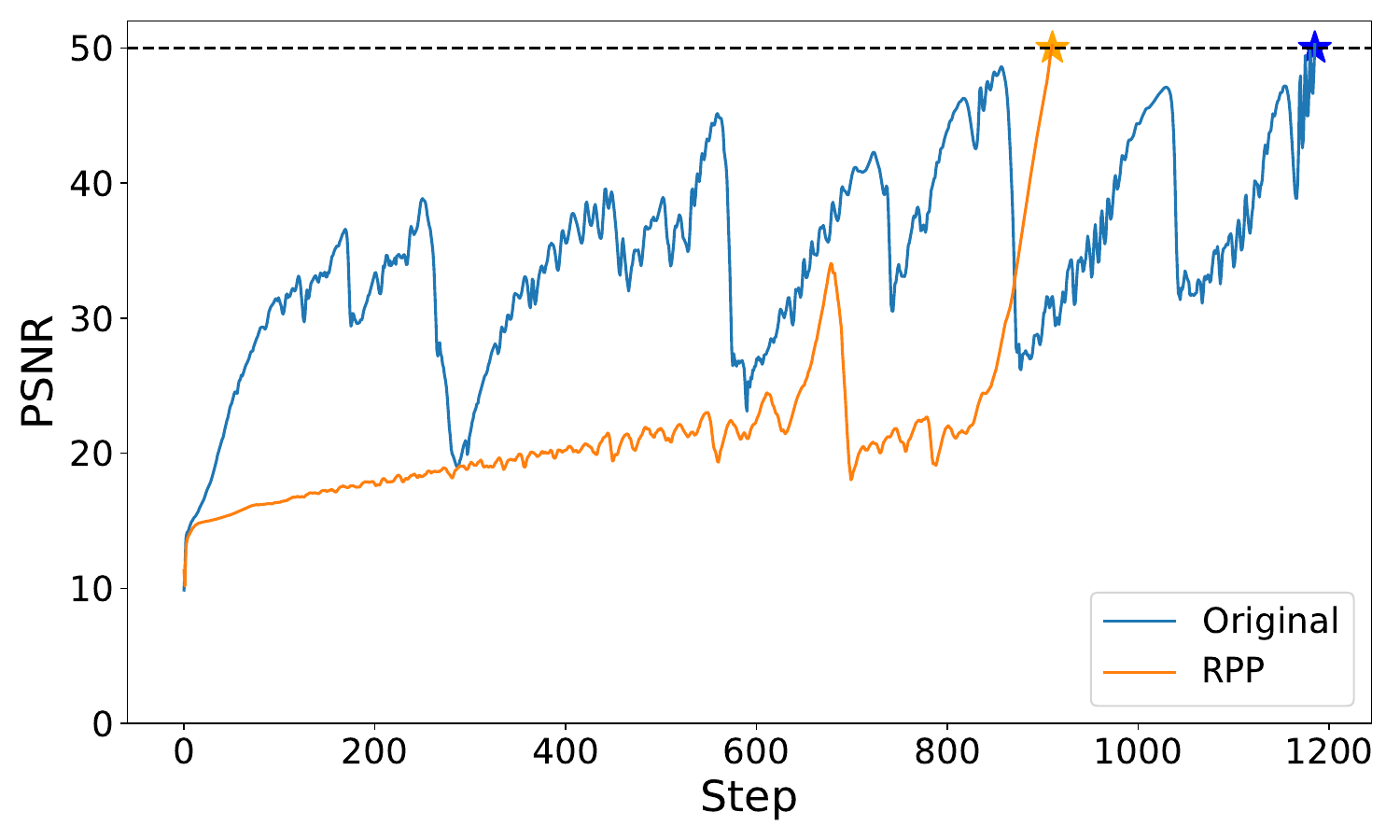}}
        \captionsetup{skip=0pt}
        \caption{Kodak\#05}
        \label{fig:lc_kodak5}
    \end{subfigure}
    \hfill
    \begin{subfigure}[t]{0.32\linewidth}
        \centering
        \includegraphics[width=\linewidth]{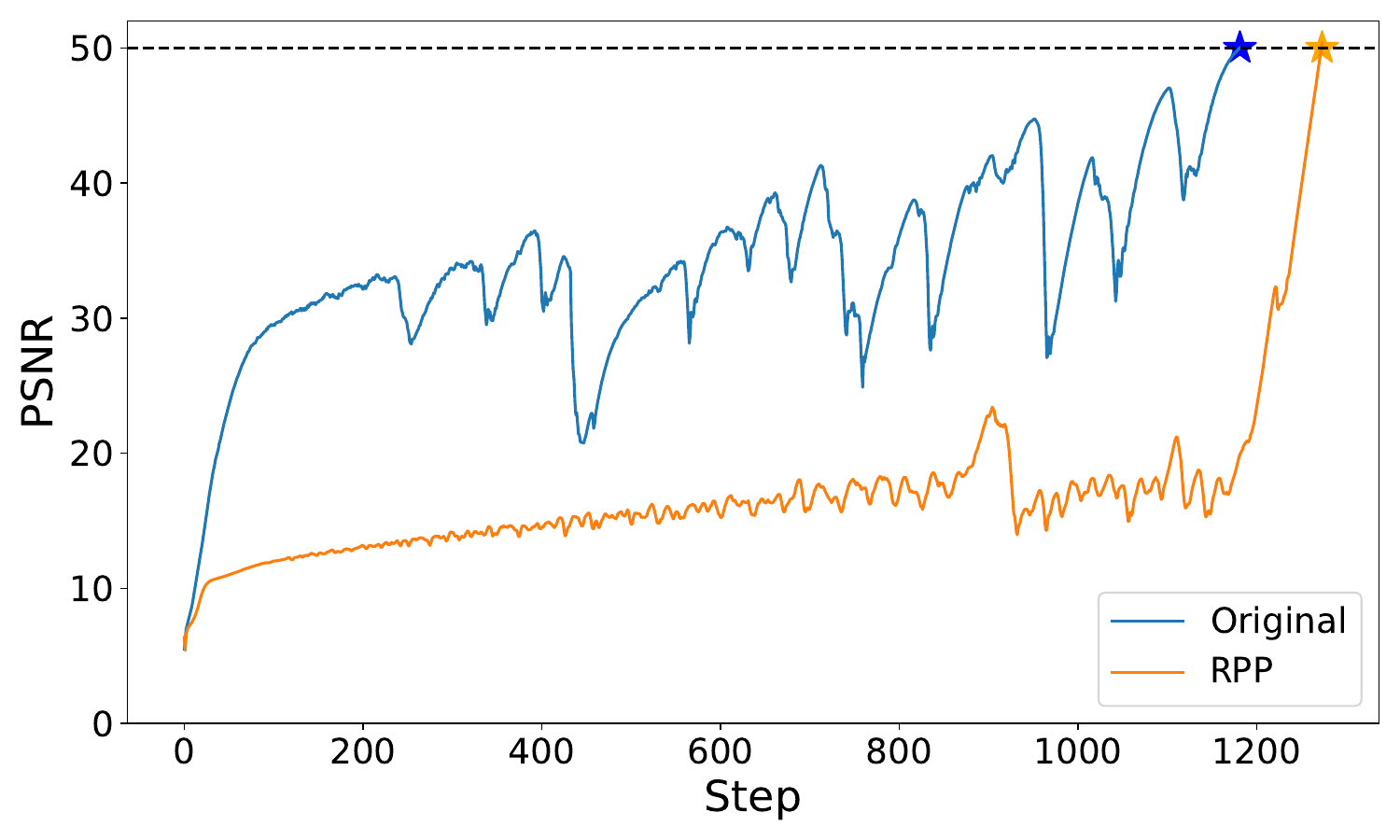}
        \captionsetup{skip=0pt}
        \caption{Kodak\#06}
        \label{fig:lc_kodak6}
    \end{subfigure}

    \vspace{0.1cm}
    \begin{subfigure}[t]{0.32\linewidth}
        \centering
        \includegraphics[width=\linewidth]{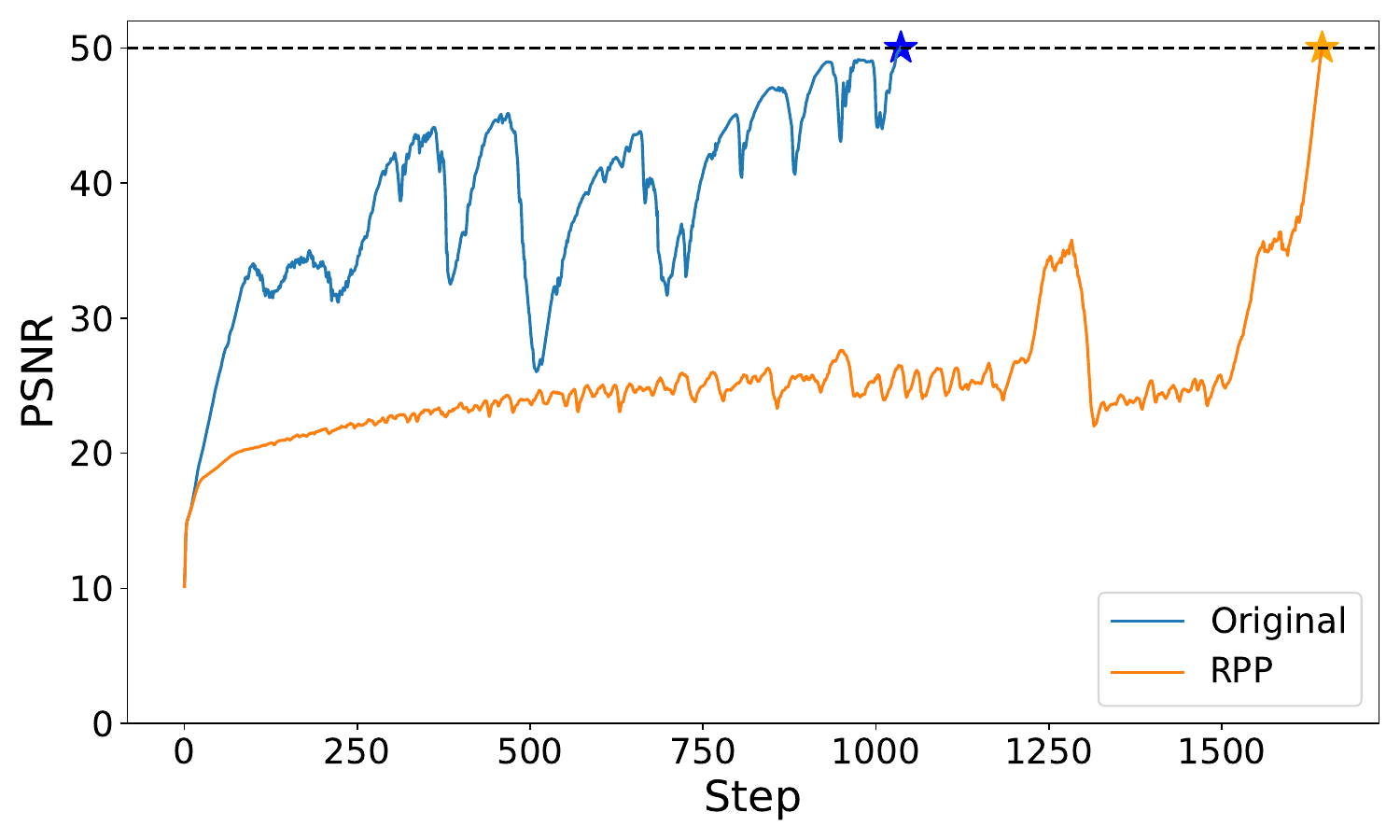}
        \captionsetup{skip=0pt}
        \caption{Kodak\#07}
        \label{fig:lc_kodak7}
    \end{subfigure}
    \hfill
    \begin{subfigure}[t]{0.32\linewidth}
        \centering
        \fcolorbox{red}{white}{\includegraphics[width=\linewidth]{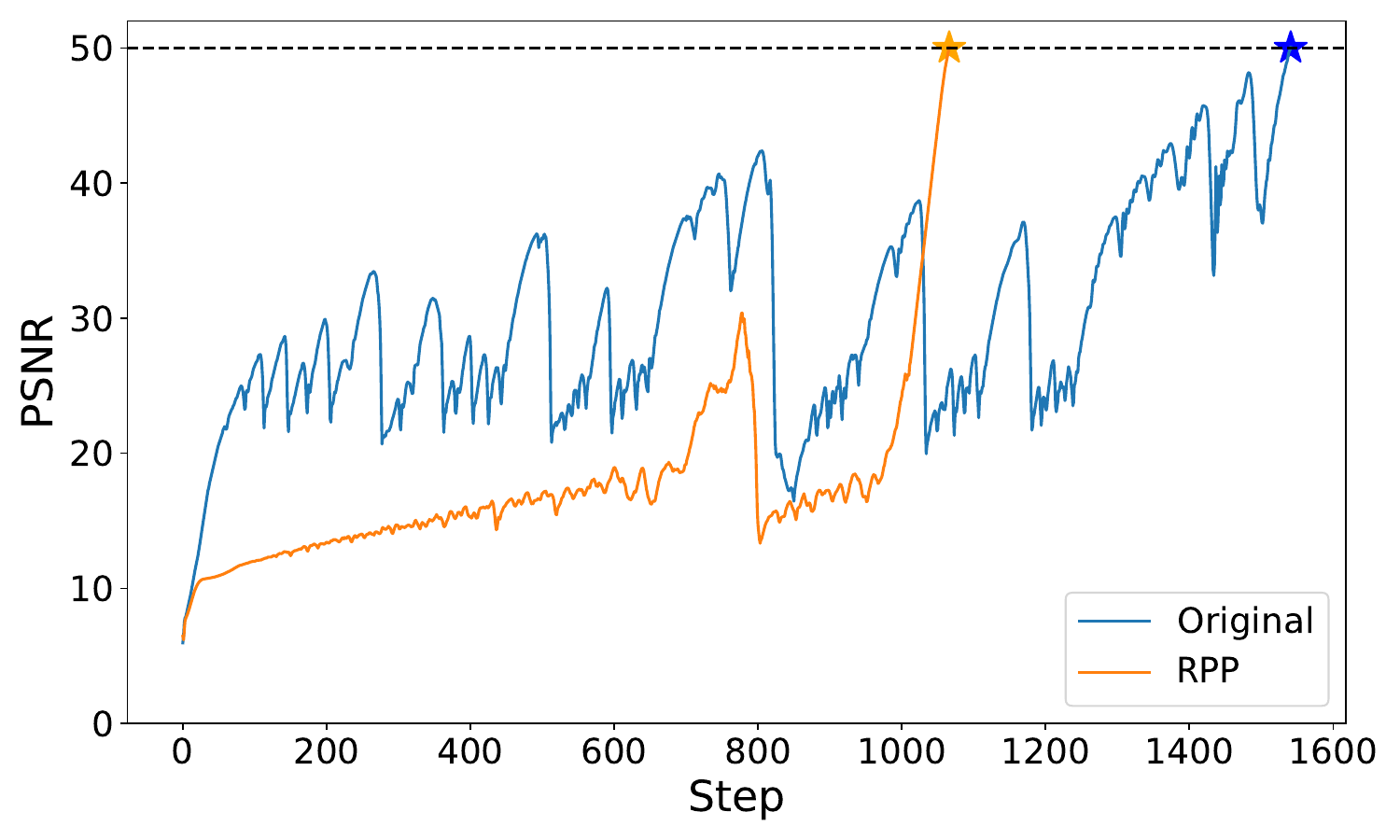}}
        \captionsetup{skip=0pt}
        \caption{Kodak\#08}
        \label{fig:lc_kodak8}
    \end{subfigure}
    \hfill
    \begin{subfigure}[t]{0.32\linewidth}
        \centering
        \includegraphics[width=\linewidth]{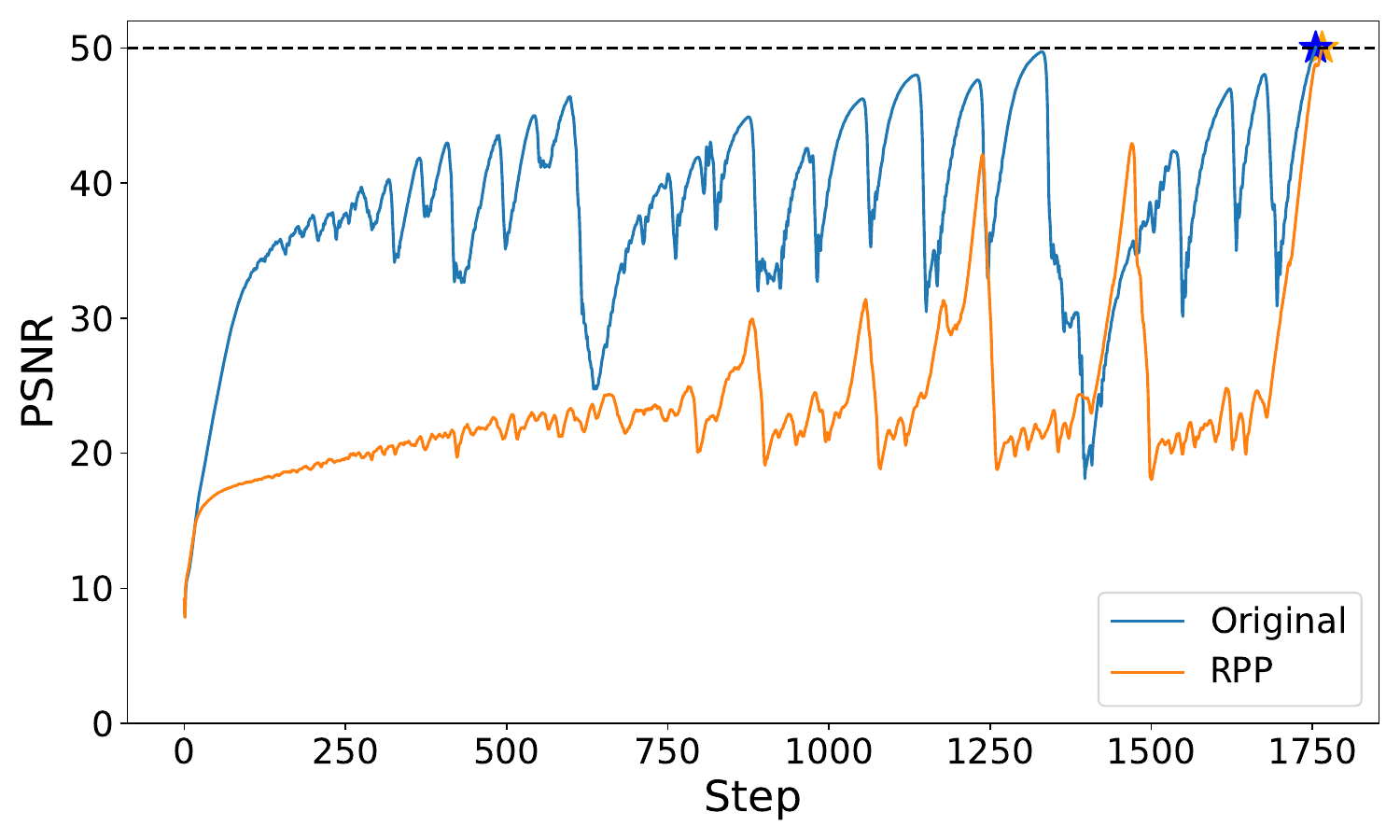}
        \captionsetup{skip=0pt}
        \caption{Kodak\#09}
        \label{fig:lc_kodak9}
    \end{subfigure}
    
    \vspace{0.1cm}
    \begin{subfigure}[t]{0.32\linewidth}
        \centering
        \fcolorbox{red}{white}{\includegraphics[width=\linewidth]{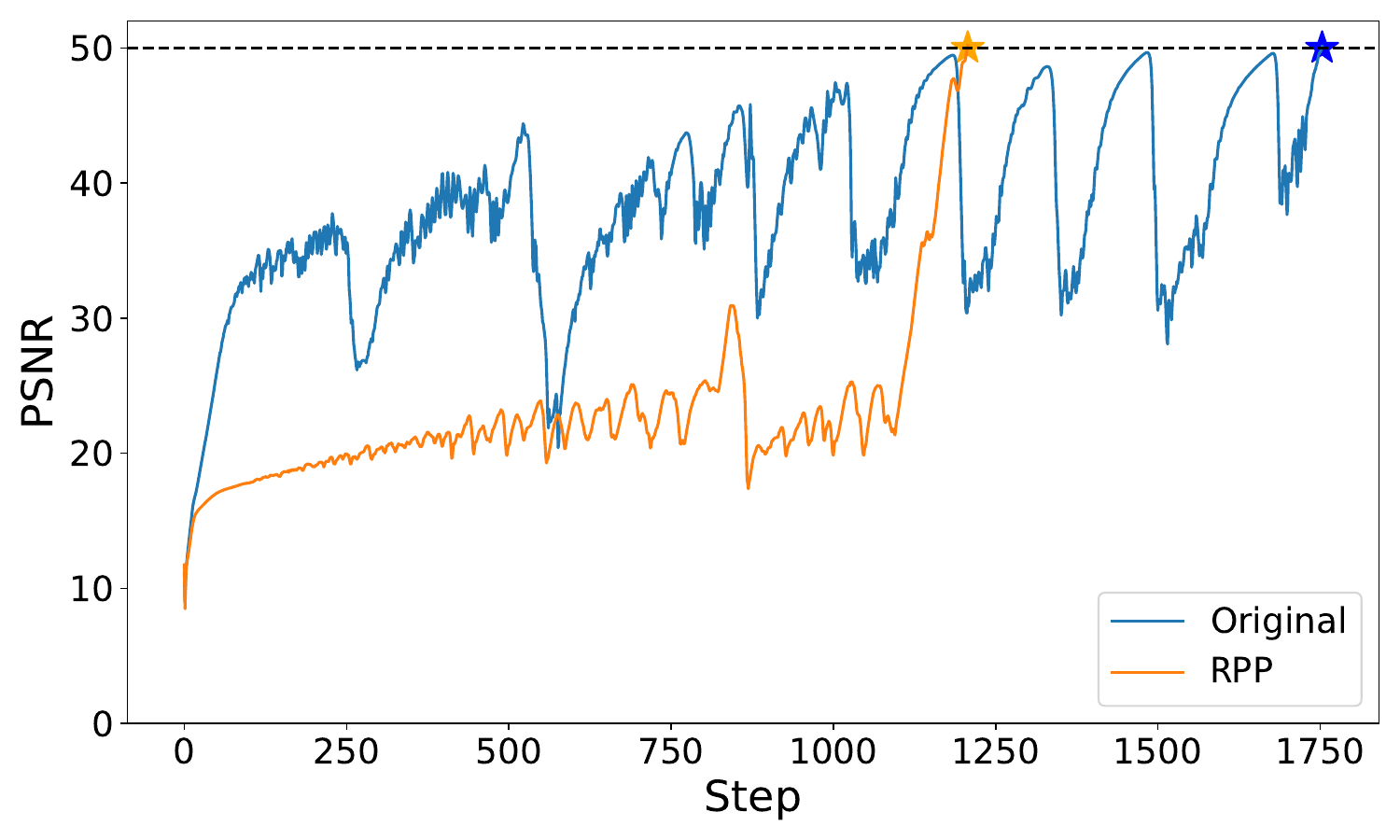}}
        \captionsetup{skip=0pt}
        \caption{Kodak\#10}
        \label{fig:lc_kodak10}
    \end{subfigure}
    \hfill
    \begin{subfigure}[t]{0.32\linewidth}
        \centering
        \fcolorbox{red}{white}{\includegraphics[width=\linewidth]{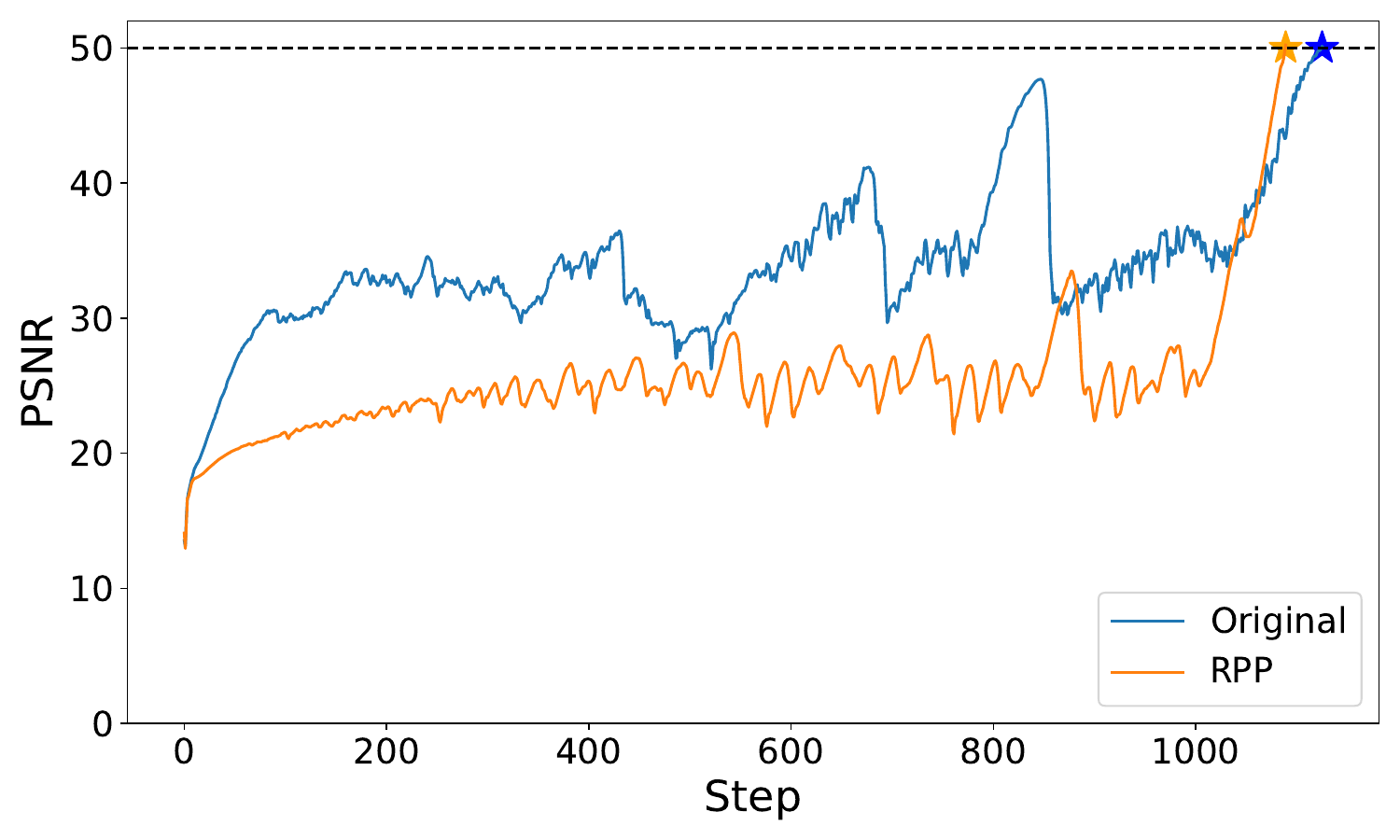}}
        \captionsetup{skip=0pt}
        \caption{Kodak\#11}
        \label{fig:lc_kodak11}
    \end{subfigure}
    \hfill
    \begin{subfigure}[t]{0.32\linewidth}
        \centering
        \fcolorbox{red}{white}{\includegraphics[width=\linewidth]{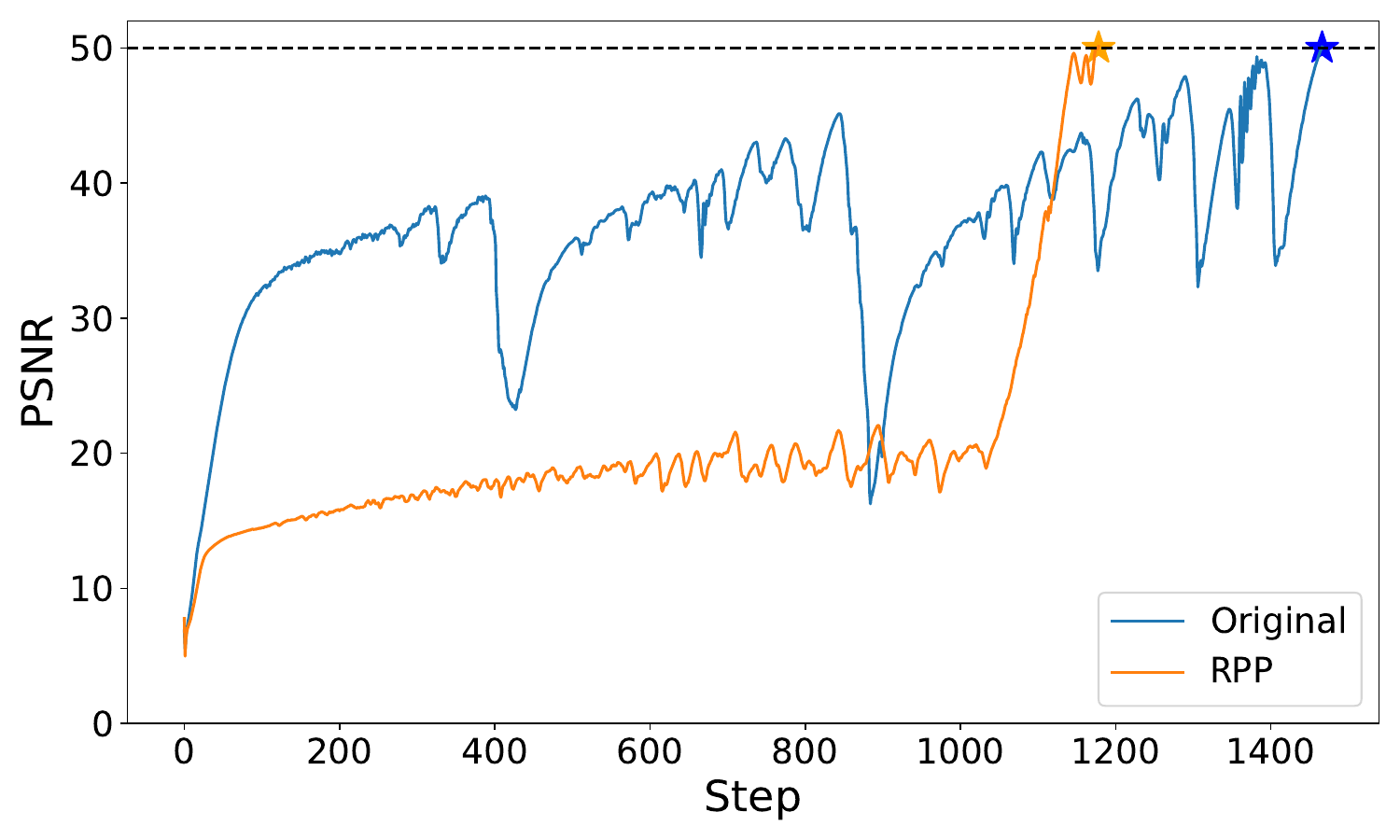}}
        \captionsetup{skip=0pt}
        \caption{Kodak\#12}
        \label{fig:lc_kodak12}
    \end{subfigure}

    \caption{\textbf{PSNR Curves of Kodak images \#01--\#12.} We report the PSNR curves of Kodak images. The {\rpp} images quickly surge to PSNR 50dB.}
    \label{fig:lc_represent_kodak1}
    \vspace{-1em}
\end{figure*}

\begin{figure*}[h]
\centering

    \vspace{0.1cm}
    \begin{subfigure}[t]{0.32\linewidth}
        \centering
        \fcolorbox{red}{white}{\includegraphics[width=\linewidth]{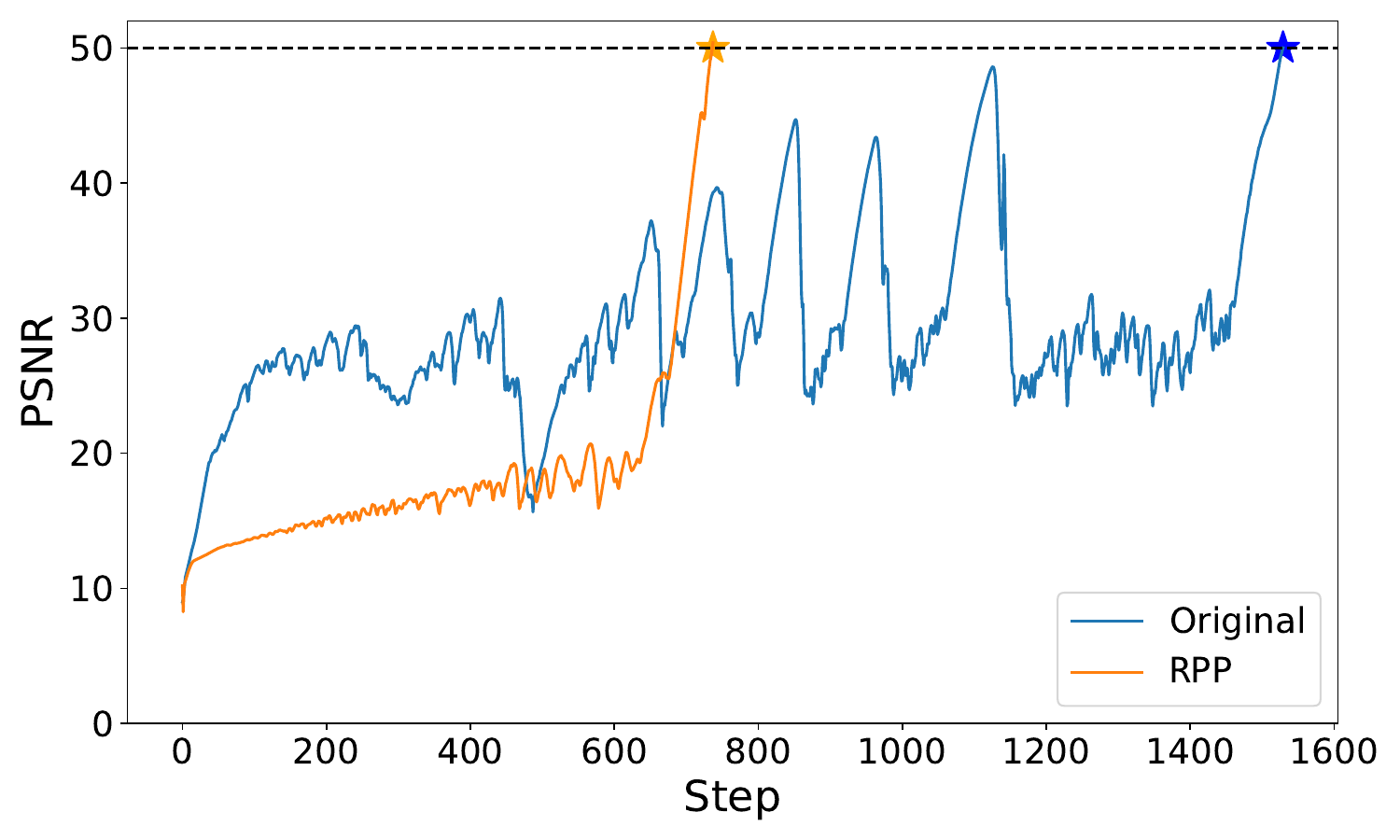}}
        \captionsetup{skip=0pt}
        \caption{Kodak\#13}
        \label{fig:lc_kodak13}
    \end{subfigure}
    \hfill
    \begin{subfigure}[t]{0.32\linewidth}
        \centering
        \fcolorbox{red}{white}{\includegraphics[width=\linewidth]{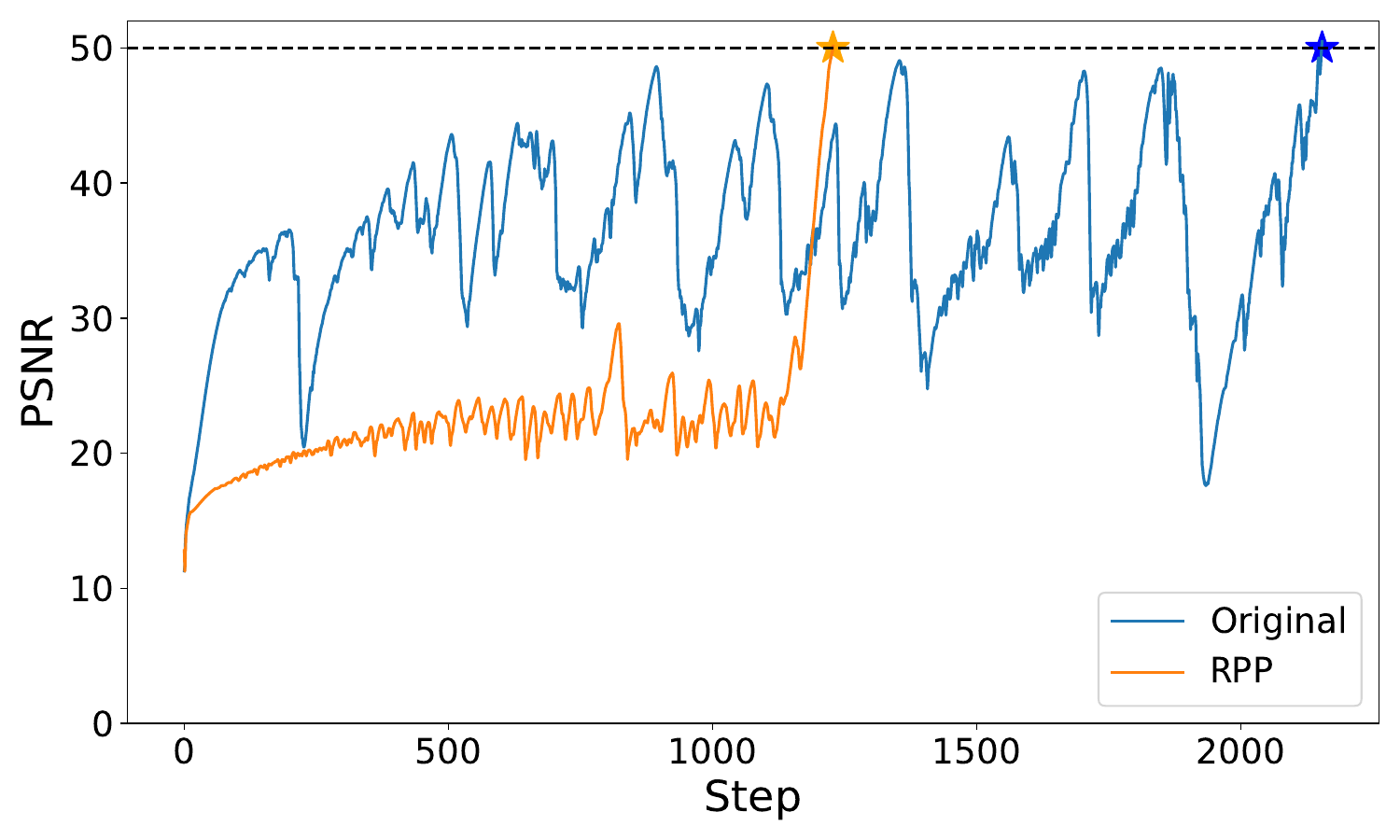}}
        \captionsetup{skip=0pt}
        \caption{Kodak\#14}
        \label{fig:lc_kodak14}
    \end{subfigure}
    \hfill
    \begin{subfigure}[t]{0.32\linewidth}
        \centering
        \fcolorbox{red}{white}{\includegraphics[width=\linewidth]{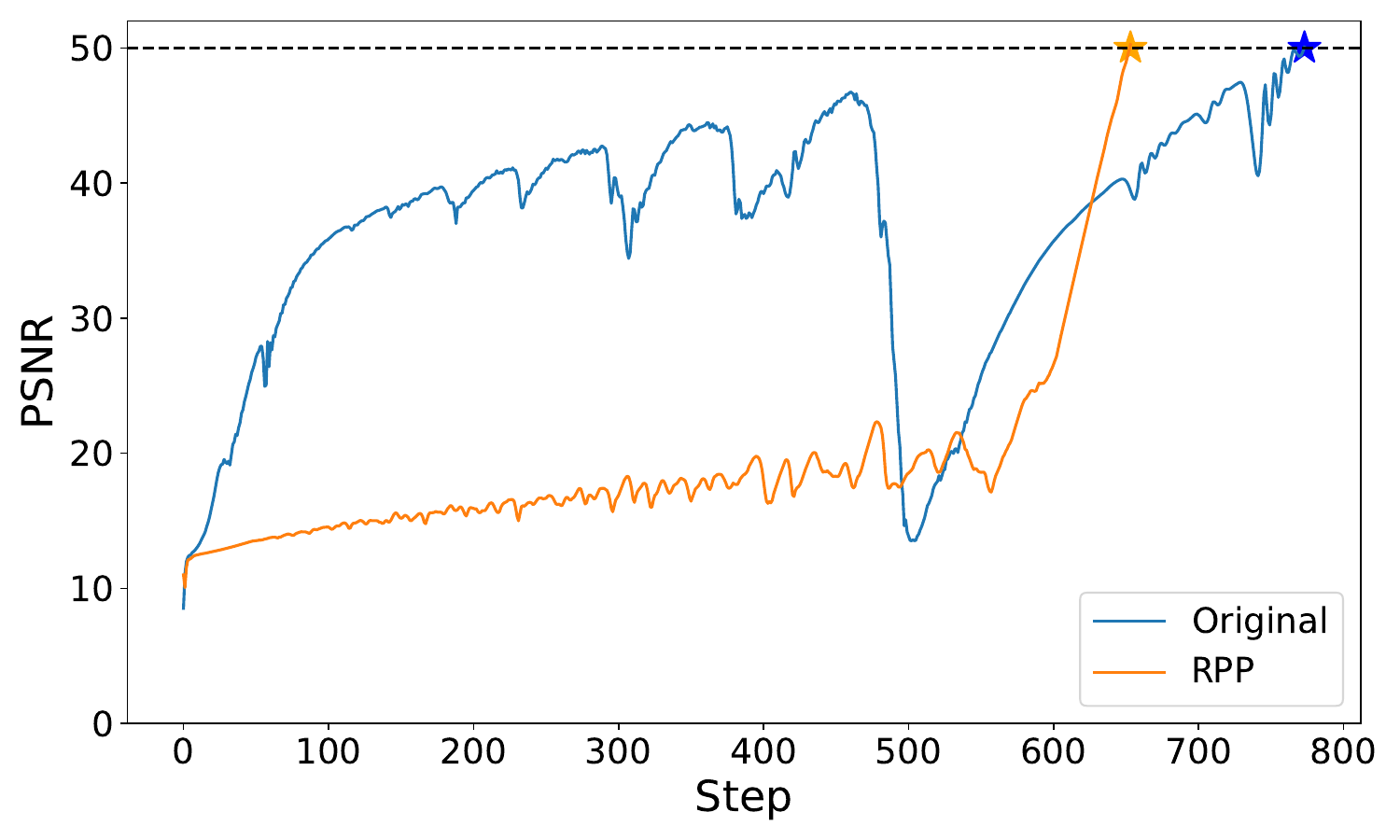}}
        \captionsetup{skip=0pt}
        \caption{Kodak\#15}
        \label{fig:lc_kodak15}
    \end{subfigure}

    \vspace{0.1cm}
    \begin{subfigure}[t]{0.32\linewidth}
        \centering
        \includegraphics[width=\linewidth]{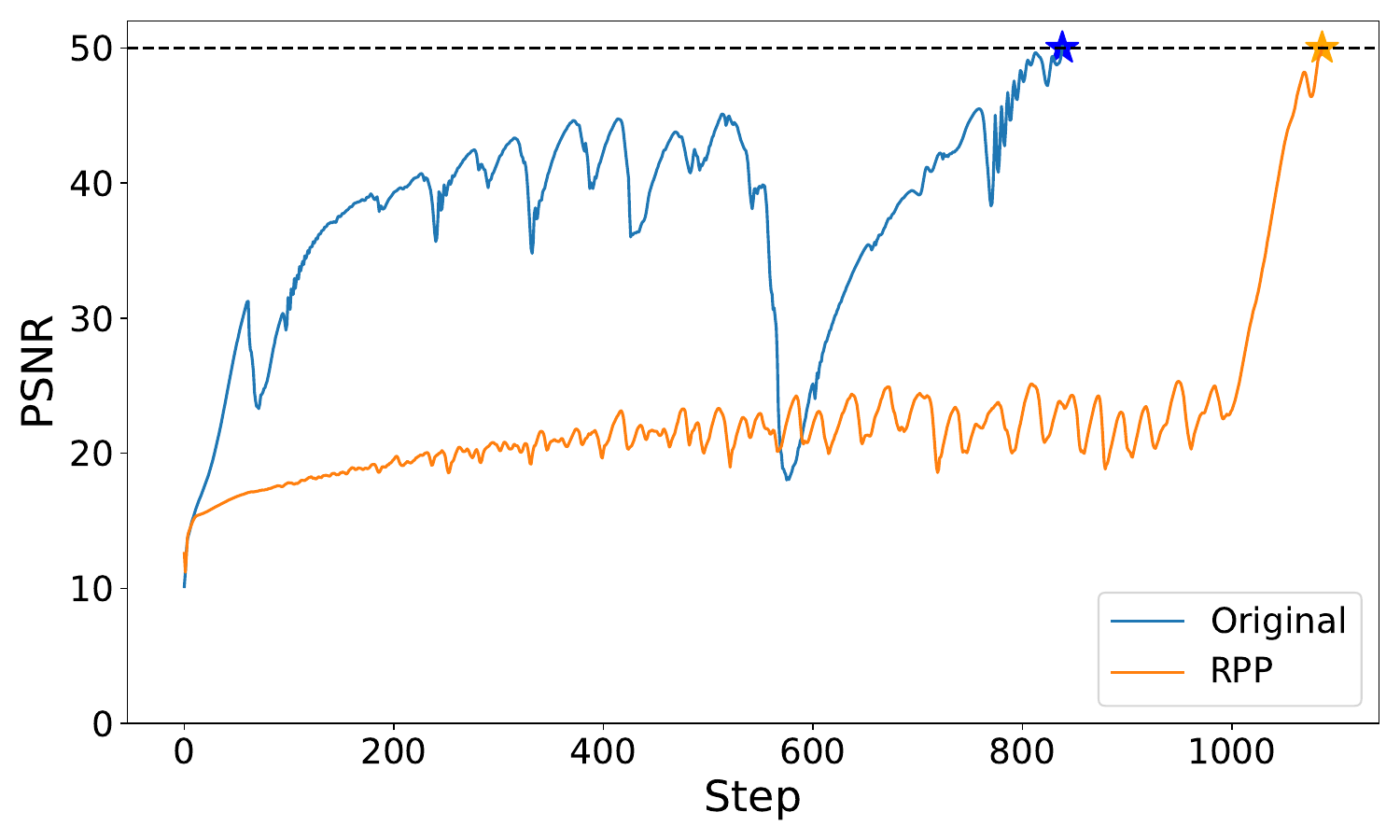}
        \captionsetup{skip=0pt}
        \caption{Kodak\#16}
        \label{fig:lc_kodak16}
    \end{subfigure}
    \hfill
    \begin{subfigure}[t]{0.32\linewidth}
        \centering
        \fcolorbox{red}{white}{\includegraphics[width=\linewidth]{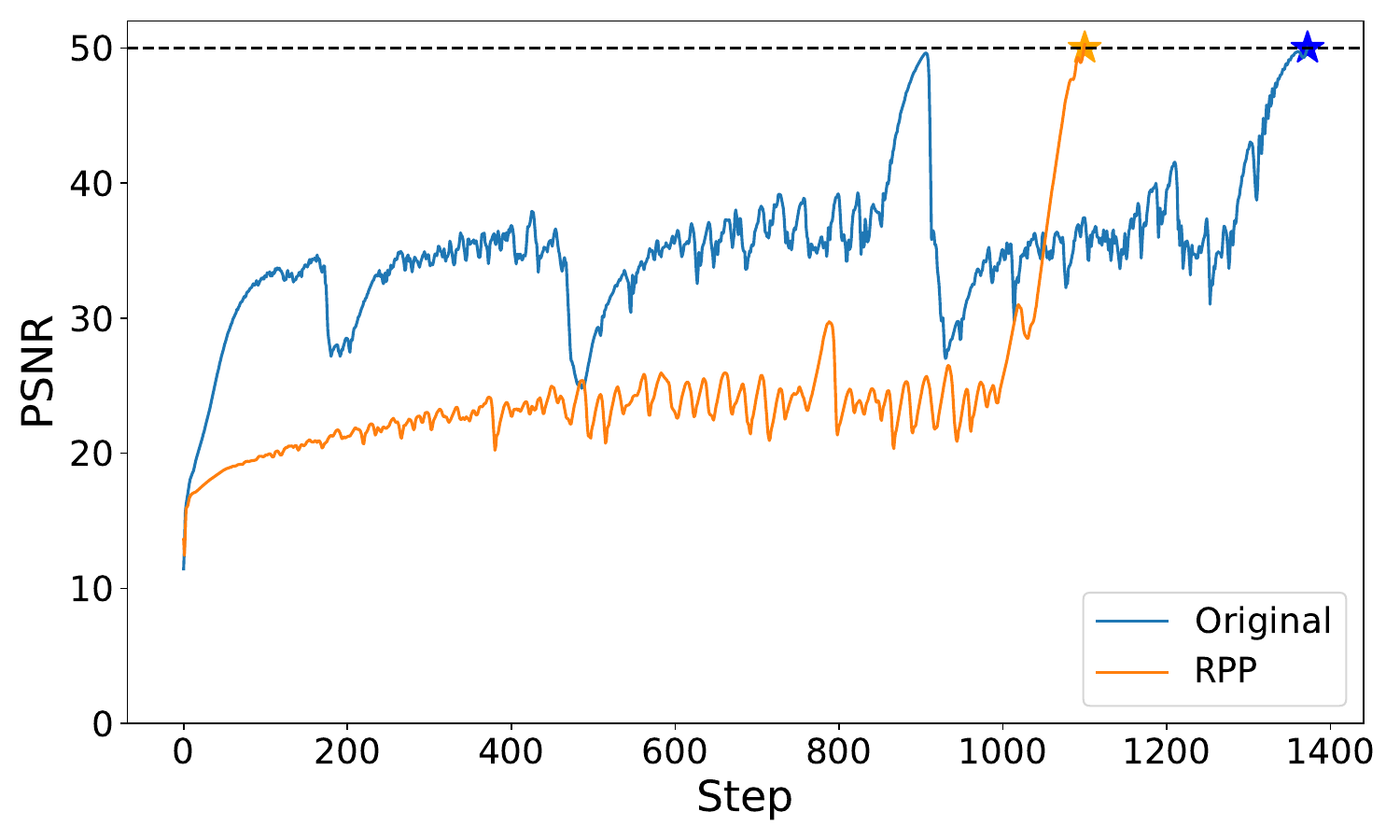}}
        \captionsetup{skip=0pt}
        \caption{Kodak\#17}
        \label{fig:lc_kodak17}
    \end{subfigure}
    \hfill
    \begin{subfigure}[t]{0.32\linewidth}
        \centering
        \fcolorbox{red}{white}{\includegraphics[width=\linewidth]{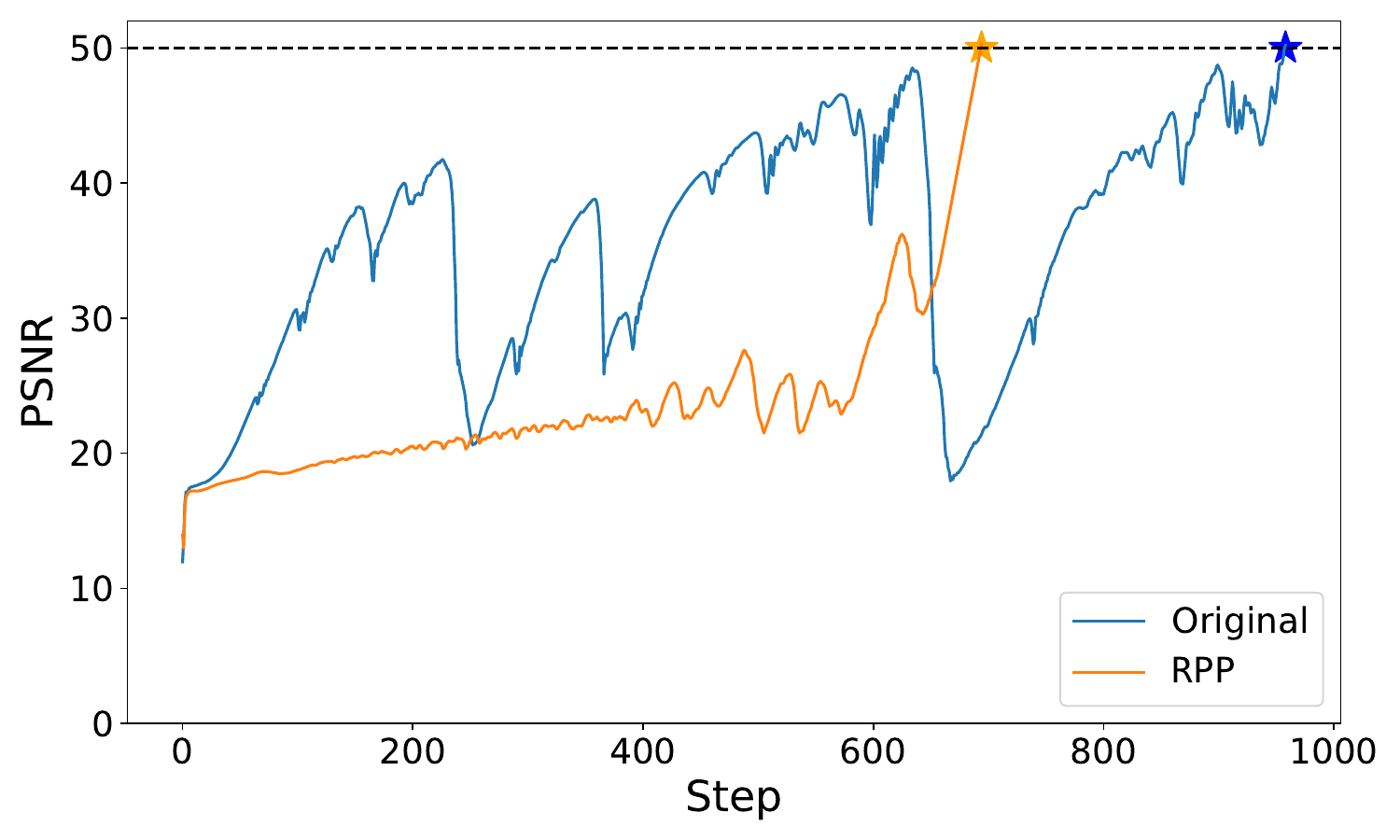}}
        \captionsetup{skip=0pt}
        \caption{Kodak\#18}
        \label{fig:lc_kodak18}
    \end{subfigure}

    \vspace{0.1cm}
    \begin{subfigure}[t]{0.32\linewidth}
        \centering
        \fcolorbox{red}{white}{\includegraphics[width=\linewidth]{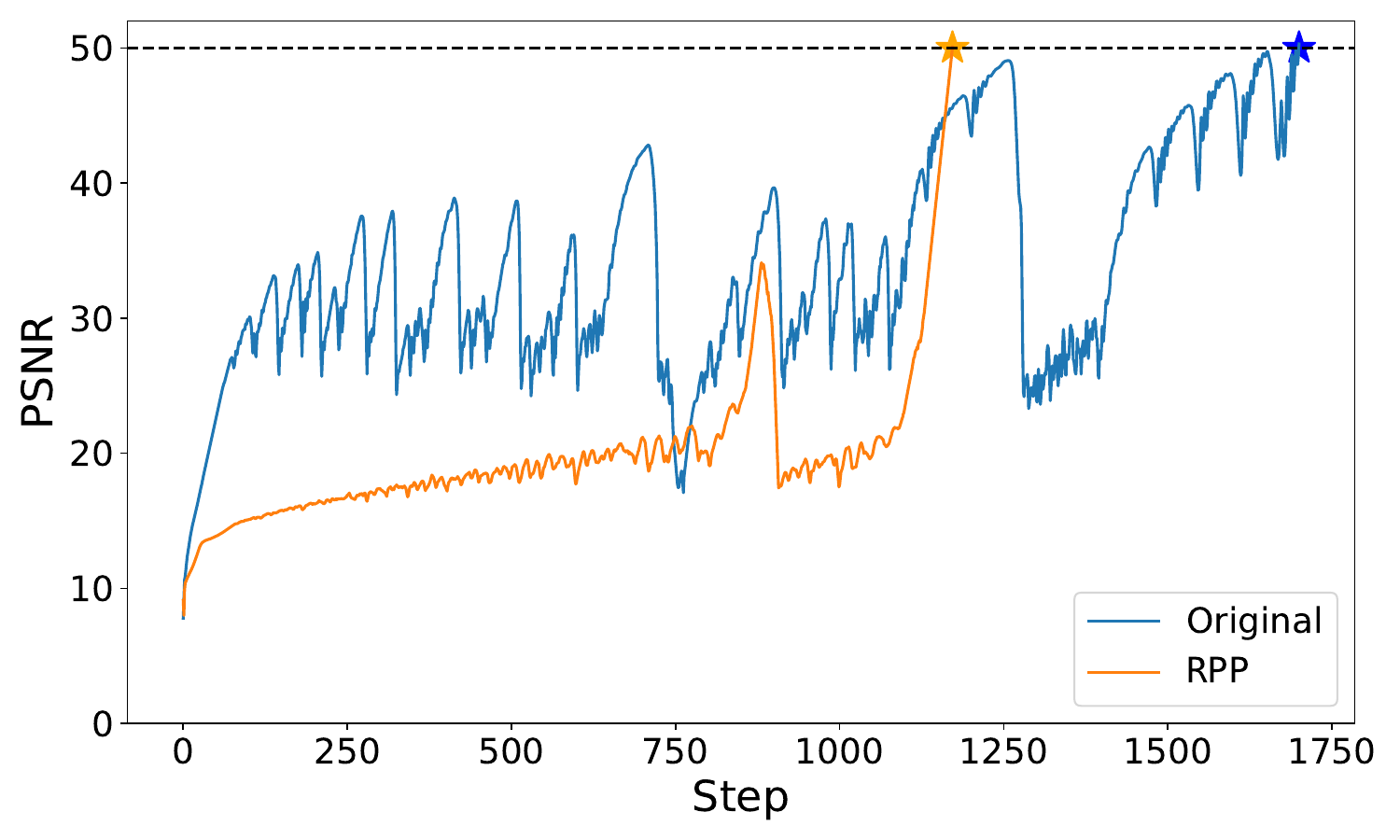}}
        \captionsetup{skip=0pt}
        \caption{Kodak\#19}
        \label{fig:lc_kodak19}
    \end{subfigure}
    \hfill
    \begin{subfigure}[t]{0.32\linewidth}
        \centering
        \fcolorbox{red}{white}{\includegraphics[width=\linewidth]{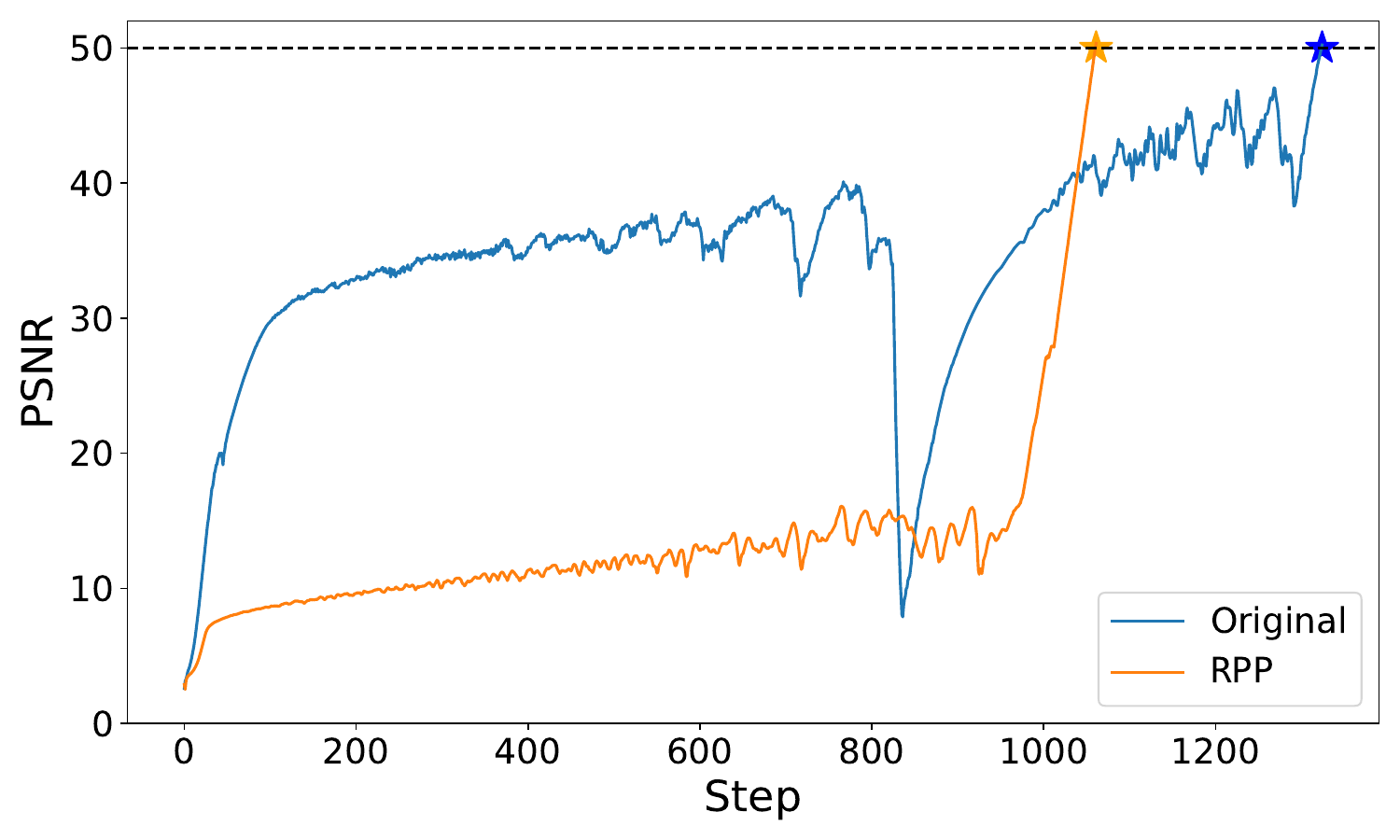}}
        \captionsetup{skip=0pt}
        \caption{Kodak\#20}
        \label{fig:lc_kodak20}
    \end{subfigure}
    \hfill
    \begin{subfigure}[t]{0.32\linewidth}
        \centering
        \includegraphics[width=\linewidth]{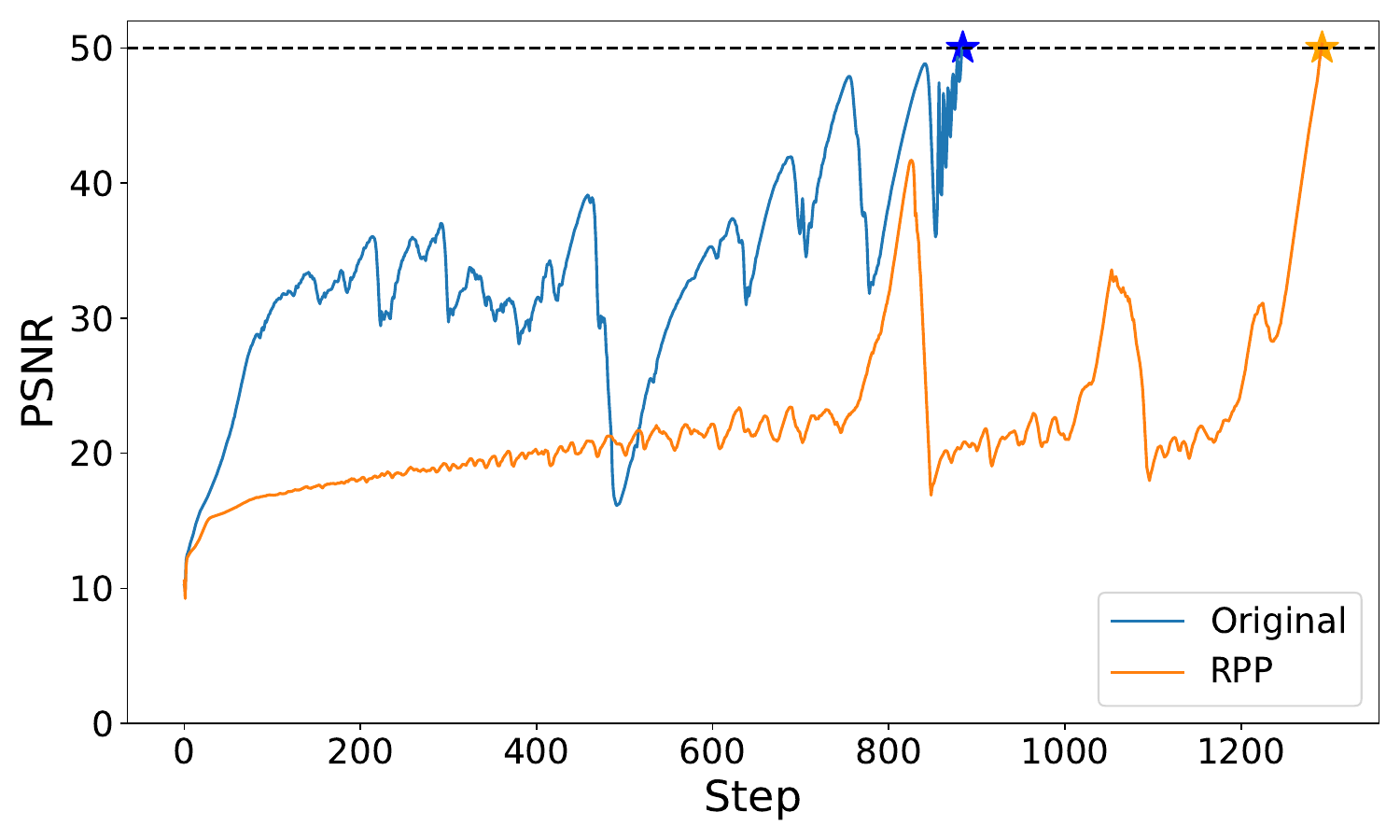}
        \captionsetup{skip=0pt}
        \caption{Kodak\#21}
        \label{fig:lc_kodak21}
    \end{subfigure}

    \vspace{0.1cm}
    \begin{subfigure}[t]{0.32\linewidth}
        \centering
        \fcolorbox{red}{white}{\includegraphics[width=\linewidth]{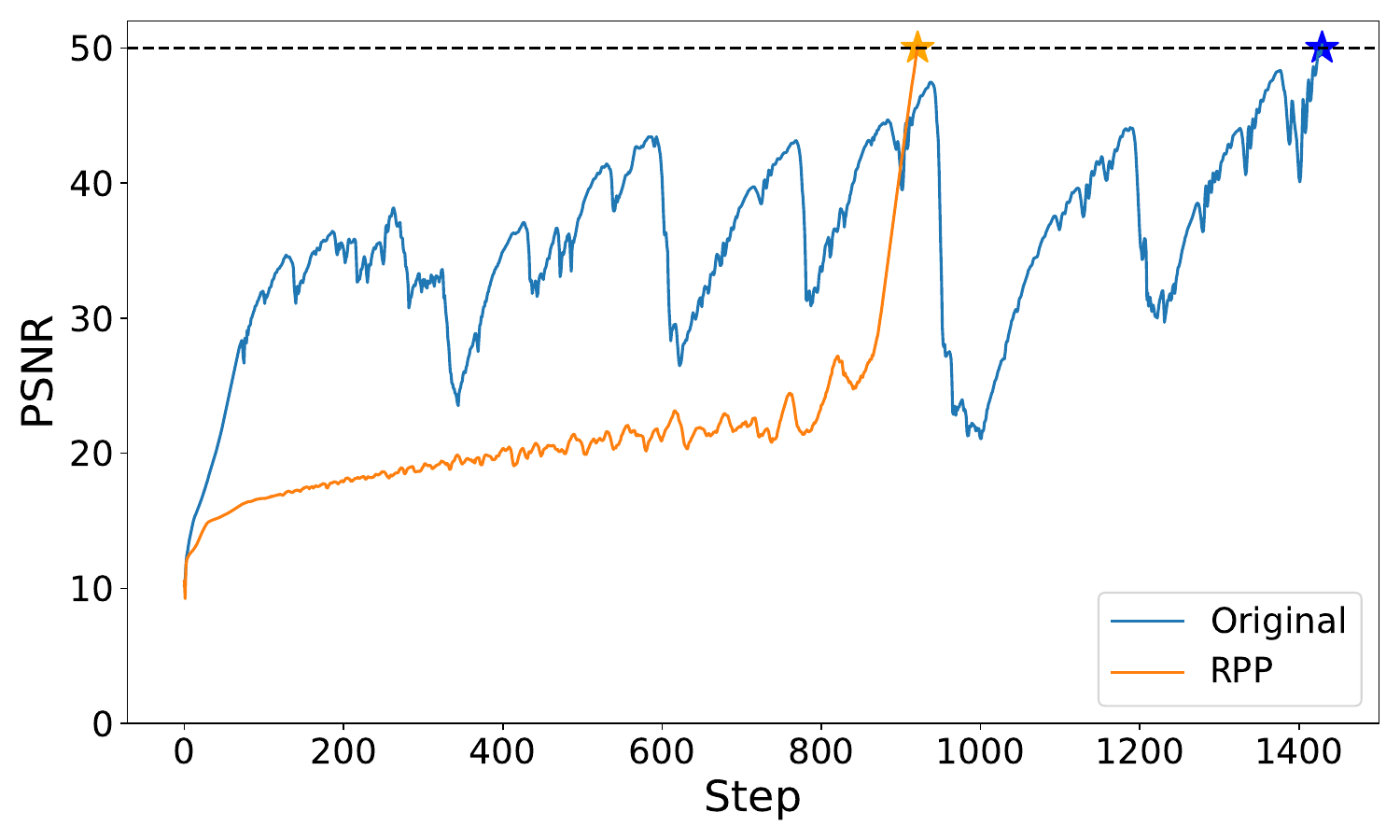}}
        \captionsetup{skip=0pt}
        \caption{Kodak\#22}
        \label{fig:lc_kodak22}
    \end{subfigure}
    \hfill
    \begin{subfigure}[t]{0.32\linewidth}
        \centering
        \includegraphics[width=\linewidth]{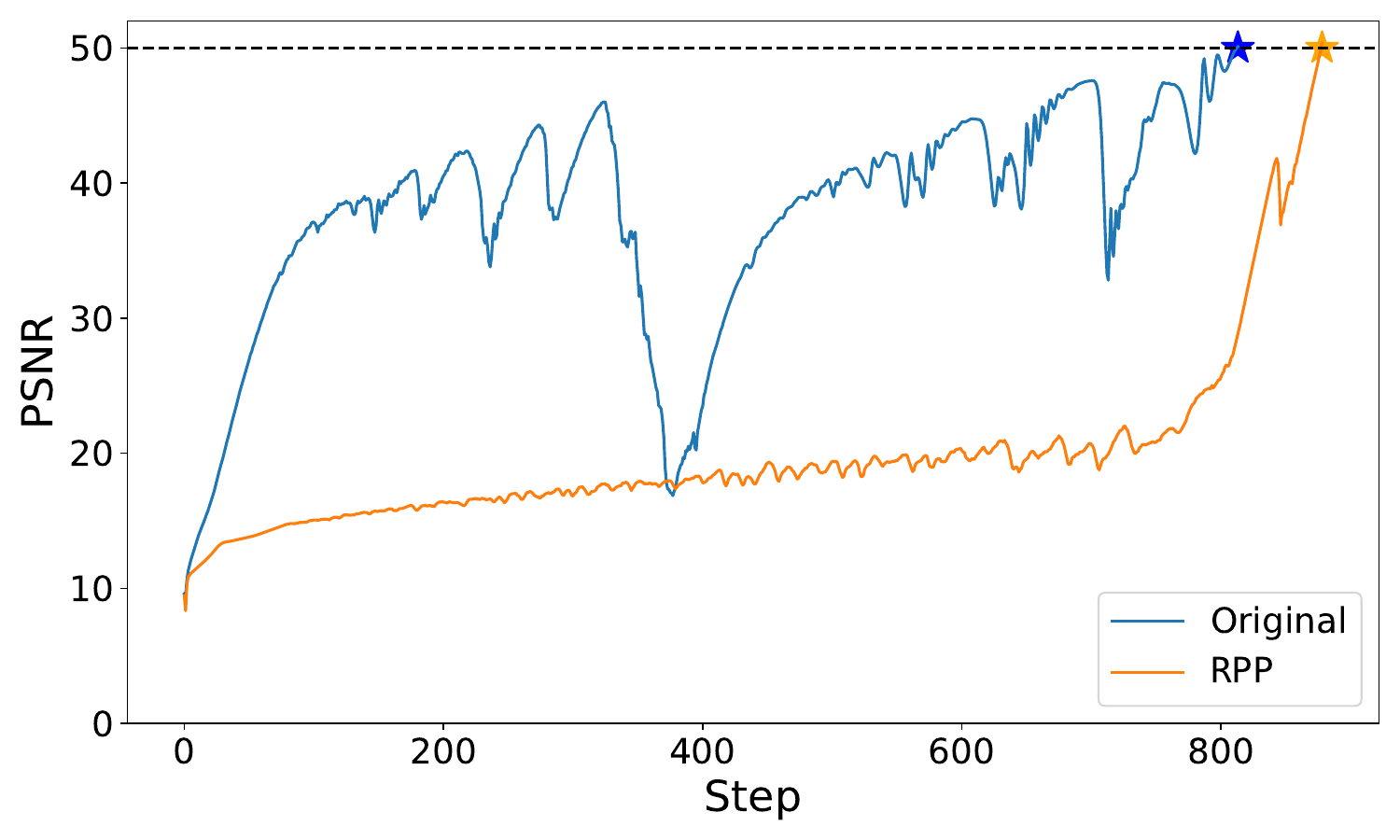}
        \captionsetup{skip=0pt}
        \caption{Kodak\#23}
        \label{fig:lc_kodak23}
    \end{subfigure}
    \hfill
    \begin{subfigure}[t]{0.32\linewidth}
        \centering
        \fcolorbox{red}{white}{\includegraphics[width=\linewidth]{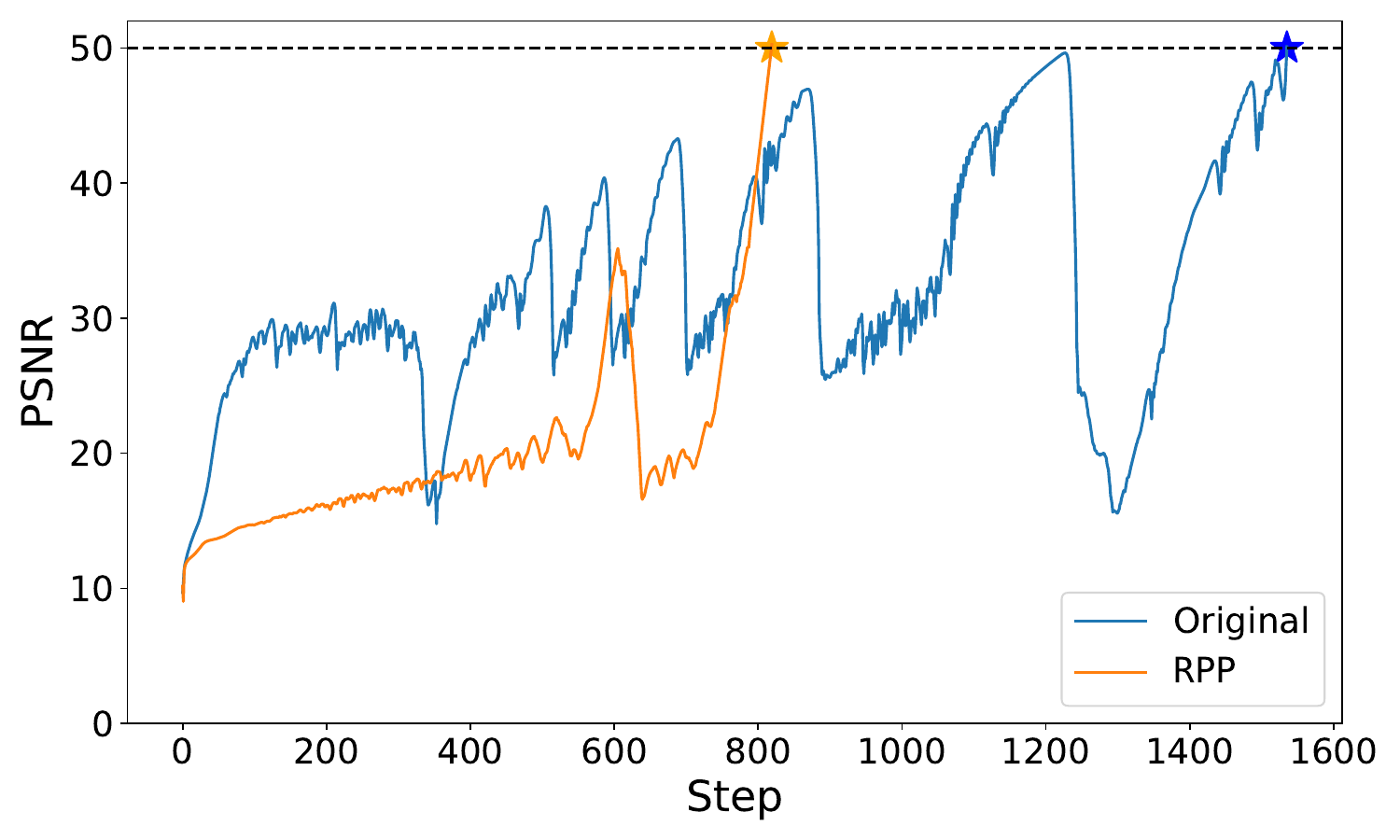}}
        \captionsetup{skip=0pt}
        \caption{Kodak\#24}
        \label{fig:lc_kodak24}
    \end{subfigure}

    \caption{\textbf{PSNR Curves of Kodak images  \#13--\#24.} We report the PSNR curves of Kodak images. The {\rpp} images quickly surge to PSNR 50dB.}
    \label{fig:lc_represent_kodak2}
    \vspace{-1em}
\end{figure*}

\Cref{fig:lc_represent_kodak1,fig:lc_represent_kodak2} show the PSNR curves of a total of 24 images in Kodak dataset. {\rpp} images fit faster than original images on 15 out of 24 images (marked with \fcolorbox{red}{white}{red borders}). Throughout all images, the {\rpp} images quickly surge to PSNR 50dB at a later stage in the majority of cases. In contrast, the original images show fluctuating PSNR values, which mostly hover above a moderate level, \ie 30dB.

\newpage

\begin{figure*}[t!]
\centering
    \begin{subfigure}[t]{0.32\linewidth}
        \centering
        \includegraphics[width=\linewidth]{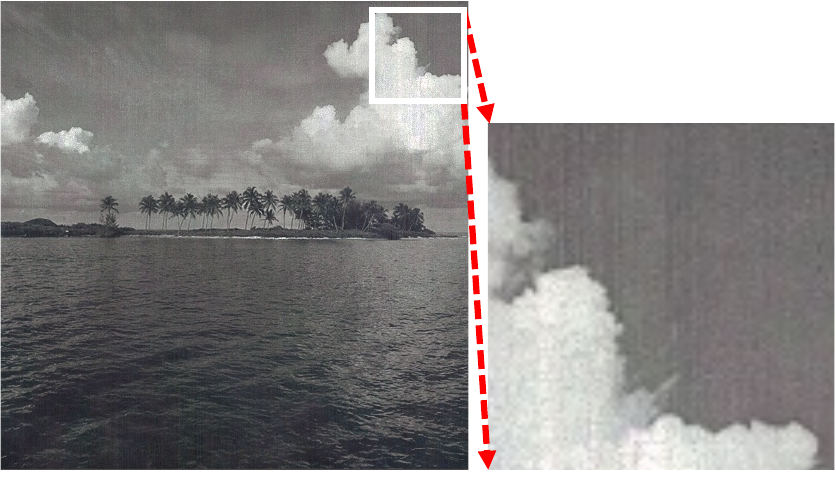}
        \caption{Kodim\#16 from permuted image}
        \label{fig:recon_ngp_16_psnr30_orig}
    \end{subfigure}
    \hfill
    \begin{subfigure}[t]{0.32\linewidth}
        \centering
        \includegraphics[width=\linewidth]{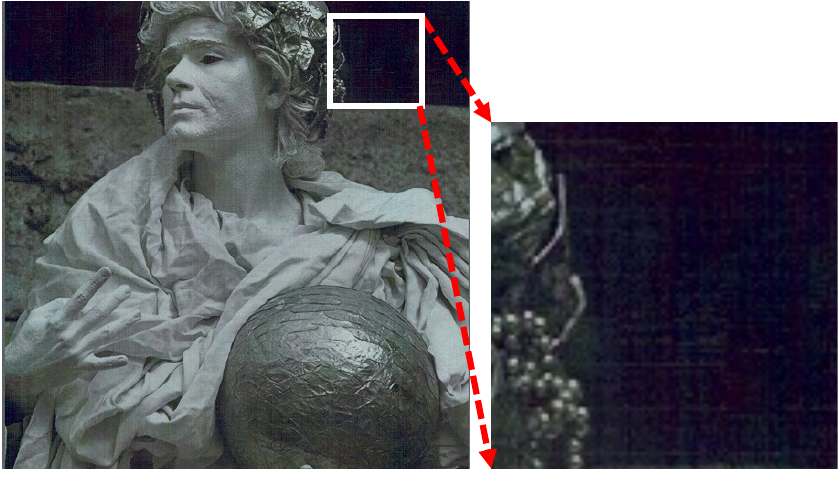}
        \caption{Kodim\#17 from permuted image}
        \label{fig:recon_ngp_17_psnr30_orig}
    \end{subfigure}
    \hfill
    \begin{subfigure}[t]{0.32\linewidth}
        \centering
        \includegraphics[width=\linewidth]{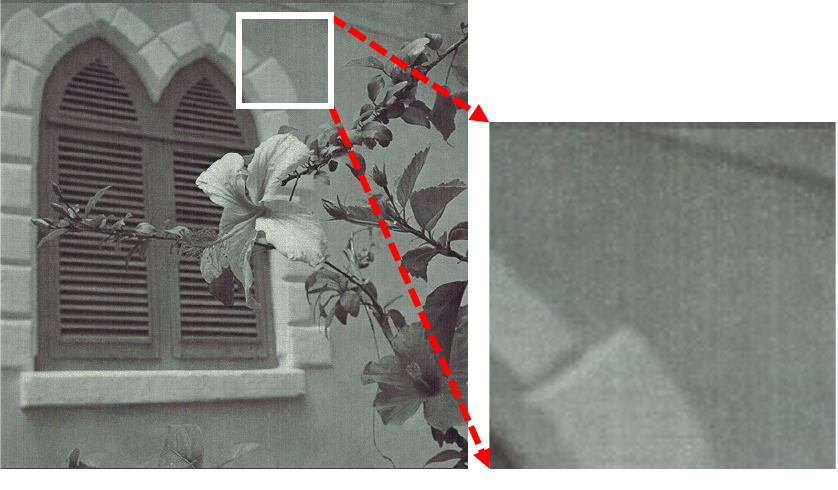}
        \caption{Kodim\#07 from permuted image}
        \label{fig:recon_ngp_7_psnr30_orig}
    \end{subfigure}

    \begin{subfigure}[t]{0.32\linewidth}
        \centering
        \includegraphics[width=\linewidth]{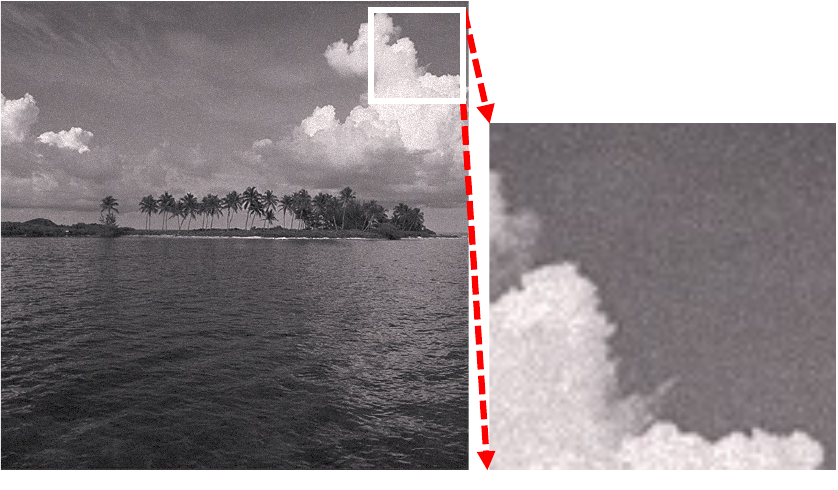}
        \caption{Kodim\#16 from permuted image}
        \label{fig:recon_ngp_16_psnr30_per}
    \end{subfigure}
    \hfill
    \begin{subfigure}[t]{0.32\linewidth}
        \centering
        \includegraphics[width=\linewidth]{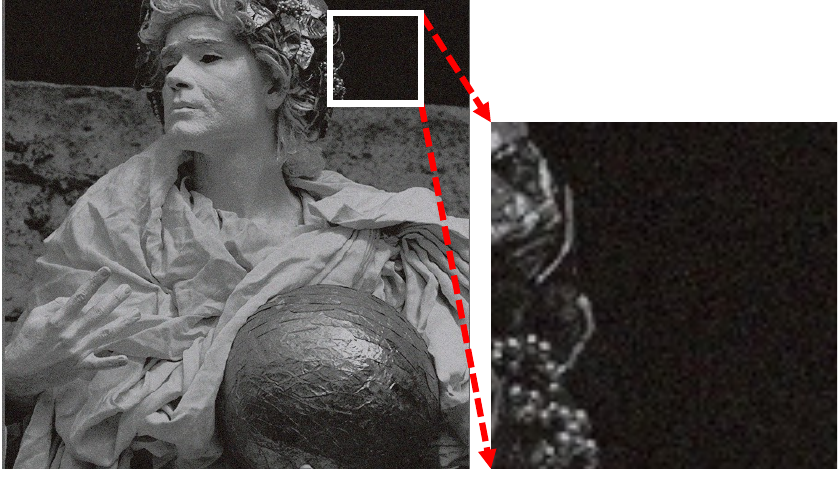}
        \caption{Kodim\#17 from permuted image}
        \label{fig:recon_ngp_17_psnr30_per}
    \end{subfigure}
    \hfill
    \begin{subfigure}[t]{0.32\linewidth}
        \centering
        \includegraphics[width=\linewidth]{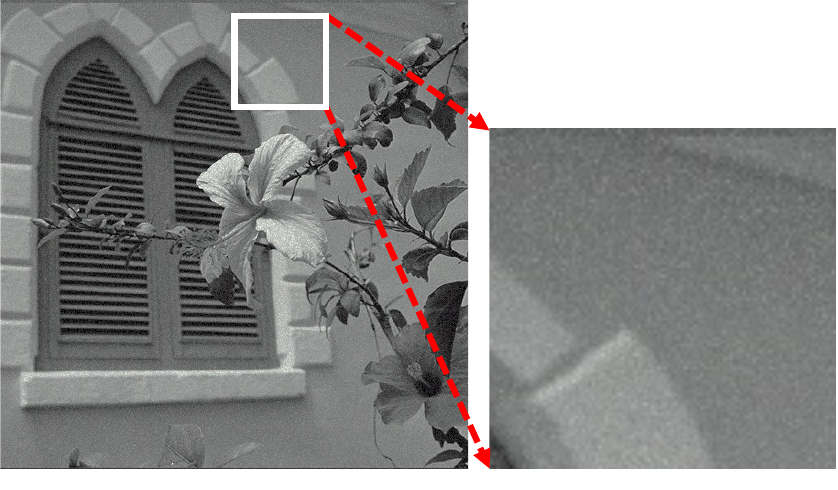}
        \caption{Kodim\#07 from permuted image}
        \label{fig:recon_ngp_7_psnr30_per}
    \end{subfigure}

    \caption{\textbf{Example reconstructions at a target PSNR value of 30dB (Instant-NGP).} We report the reconstructed images with Instant-NGP at PSNR 30dB. The original images have subtle axis-aligned artifacts. }
    \label{fig:recosntructed_image_psnr30_from_ngp}
\end{figure*}
\newpage
\quad
\newpage
\quad
\newpage
\quad
\newpage
\quad

\newpage
\section{Reconstructed images in Instant-NGP}
\label{sec:sup_reconstruct_ngp}

We present an extended analysis of the reconstructed images using Instant-NGP at a higher target PSNR value of 30dB. While the increase in PSNR typically correlates with enhanced image fidelity, our observations reveal a subtle, yet noteworthy, persistence of artifacts. These artifacts, manifesting as horizontal and vertical lines, are similar in nature to those observed at the lower PSNR of 20dB \cref{fig:recon_ngp}, albeit less pronounced. \Cref{fig:recosntructed_image_psnr30_from_ngp} illustrates these findings, showcasing the reconstructed images at PSNR 30dB. We hypothesize that these artifacts are intrinsically linked to the spatial grid encoding of the Instant-NGP model, a pattern consistent with our earlier observations at the lower PSNR threshold. 

\section{Additional loss landscapes}
\label{sec:sup_loss_landscape}
\subsection{Landscapes on other images}
\label{subsec:sup_loss_landscape_other_images}

\begin{figure*}[h]
\centering
        \begin{subfigure}[t]{0.245\linewidth}
            \centering
            \includegraphics[width=\linewidth]{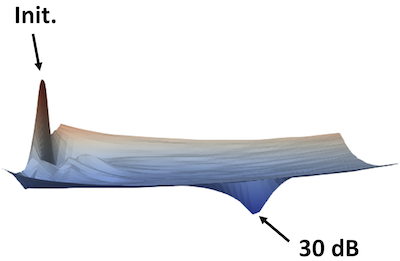}
            \caption{original; init. to 30dB}
            \label{fig:sup_kodak1_ori_30}
        \end{subfigure}
        \hfill
        \begin{subfigure}[t]{0.245\linewidth}
            \centering
            \includegraphics[width=\linewidth]{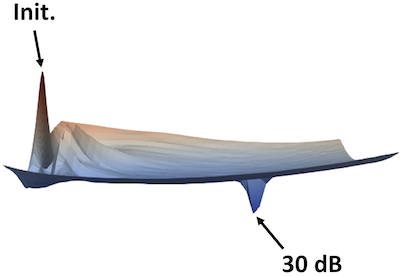}
            \caption{{\rpp}; init. to 30dB}
            \label{fig:sup_kodak1_sh_30}
        \end{subfigure}
        \hfill
        \begin{subfigure}[t]{0.245\linewidth}
            \centering
            \includegraphics[width=\linewidth]{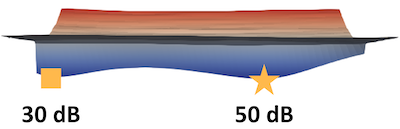}
            \caption{original; 30dB to 50dB}
            \label{fig:sup_kodak1_ori_50}
        \end{subfigure}
        \hfill
        \begin{subfigure}[t]{0.245\linewidth}
            \centering
            \includegraphics[width=\linewidth]{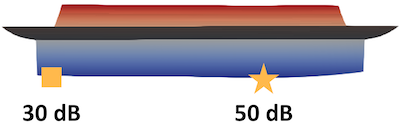}
            \caption{{\rpp} ; 30dB to 50dB}
            \label{fig:sup_kodak1_sh_50}
        \end{subfigure}
        \caption{\textbf{SIREN loss landscape: Kodak\#01 } }
\label{fig:sup_kodak0_loss_landscape}
\vspace{-1em}
\end{figure*}

\begin{figure*}[h]
\centering
        \begin{subfigure}[t]{0.245\linewidth}
            \centering
            \includegraphics[width=\linewidth]{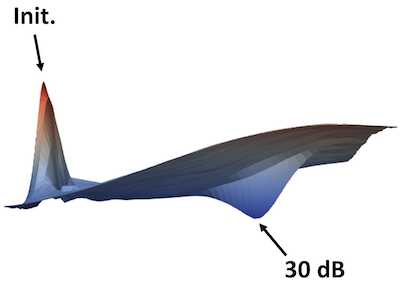}
            \caption{original; init. to 30dB}
            \label{fig:sup_kodak2_ori_30}
        \end{subfigure}
        \hfill
        \begin{subfigure}[t]{0.245\linewidth}
            \centering
            \includegraphics[width=\linewidth]{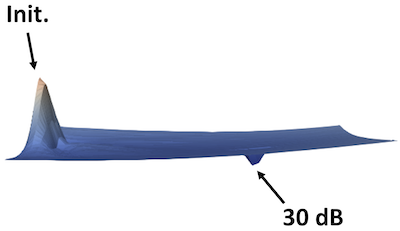}
            \caption{{\rpp}; init. to 30dB}
            \label{fig:sup_kodak2_sh_30}
        \end{subfigure}
        \hfill
        \begin{subfigure}[t]{0.245\linewidth}
            \centering
            \includegraphics[width=\linewidth]{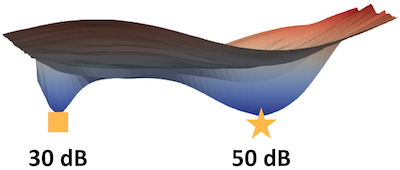}
            \caption{original; 30dB to 50dB}
            \label{fig:sup_kodak2_ori_50}
        \end{subfigure}
        \hfill
        \begin{subfigure}[t]{0.245\linewidth}
            \centering
            \includegraphics[width=\linewidth]{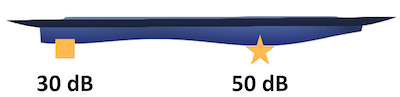}
            \caption{{\rpp} ; 30dB to 50dB}
            \label{fig:sup_kodak2_sh_50}
        \end{subfigure}
        \caption{\textbf{SIREN loss landscape: Kodak\#02} }
\label{fig:sup_kodak1_loss_landscape}
\vspace{-1em}
\end{figure*}

\begin{figure*}[h]
\centering
        \begin{subfigure}[t]{0.245\linewidth}
            \centering
            \includegraphics[width=\linewidth]{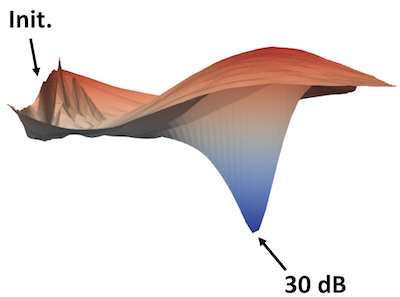}
            \caption{original; init. to 30dB}
            \label{fig:sup_div2k801_ori_30}
        \end{subfigure}
        \hfill
        \begin{subfigure}[t]{0.245\linewidth}
            \centering
            \includegraphics[width=\linewidth]{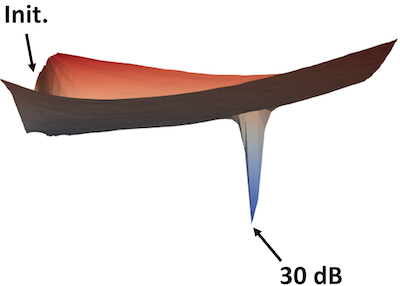}
            \caption{{\rpp}; init. to 30dB}
            \label{fig:sup_div2k801_sh_30}
        \end{subfigure}
        \hfill
        \begin{subfigure}[t]{0.245\linewidth}
            \centering
            \includegraphics[width=\linewidth]{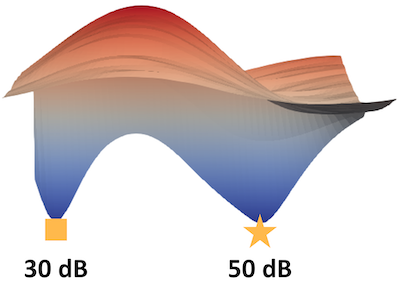}
            \caption{original; 30dB to 50dB}
            \label{fig:sup_div2k801_ori_50}
        \end{subfigure}
        \hfill
        \begin{subfigure}[t]{0.245\linewidth}
            \centering
            \includegraphics[width=\linewidth]{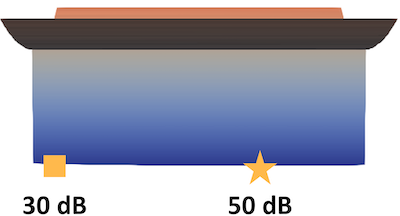}
            \caption{{\rpp} ; 30dB to 50dB}
            \label{fig:sup_div2k801_sh_50}
        \end{subfigure}
        \caption{\textbf{SIREN loss landscape: DIV2K\#801 } }
\label{fig:sup_div2k801_loss_landscape}
\vspace{-1em}
\end{figure*}

\begin{figure*}[h]
\centering
        \begin{subfigure}[t]{0.245\linewidth}
            \centering
            \includegraphics[width=\linewidth]{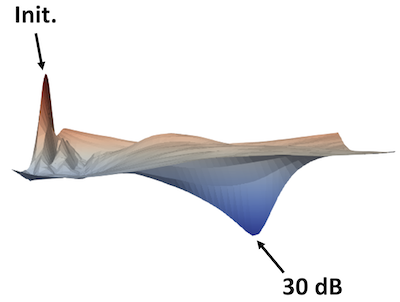}
            \caption{original; init. to 30dB}
            \label{fig:sup_div2k802_ori_30}
        \end{subfigure}
        \hfill
        \begin{subfigure}[t]{0.245\linewidth}
            \centering
            \includegraphics[width=\linewidth]{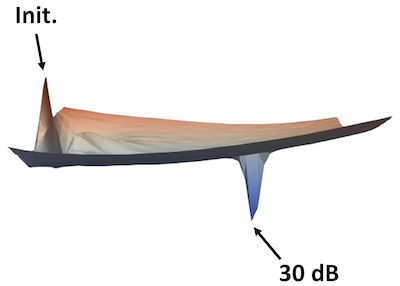}
            \caption{{\rpp}; init. to 30dB}
            \label{fig:sup_div2k802_sh_30}
        \end{subfigure}
        \hfill
        \begin{subfigure}[t]{0.245\linewidth}
            \centering
            \includegraphics[width=\linewidth]{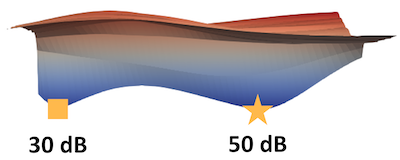}
            \caption{original; 30dB to 50dB}
            \label{fig:sup_div2k802_ori_50}
        \end{subfigure}
        \hfill
        \begin{subfigure}[t]{0.245\linewidth}
            \centering
            \includegraphics[width=\linewidth]{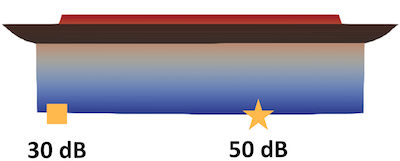}
            \caption{{\rpp} ; 30dB to 50dB}
            \label{fig:sup_div2k802_sh_50}
        \end{subfigure}
        \caption{\textbf{SIREN loss landscape: DIV2K\#802 } }
\label{fig:sup_div2k802_loss_landscape}
\vspace{-1em}
\end{figure*}

\begin{figure*}[h]
\centering
        \begin{subfigure}[t]{0.245\linewidth}
            \centering
            \includegraphics[width=\linewidth]{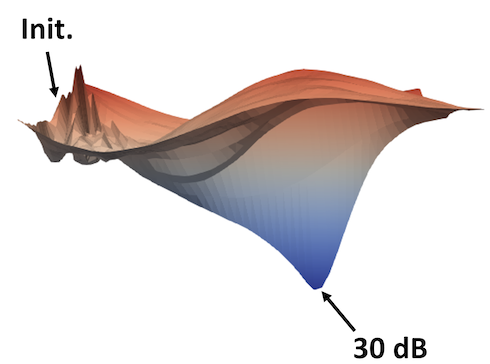}
            \caption{original; init. to 30dB}
            \label{fig:sup_clic0_ori_30}
        \end{subfigure}
        \hfill
        \begin{subfigure}[t]{0.245\linewidth}
            \centering
            \includegraphics[width=\linewidth]{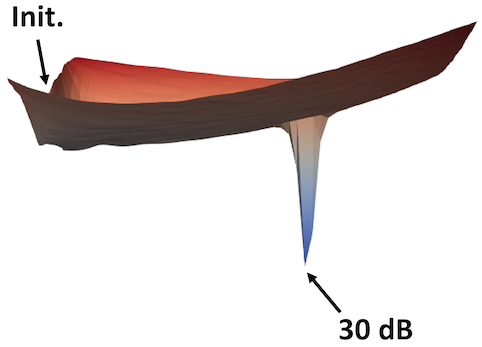}
            \caption{{\rpp}; init. to 30dB}
            \label{fig:fig:sup_clic0_sh_30}
        \end{subfigure}
        \hfill
        \begin{subfigure}[t]{0.245\linewidth}
            \centering
            \includegraphics[width=\linewidth]{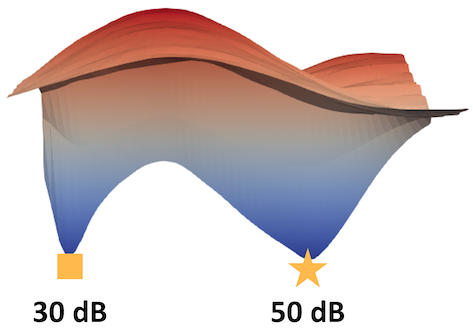}
            \caption{original; 30dB to 50dB}
            \label{fig:fig:sup_clic0_ori_50}
        \end{subfigure}
        \hfill
        \begin{subfigure}[t]{0.245\linewidth}
            \centering
            \includegraphics[width=\linewidth]{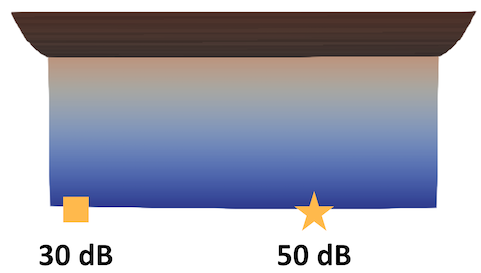}
            \caption{{\rpp} ; 30dB to 50dB}
            \label{fig:fig:sup_clic0_sh_50}
        \end{subfigure}
        \caption{\textbf{SIREN loss landscape: CLIC\#01 } }
\label{fig:sup_clic0_loss_landscape}
\vspace{-1em}
\end{figure*}

\begin{figure*}[h]
\centering
        \begin{subfigure}[t]{0.245\linewidth}
            \centering
            \includegraphics[width=\linewidth]{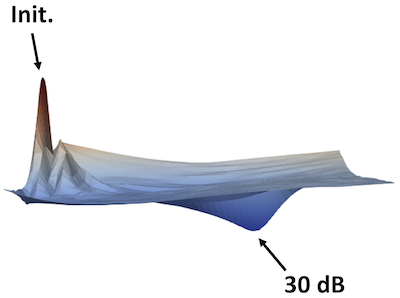}
            \caption{original; init. to 30dB}
            \label{fig:sup_clic1_ori_30}
        \end{subfigure}
        \hfill
        \begin{subfigure}[t]{0.245\linewidth}
            \centering
            \includegraphics[width=\linewidth]{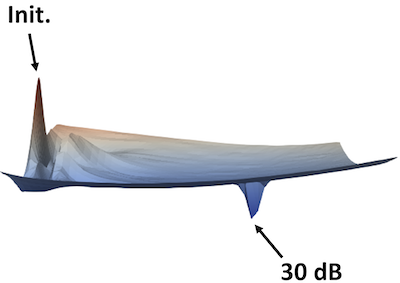}
            \caption{{\rpp}; init. to 30dB}
            \label{fig:sup_clic1_sh_30}
        \end{subfigure}
        \hfill
        \begin{subfigure}[t]{0.245\linewidth}
            \centering
            \includegraphics[width=\linewidth]{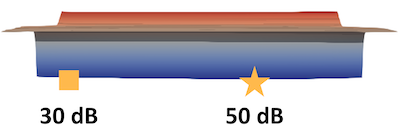}
            \caption{original; 30dB to 50dB}
            \label{fig:sup_clic1_ori_50}
        \end{subfigure}
        \hfill
        \begin{subfigure}[t]{0.245\linewidth}
            \centering
            \includegraphics[width=\linewidth]{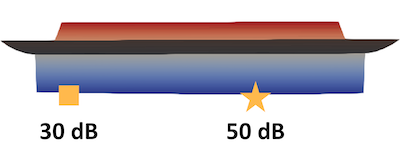}
            \caption{{\rpp} ; 30dB to 50dB}
            \label{fig:sup_clic1_sh_50}
        \end{subfigure}
        \caption{\textbf{SIREN loss landscape: CLIC\#02} In this case, the original image is quickly reaches PSNR 50dB (557 steps) than the {\rpp} image (1610 steps) }
\label{fig:sup_clic1_loss_landscape}
\vspace{-1em}
\end{figure*}

We show the loss landscapes for the first two images from Kodak, DIV2K, and CLIC. As shown in \cref{fig:loss_land_scape_0_30,fig:loss_land_scape_30_50}, we select a direction vector between two parameters and, the other direction randomly. In \cref{fig:sup_kodak0_loss_landscape,fig:sup_kodak1_loss_landscape,fig:sup_div2k801_loss_landscape,fig:sup_div2k802_loss_landscape,fig:sup_clic0_loss_landscape,fig:sup_clic1_loss_landscape}, (a) and (b) illustrate the loss landscapes during the early phase (i.e. initial point to 30dB), while (c) and (d) correspond to the late phase (i.e. 30dB to 50dB). In the early phase, as illustrated in the most of the figures, the loss landscape of the original image shows a smoother and more navigable trajectory from the initial point to a PSNR of 30dB than the {\rpp}. We also typically observe a "linear expressway' in all the loss landscapes of the {\rpp} images, which is detailed in our main paper.

\subsection{Projections on different direction vectors}
\label{subsec:sup_loss_landscape_different_vector}

\begin{figure*}[h]
\centering
        \begin{subfigure}[t]{0.245\linewidth}
            \centering
            \includegraphics[width=\linewidth]{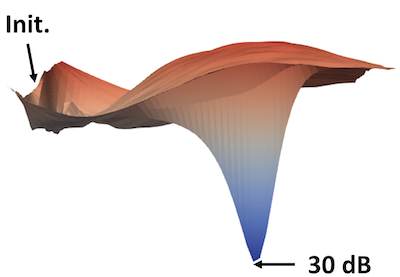}
            \caption{original; init. to 30dB}
            \label{fig:sup_kodak8_ori_30_rd}
        \end{subfigure}
        \hfill
        \begin{subfigure}[t]{0.245\linewidth}
            \centering
            \includegraphics[width=\linewidth]{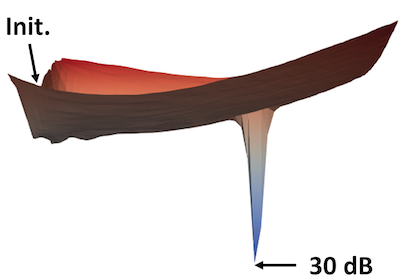}
            \caption{{\rpp}; init. to 30dB}
            \label{fig:sup_kodak8_sh_30_rd}
        \end{subfigure}
        \hfill
        \begin{subfigure}[t]{0.245\linewidth}
            \centering
            \includegraphics[width=\linewidth]{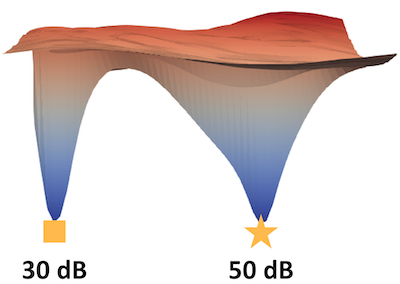}
            \caption{original; 30dB to 50dB}
            \label{fig:sup_kodak8_ori_50}
        \end{subfigure}
        \hfill
        \begin{subfigure}[t]{0.245\linewidth}
            \centering
            \includegraphics[width=\linewidth]{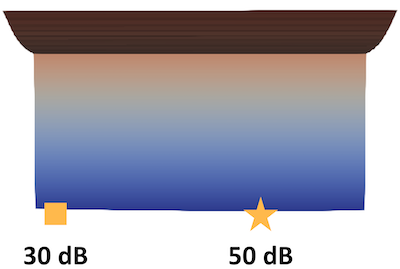}
            \caption{{\rpp} ; 30dB to 50dB}
            \label{fig:sup_kodak8_sh_50_rd}
        \end{subfigure}
        \caption{\textbf{SIREN loss landscape with different random direction: Kodak\#08}}
\label{fig:sup_kodak1_loss_landscape_rd}
\vspace{-1em}
\end{figure*}

\Cref{fig:sup_kodak1_loss_landscape_rd} shows the loss landscapes for the Kodak\#08 image, which is already used in \cref{fig:loss_land_scape_0_30,fig:loss_land_scape_30_50}. We keep the direction between two parameters, yet in this instance, another direction is chosen differently at random. Despite the difference in random directions, the overall shapes of the loss landscapes exhibit a similarity due to the insignificance of random vectors in a high-dimensional space.

\begin{figure*}[h]
\centering
        \begin{subfigure}[t]{0.245\linewidth}
            \centering
            \includegraphics[width=\linewidth]{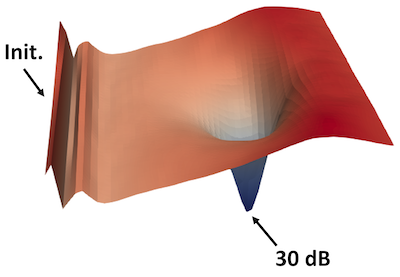}
            \caption{Elevated view; original}
            \label{fig:original_loss_landscape_0_30_hole_largest_eigen}
        \end{subfigure}
        \hfill
        \begin{subfigure}[t]{0.245\linewidth}
            \centering
            \includegraphics[width=\linewidth]{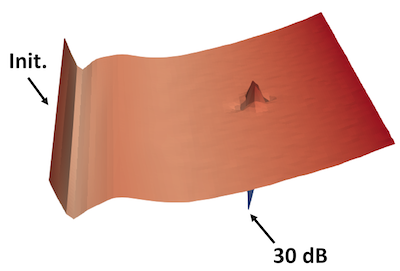}
            \caption{Elevated view; {\rpp}}
            \label{fig:permute_loss_landscape_0_30_hole_largest_eigen}
        \end{subfigure}
        \hfill
        \begin{subfigure}[t]{0.245\linewidth}
            \centering
            \includegraphics[width=\linewidth]{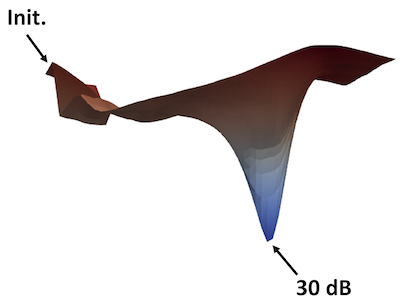}
            \caption{Side view; original}
            \label{fig:original_loss_landscape_0_30_valley_largest_eigen}
        \end{subfigure}
        \hfill
        \begin{subfigure}[t]{0.245\linewidth}
            \centering
            \includegraphics[width=\linewidth]{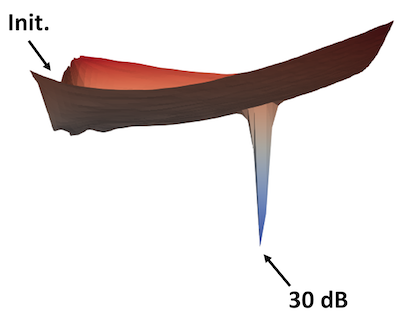}
            \caption{Side view; {\rpp}}
            \label{fig:permute_loss_landscape_0_30_valley_largest_eigen}
        \end{subfigure}
        \caption{\textbf{SIREN loss landscape with eigenvector direction, corresponding to the largest eigenvalue: Kodak\#08 (initial to 30dB)}}
\label{fig:loss_land_scape_0_30_largest_eigen}
\vspace{-1em}
\end{figure*}

\begin{figure*}[h]
\centering
        \begin{subfigure}[t]{0.245\linewidth}
            \centering
            \includegraphics[width=\linewidth]{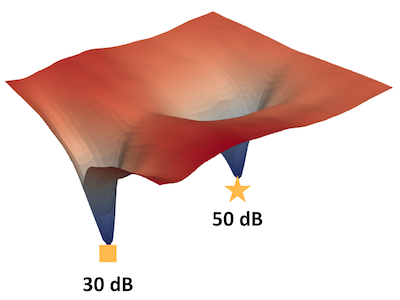}
            \caption{Elevated view; original}
            \label{fig:original_loss_landscape_0_50_hole_largest_eigen}
        \end{subfigure}
        \hfill
        \begin{subfigure}[t]{0.245\linewidth}
            \centering
            \includegraphics[width=\linewidth]{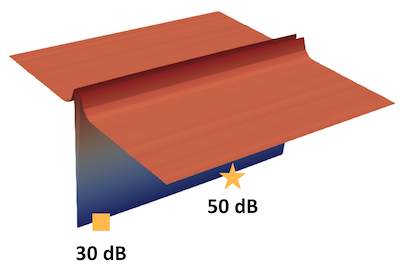}
            \caption{Elevated view; {\rpp}}
            \label{fig:permute_loss_landscape_0_50_hole_largest_eigen}
        \end{subfigure}
        \hfill
        \begin{subfigure}[t]{0.245\linewidth}
            \centering
            \includegraphics[width=\linewidth]{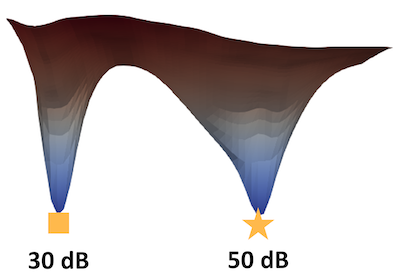}
            \caption{Side view; original}
            \label{fig:original_loss_landscape_0_50_valley_largest_eigen}
        \end{subfigure}
        \hfill
        \begin{subfigure}[t]{0.245\linewidth}
            \centering
            \includegraphics[width=\linewidth]{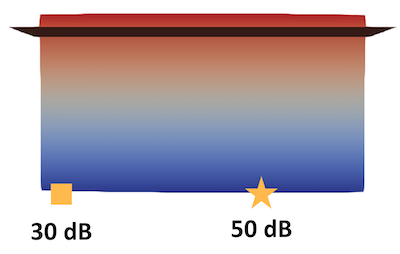}
            \caption{Side view; {\rpp}}
            \label{fig:permute_loss_landscape_0_50_valley_largest_eigen}
        \end{subfigure}
        \caption{\textbf{SIREN loss landscape with eigenvector direction, corresponding to the largest eigenvalue: Kodak\#08 (30dB to 50dB)}}
\label{fig:loss_land_scape_0_50_largest_eigen}
\vspace{-1em}
\end{figure*}

Furthermore, we provide the loss landscapes again for the Kodak\#08 image, opting not to select the direction randomly this time. Instead, we use the top-1 eigenvector, corresponding to the largest eigenvalue of the Hessian matrix of the loss function, rather than the random vector. \Cref{fig:loss_land_scape_0_30_largest_eigen,fig:loss_land_scape_0_50_largest_eigen} presents these loss landscapes, offering both elevated and side views as in \cref{fig:loss_land_scape_0_30,fig:loss_land_scape_30_50}, respectively. 

\end{document}